\documentclass{article}
\usepackage{iclr2027_conference,times}

\usepackage{amsmath,amsfonts,bm}

\def\eqref#1{equation~\ref{#1}}

\def\1{\bm{1}}

\DeclareMathAlphabet{\mathsfit}{\encodingdefault}{\sfdefault}{m}{sl}
\SetMathAlphabet{\mathsfit}{bold}{\encodingdefault}{\sfdefault}{bx}{n}

\usepackage{hyperref}
\usepackage{url}
\usepackage{booktabs}
\usepackage{multirow}
\usepackage{graphicx}
\usepackage{amsmath}
\usepackage{amssymb}
\usepackage[table]{xcolor}
\usepackage{nicematrix}
\definecolor{hred}{RGB}{168,26,38}
\definecolor{hblue}{RGB}{26,62,148}

\usepackage{wrapfig}

\newif\ifpairtables
\pairtablesfalse

\newcommand{\cmark}{\ensuremath{\checkmark}}

\newcommand{\bodyitemsep}{3pt}
\newenvironment{bodyitemize}
  {\setlength{\topsep}{\dimexpr \bodyitemsep - \parskip\relax}%
   \setlength{\partopsep}{0pt}%
   \begin{itemize}%
   \setlength{\itemsep}{\bodyitemsep}%
   \setlength{\parsep}{0pt}\setlength{\parskip}{0pt}}
  {\end{itemize}}

\newsavebox{\tblbox}
\newenvironment{fittable}
  {\begin{lrbox}{\tblbox}}
  {\end{lrbox}%
   \ifdim\wd\tblbox>\linewidth
     \resizebox{\linewidth}{!}{\usebox{\tblbox}}%
   \else
     \usebox{\tblbox}%
   \fi}

\newcommand{\nModels}{26}
\newcommand{\nMinRuns}{10}

\newcommand{\nDatasets}{15}

\newcommand{\nTotalRuns}{56K}

\newcommand{\timeGain}{+0.026}
\newcommand{\chanGain}{+0.000}
\newcommand{\timeRank}{1.72}
\newcommand{\chanRank}{1.0028}

\newcommand{\nStdRuns}{1{,}126}

\newcommand{\generalDatasets}{ETTh1, ETTh2, ETTm1, ETTm2, Electricity, Exchange, PEMS03, PEMS04, PEMS07, PEMS08, Solar, Traffic, and Weather}

\newcommand{\domGeneralRank}{1.84}

\newcommand{\domFinancialRank}{1.85}

\newcommand{\sfGain}{+0.179}
\newcommand{\sfGainSeed}{0.1625}
\newcommand{\sfGainConfig}{0.4138}
\newcommand{\sfGainN}{345}
\newcommand{\sfMargin}{+0.237}
\newcommand{\sfMarginSeed}{0.1227}
\newcommand{\sfMarginConfig}{0.2404}

\newcommand{\sfRank}{1.94}
\newcommand{\sfRankSeed}{0.1326}
\newcommand{\sfRankConfig}{1.0749}

\newcommand{\cmpModels}{26}
\newcommand{\cmpDatasets}{15}
\newcommand{\cmpSeeds}{3}
\newcommand{\cmpShapes}{4}

\newcommand{\calibN}{349}

\newcommand{\calibRankMax}{16}
\newcommand{\calibModels}{28}

\newcommand{\limMaxCorr}{0.07}
\newcommand{\limPeriodCorr}{-0.169}

\newcommand{\relSurrogate}{TimeX++ (ICML'24)}
\newcommand{\relLearned}{Dynamask (ICML'21), ExtremalMask (ICML'23), ContraLSP (ICLR'24)}
\newcommand{\relGradient}{Saliency (ICLR-W'14), Integrated gradients (ICML'17), TSR (NeurIPS'20), TimeSHAP (KDD'21), TsSHAP (arXiv'23)}

\newcommand{\fmRuns}{16{,}729}
\newcommand{\fmModels}{8}

\newcommand{\fmShuffle}{+0.118}

\newcommand{\fmAdapterErr}{2e-6}

\newcommand{\fmDataSig}{32}
\newcommand{\fmDataNeg}{15}
\newcommand{\fmDataTotal}{66}

\newcommand{\gtmModels}{26}
\newcommand{\gtmGens}{6}

\newcommand{\syN}{1{,}386}

\newcommand{\syGens}{6}
\newcommand{\syShapes}{3}
\newcommand{\sySeeds}{3}
\newcommand{\syPerCell}{54}

\title{Explaining Time Series Forecasting \\ with Horizon-Resolved Attribution}

\author{%
Seunghan Lee, Jun Seo, Jaehoon Lee, Junhyeok Kang, Sangjun Han, Sungdong Yoo, \\
\bfseries Minjae Kim, Tae Yoon Lim, Dongwan Kang, Hwanil Choi, Soonyoung Lee, Wonbin Ahn \\
\normalfont LG AI Research
}

\iclrfinalcopy

\begin{document}

\maketitle
\lhead{Preprint}

\maketitle

\begin{abstract}
Recent advances in explaining time series (TS) models have produced methods that identify
which past values a prediction depends on.
However, most existing methods return a \textit{single} importance vector, assuming that every
predicted step depends on the \textit{same} past values.
In this paper, we show that this assumption does not hold, as
\textit{different forecast steps depend on different past values}.
Motivated by this observation, we propose \textbf{H}orizon-\textbf{R}esolved e\textbf{X}planation
(\textbf{HRX}), which adds a \textit{horizon axis} to the explanation, so that every forecast
step receives its \textit{own} importance map.
HRX is a simple yet effective plug-in framework with three components: 1) an
\textbf{estimator} that reads these maps out of any differentiable forecaster without modifying
the TS backbone,
2) an \textbf{evaluation protocol} that validates the horizon axis by measuring how much a single
forecast step changes when the inputs an importance map ranks highest are removed, and 3) a \textbf{rank criterion} that predicts in advance
whether the axis is worth resolving on a given TS.
We further show that this step-wise dependence is \textit{low-dimensional}, as the explanations of all
steps are built from a few shared maps whose number does not grow with the forecast length.
Extensive experiments across various backbones and datasets show that the improvement comes
from the horizon axis and holds for estimators of previous explanation methods.
Code is available at \url{https://github.com/seunghan96/HRX}.
\end{abstract}

\section{Introduction}

Time series (TS) forecasting is widely used in various fields, including finance
\citep{lee2026beyond, lee2026finstar} and traffic \citep{liu2024itransformer}.
A range of TS forecasting methods have been developed based on different architectures, such as
Transformers \citep{liu2024itransformer}, multi-layer perceptrons (MLPs)
\citep{zeng2023dlinear, lee2024learning}, and convolutional networks
\citep{wu2023timesnet}.
In parallel, TS explanation methods ask \textit{why a model produces a particular
output} \citep{queen2023timex}.

However, existing TS explanation methods return a \textit{single explanation vector} $e \in \mathbb{R}^{L}$ for
the entire output, whether that output is a class label (classification) or an entire forecast
horizon (forecasting).
A forecast, unlike a class label, is a trajectory of $H$ values, and a single vector collapses it
into one summary.
This implicitly assumes that every future value depends on the \textit{same} past values.

\begin{wrapfigure}{r}{0.47\linewidth}
\centering
\vspace{-13.2pt}
\includegraphics[width=\linewidth]{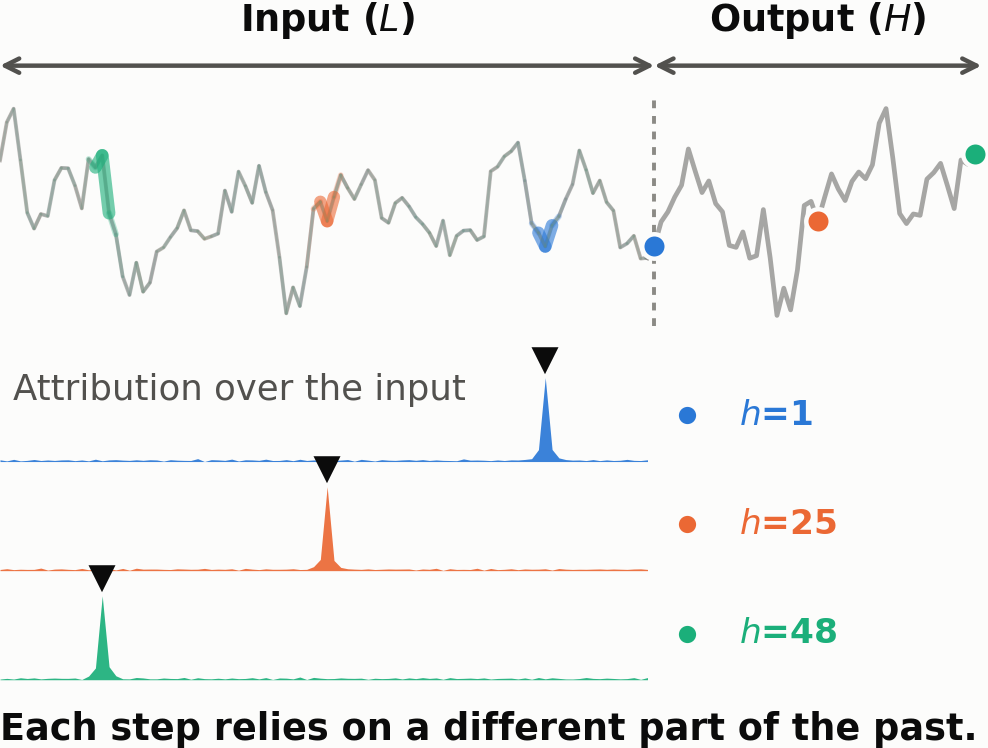}
\caption{\textbf{Different steps, different evidence.} Each step needs its own explanation.}
\label{fig:teaser}
\vspace{-11pt}
\end{wrapfigure}

In this paper, we argue that \textbf{\emph{each forecast step needs its own explanation}}, not one shared
across all horizons, since different steps may rely on different past values.
Figure~\ref{fig:teaser} shows this on a synthetic series, where three forecast steps of the
same forecast rely on three different parts of the past.
To this end, we propose \textbf{H}orizon-\textbf{R}esolved e\textbf{X}planation (\textbf{HRX}), a
simple yet effective plug-in framework that explains each forecast step separately instead
of the entire forecast as a whole.
Specifically, it replaces the explanation \textit{vector} $e \in \mathbb{R}^{L}$, which is shared across all
forecast steps, with a \textit{matrix} $E \in \mathbb{R}^{H \times L}$, one row per forecast step, and applies to any differentiable
forecaster without modifying the TS backbone.
Note that our contribution is the \textit{horizon axis rather than the estimator} that
produces each row of $E$, since the estimator of any existing explanation method can fill a row and we adopt an
input gradient as our default estimator only because it is the simplest.

We further propose an evaluation protocol that scores one forecast step at a time, which no prior
metric can do.
We also propose a rank criterion, the rank of $E$, that says in advance whether the axis brings
any gain over the shared vector on a given TS.
The main contributions are summarized as:

\begin{bodyitemize}
\item We show that TS forecasters read different parts of the lookback window for different
forecast steps, and that the single explanation vector returned by prior works cannot express
this structure.

\item We propose \textbf{H}orizon-\textbf{R}esolved e\textbf{X}planation (\textbf{HRX}), a simple yet
effective plug-in framework that resolves the explanation along the horizon axis, so that the
explanation becomes a matrix with one row per forecast step.
HRX is applicable to the estimator of any explanation method and to any differentiable
forecaster, and needs no change to the TS backbone or auxiliary training.

\item We propose an evaluation protocol, since established TS explanation metrics score one
label per input and therefore cannot score a single forecast step. It measures how much one step
degrades when we delete what its own row of $E$ marks rather than what the shared vector marks.

\item We show that the horizon axis is low-dimensional, as keeping only the leading directions
of $E$ recovers the full gain while a single direction recovers none of it. We turn the rank of
$E$ into a criterion that predicts the gain from the matrix alone, without running the protocol.

\item We conduct extensive experiments over \nTotalRuns{} runs across \nModels{} backbones and
\nDatasets{} benchmarks, and show that the improvement comes from the horizon axis rather than
any particular estimator. The axis also holds under the explanation metrics of prior works and
against known ground truth.
\end{bodyitemize}

\section{Related Works}
\label{sec:related}

\begin{table}[t]
\caption{\textbf{Comparison with existing TS explanation methods.} Our work is the first to
resolve the horizon axis, returning one importance map per forecast step rather than one for the
entire forecast, while remaining a plug-in that needs no change to the backbone and no auxiliary
training.}
\label{tab:related}
\begin{center}
\begin{fittable}\begin{tabular}{l|cc|cc|c}
\toprule
 & \multicolumn{2}{c|}{\textbf{Resolved axes}} & \multicolumn{2}{c|}{\textbf{Requirements}} & \\
\cmidrule(lr){2-3}\cmidrule(lr){4-5}
\raisebox{1.6ex}[0pt][0pt]{\textbf{Method}} & \textbf{Time} & \textbf{Horizon} & \textbf{Plug-in} & \textbf{No train.} & \raisebox{1.6ex}[0pt][0pt]{\textbf{Explanation}} \\
\midrule
Surrogate methods\textsuperscript{1} & \cmark &  &  &  & $\mathbb{R}^{L}$ \\
Learned mask methods\textsuperscript{2} & \cmark &  & \cmark &  & $\mathbb{R}^{L}$ \\
Gradient and Shapley methods\textsuperscript{3} & \cmark &  & \cmark & \cmark & $\mathbb{R}^{L}$ \\
\rowcolor{yellow!18} \textbf{HRX (Ours)} & \cmark & \cmark & \cmark & \cmark & $\mathbb{R}^{H \times L}$ \\
\bottomrule
\end{tabular}
\end{fittable}
\end{center}
{\scriptsize\setlength{\parindent}{0pt}\setlength{\leftskip}{1.6em}
\textsuperscript{1}\relSurrogate{}.\par
\textsuperscript{2}\relLearned{}.\par
\textsuperscript{3}\relGradient{}.\par}
\end{table}

\noindent\textbf{Three lines of related work (Appendix~\ref{sec:relatedfull}).}
Work related to ours falls into three lines:
\begin{bodyitemize}
\item TS forecasting models \citep{nie2023patchtst, liu2024itransformer}.
\item Gradient-based explanation \citep{simonyan2013saliency, sundararajan2017ig}.
\item Perturbation-based explanation \citep{crabbe2021dynamask, enguehard2023extremal}.
\end{bodyitemize}
\noindent The first is what we attach to, since our framework treats any forecaster as a black
box and changes nothing inside it.
The second reads importance from the derivative of the output with respect to the input, which
is what keeps it cheap, while the third changes the input and reads importance from how much the
output moves, which is what makes it expensive.
Details are discussed in Appendix~\ref{sec:relatedfull}.

\noindent\textbf{Positioning of our work.}
Every method above returns one explanation vector per instance, and that vector is attached
\textit{to the whole forecast rather than to any one step}.
Table~\ref{tab:related} compares TS explanation methods along the axes an explanation can
resolve.
Every method resolves the time axis, while ours is the only one that also resolves the
\textit{forecast step}.
Additional related works are discussed in Appendix~\ref{sec:extrarelated}, and details of the
backbones we use are in Appendix~\ref{sec:setup}.

\begin{figure}[t]
\centering
\includegraphics[width=0.95\linewidth]{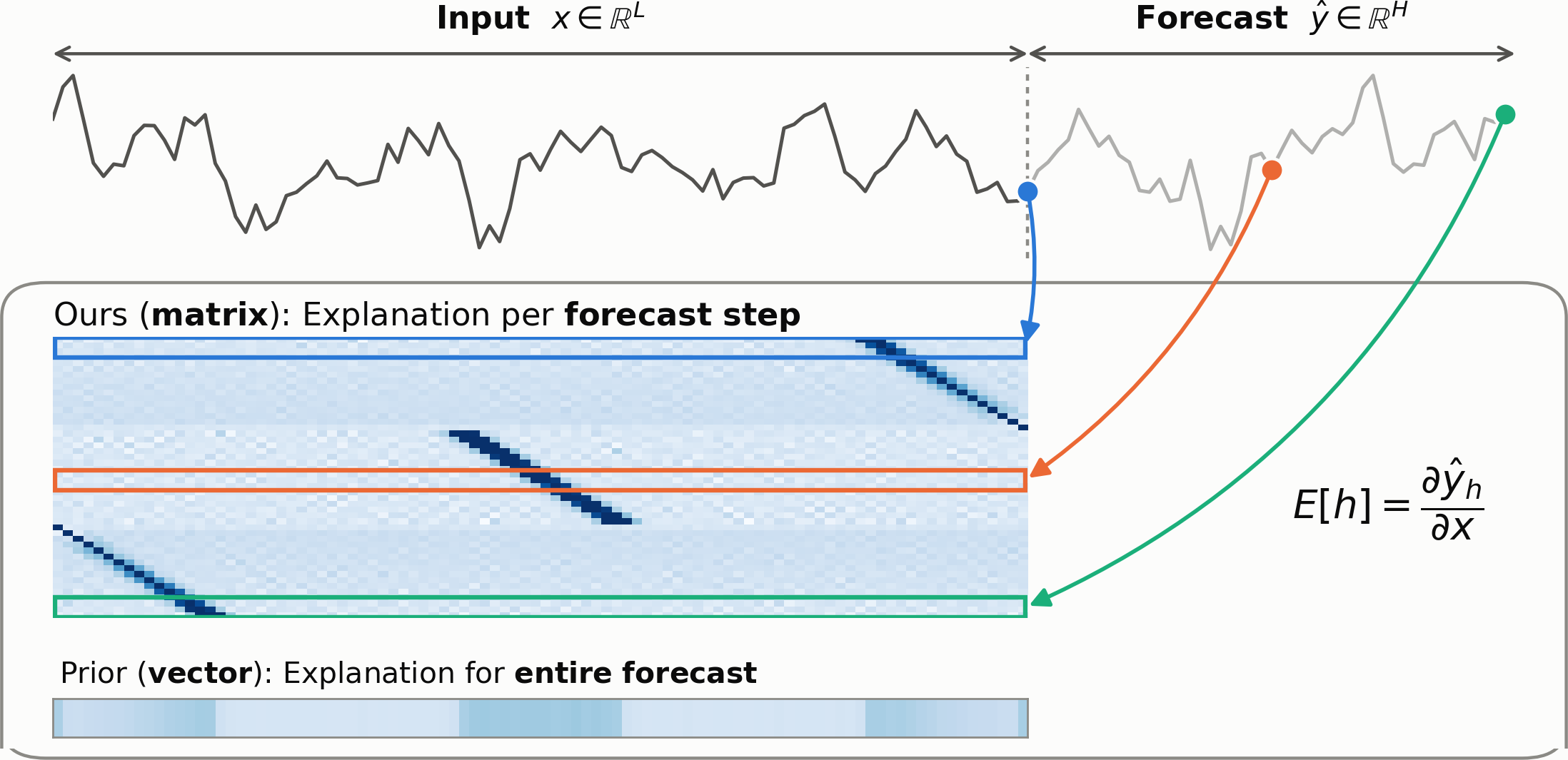}
\caption{\textbf{Explanation matrix $E$.} A forecaster reads the input and outputs $H$ steps, and HRX
returns one row of $E$ per step, where each colored row explains the step in the same color
above.}
\label{fig:matrix}
\end{figure}

\section{Methodology}
\label{sec:method}

The framework has four parts: (Section~\ref{sec:matrix}) what an explanation should be,
(Section~\ref{sec:estimator}) how we compute it, (Section~\ref{sec:protocol}) how we validate it,
and (Section~\ref{sec:rank}) how many directions it needs.
\begin{itemize}
\setlength{\itemsep}{0pt}\setlength{\parsep}{0pt}
\item \textbf{Explanation matrix.} From one shared vector to a matrix with one row per forecast step.
\item \textbf{Estimator.} The input gradient of one forecast step fills a row, and any estimator can replace it.
\item \textbf{Evaluation protocol.} A test of whether a row explains its own step better than another row.
\item \textbf{Rank criterion.} The singular values of $E$ say how many of the $H$ rows are effectively distinct.
\end{itemize}

\subsection{Explanation Matrix}
\label{sec:matrix}

\noindent\textbf{From explanation vector to matrix.}
Let $f_\theta : \mathbb{R}^{L} \to \mathbb{R}^{H}$ be a trained forecaster that maps a lookback
window $x$ of length $L$ to a forecast $\hat{y} = f_\theta(x)$ of length $H$.
Existing TS explanation methods produce $e \in \mathbb{R}^{L}$, a single vector over the lookback
that every forecast step has to share, which we call the \textit{shared vector} throughout.
As shown in Figure~\ref{fig:matrix}, we instead ask for
\begin{equation}
E \in \mathbb{R}^{H \times L}, \qquad
E[h, t] = \text{importance of input } t \text{ for the forecast at step } h .
\end{equation}
The shared vector of prior works is recovered as $e = \sum_h E[h]$, so $E$
\textit{strictly refines it rather than replacing it}.
Note that $E$ is only useful \textit{when its rows actually differ}, since a matrix whose rows are copies
of one vector carries no more than $e$ does, and Section~\ref{sec:rank} makes that condition
measurable.
We take the univariate setting as our default, and the extension to multivariate TS follows
below.

\noindent\textbf{Extension to multivariate TS.}
A forecaster with $C$ channels, $f_\theta : \mathbb{R}^{L \times C} \to \mathbb{R}^{H \times C}$,
carries a second input axis, so the explanation of prior work becomes
$e \in \mathbb{R}^{L \times C}$ and ours becomes
\begin{equation}
\tilde{E} \in \mathbb{R}^{H \times L \times C}, \qquad
\tilde{E}[h, t, c] = \text{importance of input } (t, c) \text{ for the forecast at step } h .
\end{equation}
Note that \textit{nothing in the framework depends on the channel axis}, as a row belongs to one
forecast step no matter how many input axes there are.
We choose to report the time axis alone, taking
$E[h, t] = \sum_{c} \lvert \tilde{E}[h, t, c] \rvert$ and keeping the $H \times L$ form
throughout, for two reasons.
First, which channels a forecaster reads does not change with the forecast step, so resolving
that axis would store the same values once per step (Section~\ref{sec:timechan}).
Second, the fixed $H \times L$ form makes the explanation the same shape on every dataset, so a
rank or a gain is directly comparable across different $C$.

\subsection{Estimator}
\label{sec:estimator}

\noindent\textbf{Gradient of a single forecast step.}
For a differentiable $f_\theta$, we take the row of $E$ at step $h$ to be the input gradient of
that step alone, which is $E[h] = \partial \hat{y}[h] / \partial x$.
For a multivariate forecaster we differentiate the channel sum $\sum_{c} \hat{y}[h, c]$ and then
reduce it over input channels as above.
Note that HRX is not tied to this gradient, since \textit{any estimator conditioned on a
single output coordinate} can fill a row of $E$.
We adopt the gradient form as our default estimator because it is the simplest, needing no
optimization of its own, and any of seven other estimators can take its place
(Section~\ref{sec:methods}).

\noindent\textbf{Efficient implementation: Batched backward calls.}
A backward call propagates from a single scalar, so a naive implementation needs one pass per
forecast step and $H$ passes in total.
To fill several rows in one call, we replicate the input, give each replica a different $h$, and
differentiate the sum of the selected outputs, which keeps the rows separate because
\textit{each replica is its own leaf in the graph}.
The cost of the whole matrix then stays within a small multiple of a forward pass.

\begin{figure}[t]
\centering
\includegraphics[width=0.91\linewidth]{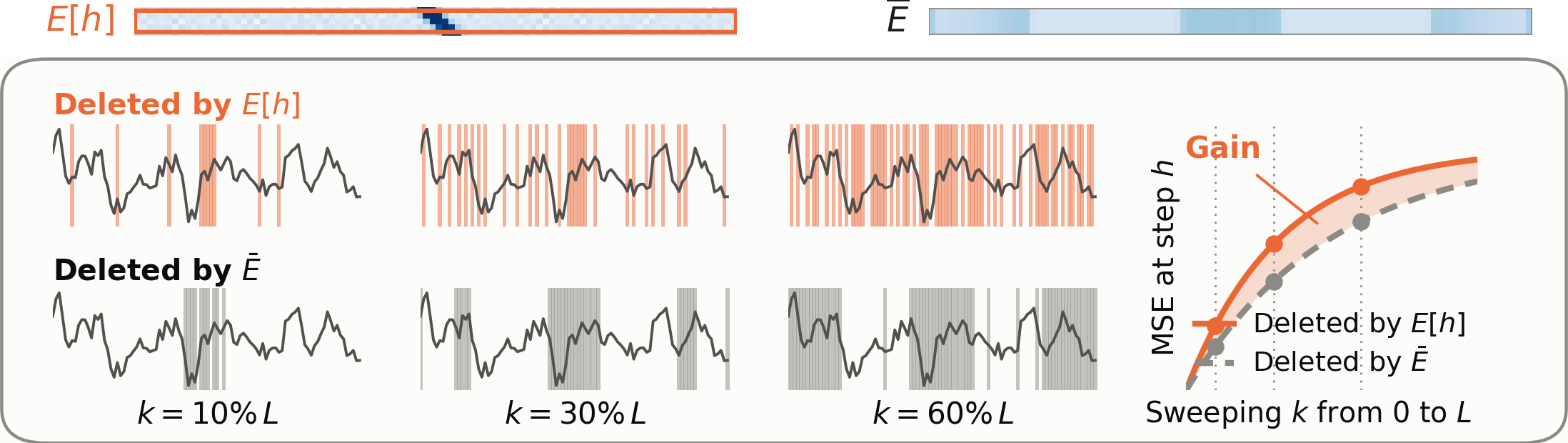}
\caption{\textbf{Evaluation protocol.} For a forecast step $h$ we delete the top-$k$ lookback
positions in two orders: (1) by its own row $E[h]$ and (2) by the shared vector $\bar{E}$, and
sweep $k$ from $0$ to $L$. The gap between the two error curves, that is
$\mathrm{AUC}(h; E[h]) - \mathrm{AUC}(h; \bar{E})$, is the gain.}
\label{fig:protocol}
\end{figure}

\subsection{Evaluation Protocol}
\label{sec:protocol}

An explanation is scored by \textit{deletion}
\citep{liu2024timexpp, liu2024contralsp}, which \textit{removes the positions
the explanation ranks highest}\footnote{Removed positions are resampled from elsewhere in the
same window rather than set to zero, which keeps the input in distribution
\citep{hooker2019roar}.} and measures how much the forecast changes
(Figure~\ref{fig:protocol}).
Since removing more positions changes the forecast further, sweeping the number removed traces
a curve rather than a scalar, summarized by its area (AUC).
As a forecast has $H$ values, we read the error one step at a time and
\textit{compare the AUCs of different importance vectors}, giving the quantities
of Table~\ref{tab:protocol}.

\begin{table}[t]
\caption{\textbf{Explanation quantities.} Each compares two AUCs with different
importance vectors (own step $E[h]$ vs. other step $E[\pi(h)]$ vs. shared $\bar{E}$). Last column
gives the ideal direction, where a step's own row should break that step more than the shared
vector does and another step's row less.}
\label{tab:protocol}
\begin{center}
\small
\begin{fittable}
\begin{tabular}{@{}c@{\ }l|ccc}
\toprule
 & \textbf{Quantity} & \textbf{AUC comparison} & \textbf{Question it answers} & \textbf{Ideal dir.} \\
\midrule
 & $g_{\text{own}}$ & $E[\textcolor{hred}{h}]$ vs. $\bar{E}$ & Does \textbf{\textcolor{hred}{own step's}} row explain it better than the shared one? & $> 0$ \\
$-$ & $g_{\text{shuffled}}$ & $E[\textcolor{hblue}{\pi(h)}]$ vs. $\bar{E}$ & Does \textbf{\textcolor{hblue}{another step's}} row explain it better than the shared one? & $< 0$ \\
\midrule
$=$ & $g_{\text{margin}}$ & $E[\textcolor{hred}{h}]$ vs. $E[\textcolor{hblue}{\pi(h)}]$ & Does \textbf{\textcolor{hred}{own step's}} row explain it better than \textbf{\textcolor{hblue}{another step's}} row? & $> g_{\text{own}}$ \\
\bottomrule
\end{tabular}
\end{fittable}
\end{center}
\end{table}

\noindent\textbf{[1] Gain with own explanation ($g_{\text{own}}$).}
Write $\mathrm{AUC}(h; v)$ for the AUC at step $h$ with importance vector $v$ (i.e., the top-$k$
lookback positions are deleted in the order $v$ ranks them, sweeping $k$ from $0$ to $L$).
The error is read at step $h$ alone rather than over the whole forecast, no matter which $v$
chose the deleted positions.
Then, the gain with own explanation ($g_{\text{own}}$) compares 1) the row of $E$ that belongs to
step $h$ ($E[h]$) against 2) the shared vector $\bar{E} = \frac{1}{H}\sum_{h'} E[h']$, so it asks
whether step $h$ is better explained by its own row than by one vector shared across the
horizon,
\begin{equation}
g_{\text{own}}(h) \;=\; \mathrm{AUC}\bigl(\textcolor{hred}{h};\, E[\textcolor{hred}{h}]\bigr)
                       - \mathrm{AUC}\bigl(\textcolor{hred}{h};\, \bar{E}\bigr),
\qquad
g_{\text{own}} \;=\; \frac{1}{|\mathcal{H}|}\sum_{h \in \mathcal{H}} g_{\text{own}}(h) .
\label{eq:gain}
\end{equation}
A positive $g_{\text{own}}$ says that deleting by a step's own row breaks that step more than
deleting by the shared vector does, so the row points at what that step actually uses.
A method without a horizon axis is pinned to $g_{\text{own}} = 0$, since the two terms then read
the same vector.

\noindent\textbf{[2] Gain with other explanation ($g_{\text{shuffled}}$).}
A positive $g_{\text{own}}$ is not enough, as a row of $E$ is concentrated on fewer
positions than $\bar{E}$ is, and deleting a few concentrated positions
breaks the forecast more than deleting the same number of scattered ones, whether or not they
are the right positions.
Thus, $g_{\text{own}}$ can be positive even when the row is not the right one for step $h$.
To rule that out, we score step $h$ with the row of another step, drawing a random permutation
$\pi$ of $\mathcal{H}$ and using $E[\pi(h)]$,
\begin{equation}
g_{\text{shuffled}}(h) \;=\; \mathrm{AUC}\bigl(\textcolor{hred}{h};\, E[\textcolor{hblue}{\pi(h)}]\bigr)
                            - \mathrm{AUC}\bigl(\textcolor{hred}{h};\, \bar{E}\bigr),
\qquad
g_{\text{shuffled}} \;=\; \frac{1}{|\mathcal{H}|}
                          \sum_{h \in \mathcal{H}} g_{\text{shuffled}}(h) .
\label{eq:shuffle}
\end{equation}
The only difference between $g_{\text{shuffled}}$ and $g_{\text{own}}$ is which row is used to
explain step $h$.
The shared vector keeps a share of every row, including the one that belongs to step $h$, whereas
another step's row keeps none of it, \textit{so the ideal value of $g_{\text{shuffled}}$ is
negative rather than merely zero}.

\noindent\textbf{[1$-$2] Shuffle margin ($g_{\text{margin}}$).}
The decisive quantity of the protocol is their difference,
\begin{equation}
g_{\text{margin}} \;=\; g_{\textcolor{hred}{\text{own}}} - g_{\textcolor{hblue}{\text{shuffled}}} ,
\label{eq:margin}
\end{equation}
which we average over three permutations of $\mathcal{H}$.
The shared vector appears in both gains and cancels, leaving
$\mathrm{AUC}(\textcolor{hred}{h}; E[\textcolor{hred}{h}])
- \mathrm{AUC}(\textcolor{hred}{h}; E[\textcolor{hblue}{\pi(h)}])$, so the margin compares 1) the row of
step $h$ ($E[\textcolor{hred}{h}]$) directly against 2) the row of another step
($E[\textcolor{hblue}{\pi(h)}]$).
The margin therefore measures only \textit{whether a row is used on the step it was computed
for}, and is expected not only to be positive but to exceed $g_{\text{own}}$, since the
ideal value of $g_{\text{shuffled}}$ is negative.

\noindent\textbf{Scale normalization.}
All three quantities are error differences, so they grow with the variance of the forecast.
Without a correction, a dataset whose targets vary more would look more horizon-dependent than
one whose targets are flat.
We therefore divide each of them by $\mathrm{Var}(\hat{y})$.

\subsection{Rank of the Explanation Matrix}
\label{sec:rank}

The matrix is worth its $H$ rows only when those rows differ.
We check it with the singular values of $E$, which say how many are effectively distinct, taking
the effective rank to be the participation ratio \citep{litwinkumar2017dim}
\begin{equation}
r(E) \;=\; \bigl(\textstyle\sum_i \sigma_i^2\bigr)^2 \big/ \textstyle\sum_i \sigma_i^4 ,
\qquad \sigma_1 \ge \dots \ge \sigma_{\min(H,L)} \ge 0 ,
\label{eq:rank}
\end{equation}
which counts how many directions carry the magnitude of the matrix rather than how many
are nonzero.
Truncating $E$ to the leading $r$ of those directions writes every row as
\begin{equation}
E_r[h] \;=\; \textstyle\sum_{k=1}^{r} c_{h,k} \, v_k ,
\qquad c_{h,k} = \sigma_k u_k[h] ,
\label{eq:trunc}
\end{equation}
where $u_k$ and $v_k$ are the singular vectors of $E$, so that the $r$ maps
$v_1, \dots, v_r \in \mathbb{R}^{L}$ are shared by every forecast step and only the weights
$c_{h,k}$ change with $h$.
At $r = 1$ every row is a positive multiple of $v_1$, whereas a larger $r$ lets the rows rank the
lookback differently.
Note that $r(E) = 1$ is the setting of prior works, which pins every quantity of
Table~\ref{tab:protocol} to zero.
Unlike the evaluation protocol in Section~\ref{sec:protocol}, $r(E)$ needs only the matrix itself
and no deletion sweep, so it says before any measurement whether the axis is worth resolving on
a given TS.

\section{Experiments}

\noindent\textbf{Datasets.}
We evaluate on \nDatasets{} benchmarks grouped into two domains.
The first is a widely used general-domain suite, which is \generalDatasets{}
\citep{wu2021autoformer}.
The second is a new financial benchmark we assemble from the daily closing prices of the 100
largest listed US companies between 2018 and 2025 \citep{yfinance}.
We add the financial domain, where decisions have to be justified to auditors and regulators
\citep{yeo2024financialxai}.
Details are discussed in Appendix~\ref{sec:setup}.

\noindent\textbf{Backbones.}
We build on the Time-Series-Library \citep{wang2024tslib} and evaluate \nModels{} backbones
spanning linear, convolutional, recurrent, and Transformer families.
We add three deliberately minimal architectures, which are a single linear map, a small CNN, and
a vanilla Transformer, so that the phenomenon can be separated from any modern design choice.
Details are discussed in Appendix~\ref{sec:setup}.

\noindent\textbf{Experimental setup.}
To test the horizon axis where it would have to hold in practice, we sweep the lookback length,
the forecast length, and the depth over the ranges the forecasting literature uses, and we run
every configuration with three random seeds.
Details are discussed in Appendix~\ref{sec:setup}.

\noindent\textbf{Metrics.}
We report our own protocol of Section~\ref{sec:protocol}, where $g_{\text{own}}$ asks whether a
step is better explained by its own row than by one shared vector and $g_{\text{margin}}$ asks
how much of that needs the rows matched to their steps.
We also report metrics of prior works, which are 1) AUPRC, AUROC, AUP and AUR on the synthetic
data, where we know the ground truth, and 2) Comprehensiveness, AOPC, Sufficiency and Insertion
on the real-world data, with details discussed in Appendix~\ref{sec:stddefs}.

\subsection{Sanity Check on Synthetic Dataset}
\label{sec:sanity}

\noindent\textbf{Why this check comes first.}
On real data, we never know which past values a forecast step actually depends on.
We therefore cannot tell whether our method reports the truth or only an artifact of the metric.
To find out, we build six kinds of synthetic dataset in which we decide that dependence
ourselves and check what our method returns, which we assess both qualitatively and
quantitatively.

\noindent\textbf{Synthetic datasets.}
The target of each generator is an explicit function of a few chosen lookback positions, so \textit{the
dependence is fixed by how we build the data} rather than asserted afterwards.
Note that $\partial y_h / \partial x_t$ then follows from the definition and is exactly zero at
every position we did not choose.
Each generator takes the same form $y_h = w^{(h)} \cdot x_{S(h)}$, where $S(h)$ is the small set
of lookback positions that step $h$ reads and $w^{(h)}$ is a unit-norm weight vector that varies
smoothly with $h$.
They differ only in how $S(h)$ moves with $h$, and Appendix~\ref{sec:gt} defines all \syGens{}
of them.

\begin{table}[t]
\caption{\textbf{Explanation performance on the synthetic dataset.} On the ground-truth metrics
the explanation matrix beats the shared vector. On the proposed metrics the shared vector has no
per-step value, so every $g$ is zero, while ours is ideally $g_{\text{own}} > 0$,
$g_{\text{shuffled}} < 0$, $g_{\text{margin}} > g_{\text{own}}$. \textbf{Bold} marks the
outcome the horizon axis predicts, here and in every table that follows.}
\label{tab:synth}
\begin{center}
\small
\begin{fittable}\begin{NiceTabular}{l|c|c|c|cc|cc|cc|cc}
\toprule
\Block{3-1}{\textbf{Backbone}} & \multicolumn{3}{c|}{\textbf{Proposed metrics}\textsuperscript{1}} & \multicolumn{8}{c}{\textbf{Ground-truth metrics}\textsuperscript{2}} \\
\cmidrule(lr){2-4}\cmidrule(lr){5-12}
 & \Block{2-1}{\textbf{$g_{\text{own}}$}} & \Block{2-1}{\textbf{$g_{\text{shuffled}}$}} & \Block{2-1}{\textbf{$g_{\text{margin}}$}} & \multicolumn{2}{c|}{\textbf{AUPRC}} & \multicolumn{2}{c|}{\textbf{AUROC}} & \multicolumn{2}{c|}{\textbf{AUP}} & \multicolumn{2}{c}{\textbf{AUR}} \\
\cmidrule(lr){5-6}\cmidrule(lr){7-8}\cmidrule(lr){9-10}\cmidrule(lr){11-12}
 &  &  &  & Vector & Matrix & Vector & Matrix & Vector & Matrix & Vector & Matrix \\
\midrule
Linear & \textcolor{hred}{$\mathbf{0.142}$} & \textcolor{hblue}{$\mathbf{-0.215}$} & \textcolor{hred}{$\mathbf{0.357}$} & $0.586$ & \textcolor{hred}{$\mathbf{0.983}$} & $0.961$ & \textcolor{hred}{$\mathbf{0.998}$} & $0.132$ & \textcolor{hred}{$\mathbf{0.154}$} & $0.960$ & \textcolor{hred}{$\mathbf{0.985}$} \\
CNN & \textcolor{hred}{$\mathbf{0.161}$} & \textcolor{hblue}{$\mathbf{-0.234}$} & \textcolor{hred}{$\mathbf{0.395}$} & $0.578$ & \textcolor{hred}{$\mathbf{0.993}$} & $0.962$ & \textcolor{hred}{$\mathbf{0.999}$} & $0.132$ & \textcolor{hred}{$\mathbf{0.154}$} & $0.963$ & \textcolor{hred}{$\mathbf{0.984}$} \\
Transformer & \textcolor{hred}{$\mathbf{0.121}$} & \textcolor{hblue}{$\mathbf{-0.252}$} & \textcolor{hred}{$\mathbf{0.373}$} & $0.577$ & \textcolor{hred}{$\mathbf{0.993}$} & $0.960$ & \textcolor{hred}{$\mathbf{0.999}$} & $0.130$ & \textcolor{hred}{$\mathbf{0.154}$} & $0.963$ & \textcolor{hred}{$\mathbf{0.985}$} \\
\bottomrule
\end{NiceTabular}
\end{fittable}
\end{center}
{\scriptsize\setlength{\parindent}{0pt}\setlength{\leftskip}{1.6em}
\textsuperscript{1}A method without a horizon axis has only the shared vector, so $E[h]$ and
$\bar{E}$ are the same vector there and every $g$ is \textit{exactly zero}.\par
\textsuperscript{2}AUPRC, AUROC, AUP and AUR compare an explanation against the positions that
actually generated the target, which we know here because we built the generator. They are
therefore \textit{defined on this synthetic data and not on real data}.\par}
\end{table}

\begin{figure}[t]
\centering
\includegraphics[width=\linewidth]{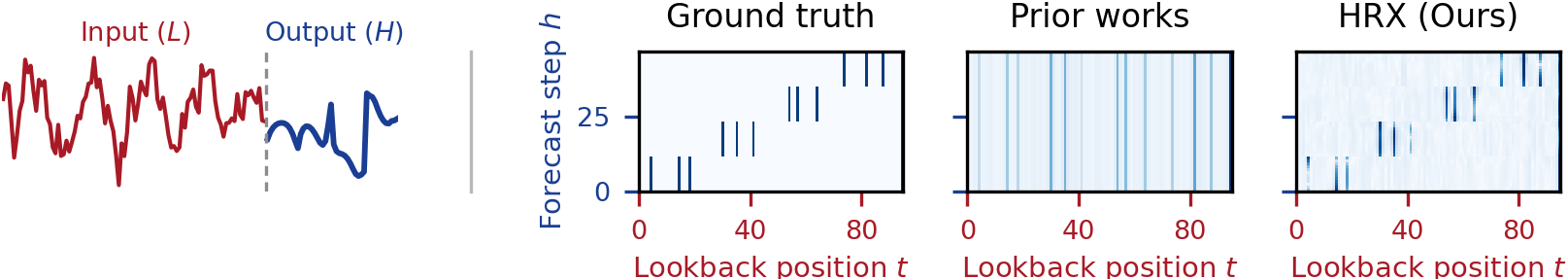}
\caption{\textbf{Explanation on synthetic dataset.} \textrm{[Left]} TS forecasting
visualization. \textrm{[Right]} 1) Ground truth, 2) shared vector that stands in for prior
works, and 3) explanation matrix that HRX returns.}
\label{fig:synth}
\end{figure}

\noindent\textbf{Quantitative results.}
Table~\ref{tab:synth} reports the performance over the six synthetic datasets (generators) with
three basic backbones, where every cell is a median over \syPerCell{} runs, namely \syGens{}
generators, \syShapes{} window shapes, and \sySeeds{} seeds.
The explanation matrix beats the shared vector on all four ground-truth metrics and all three
backbones.
On our proposed metrics, all three backbones show the ideal direction, $g_{\text{own}} > 0$,
$g_{\text{shuffled}} < 0$, and $g_{\text{margin}} > g_{\text{own}}$.
Note that these three are exactly zero for prior works, since they score one forecast step at
a time and the shared vector has no per-step value.

\noindent\textbf{Qualitative results.}
To validate that the matrix recovers the true dependence, we compare the positions that actually
generated each forecast step with what each method returns on the same trained backbone
(Figure~\ref{fig:synth}).
Prior works repeat the shared vector down the horizon, which flattens the structure the
generator planted into vertical stripes.
The explanation matrix keeps one row per forecast step and reproduces that structure, and
Appendix~\ref{sec:gt} shows the same for every generator.

\subsection{Main Experiments}
\label{sec:mainresults}

\begin{figure}[t]
\centering
\includegraphics[width=\linewidth]{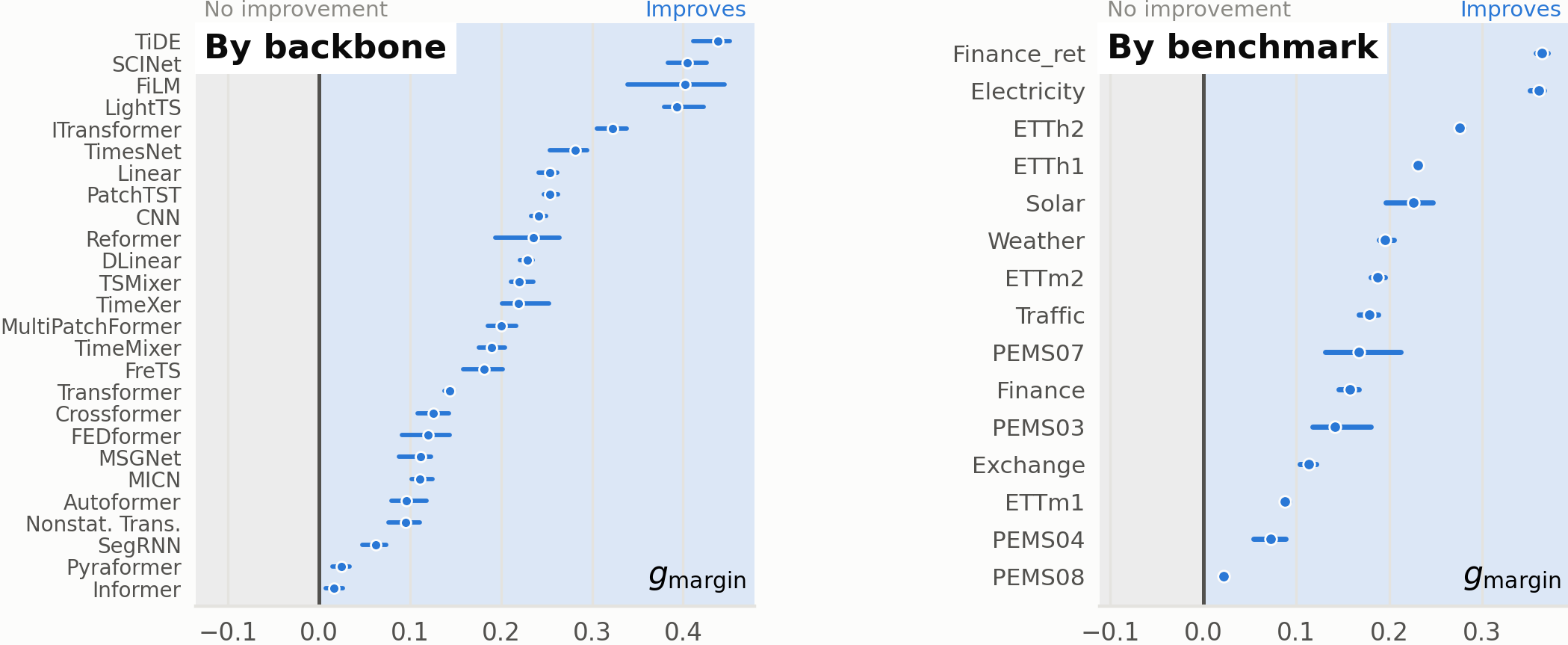}
\caption{\textbf{Results under our own metric ($g_{\text{margin}}$).} \textrm{[Left]} By backbone.
\textrm{[Right]} By benchmark. Points are medians and bars are bootstrap 95\% confidence
intervals, and a row right of zero means the step is explained better by its own row than by
another step's row.}
\label{fig:forest}
\end{figure}

\noindent\textbf{Results under our own metric.}
To validate that the horizon axis holds beyond the synthetic setting, we aggregate every
real-world run under the protocol of Section~\ref{sec:protocol}.
As shown in Figure~\ref{fig:forest}, $g_{\text{margin}}$ is consistently positive across
backbones and benchmarks, indicating that \textit{each step is explained better by its own row}
than by the row of another step.
The gain $g_{\text{own}}$ is consistently positive while $g_{\text{shuffled}}$ falls below
zero, which means \textit{a row handed to the wrong step actively hurts}, and
Appendix~\ref{sec:detailed} has both.
All quantities are normalized by the forecast variance, which never changes a run's sign.

\begin{wraptable}{r}{0.552\linewidth}
\vspace{-\intextsep}
\vspace{-10.6pt}
\caption{\textbf{Results under metrics of prior works.} Every row compares a step against the
shared vector.}
\label{tab:stdmetrics}
\begin{center}
\small
\begin{fittable}\begin{tabular}{l|cc|c}
\toprule
\textbf{Metric} & \textbf{Median} & \textbf{Mean} & \textbf{Positive} \\
\midrule
\multicolumn{4}{l}{\textbf{Metrics of prior works}} \\
Comprehensiveness & \textcolor{hred}{$\mathbf{+.0038}$} & \textcolor{hred}{$\mathbf{+.0305}$} & \textcolor{hred}{\textbf{69.6\%}} \\
AOPC & \textcolor{hred}{$\mathbf{+.0030}$} & \textcolor{hred}{$\mathbf{+.0271}$} & \textcolor{hred}{\textbf{68.7\%}} \\
Sufficiency & \textcolor{hred}{$\mathbf{+.0022}$} & \textcolor{hred}{$\mathbf{+.0114}$} & \textcolor{hred}{\textbf{72.2\%}} \\
Insertion & \textcolor{hred}{$\mathbf{+.0320}$} & \textcolor{hred}{$\mathbf{+.0475}$} & \textcolor{hred}{\textbf{73.9\%}} \\
\midrule
\multicolumn{4}{l}{\textbf{Metrics of ours}} \\
$g_{\text{own}}$ & \textcolor{hred}{$\mathbf{+.0888}$} & \textcolor{hred}{$\mathbf{+.1365}$} & \textcolor{hred}{\textbf{70.3\%}} \\
$g_{\text{shuffled}}$ & \textcolor{hblue}{$\mathbf{-.0530}$} & \textcolor{hblue}{$\mathbf{-.1011}$} & \textcolor{hblue}{\textbf{34.6\%}} \\
$g_{\text{margin}}$ & \textcolor{hred}{$\mathbf{+.1418}$} & \textcolor{hred}{$\mathbf{+.2376}$} & \textcolor{hred}{\textbf{80.6\%}} \\
\bottomrule
\end{tabular}
\end{fittable}
\end{center}
\vspace{-11pt}
\end{wraptable}

\noindent\textbf{Results under metrics of prior works.}
\label{sec:stdmetrics}
To validate the horizon axis under the metrics of prior works, which were defined for a model
with a single output (e.g., a class label), we re-express four of them for forecasting
and score the explanation of step $h$ against the shared vector.
Table~\ref{tab:stdmetrics} reports the median and the trimmed mean of that comparison for each
metric, and all four metrics of prior works are positive under a sign test over \nStdRuns{} runs.
The results indicate that \textit{the horizon axis helps regardless of how faithfulness is
measured}.
The four metrics are re-expressed as follows, with details discussed in
Appendix~\ref{sec:stddefs}.
{\renewcommand{\bodyitemsep}{0pt}
\begin{bodyitemize}
\item Comprehensiveness \citep{deyoung2020eraser} deletes the highest-ranked positions.
\item AOPC \citep{samek2016evaluating} deletes them and averages over how many are deleted.
\item Sufficiency \citep{deyoung2020eraser} keeps the highest-ranked positions instead.
\item Insertion \citep{petsiuk2018rise} keeps them and reads how far the forecast recovers.
\end{bodyitemize}}

\subsection{Application to Other Explanation Methods}
\label{sec:methods}

\noindent\textbf{Setup.}
To validate that the horizon axis is what improves the explanation rather than any particular
estimator, we apply the axis to the estimators of seven other explanation
methods\footnote{Methods published for classification need one change, since the quantity a mask
or a gradient is asked to preserve becomes the forecast at step $h$ instead of a class score.
Details are provided in Appendix~\ref{sec:porting}.} and score them under the identical
protocol, on the synthetic data of Section~\ref{sec:sanity} and on the real-world benchmarks.

\noindent\textbf{Results on synthetic datasets.}
The synthetic data admits the ground-truth metrics, since the answer is known in advance.
Each metric is read twice, once with the shared vector of prior works and once with our matrix,
and each cell averages \gtmModels{} backbones and \gtmGens{} generators.
As shown in Table~\ref{tab:gt_body}, resolving the axis wins in almost every cell, indicating
that the gain is robust across estimators.

\begin{table}[t]
\caption{Application of HRX to various estimators (synthetic data).}
\label{tab:gt_body}
\begin{center}
\small
\begin{fittable}\begin{tabular}{l|cc|cc|cc|cc}
\toprule
\multirow{2.5}{*}{\textbf{Estimator}} & \multicolumn{2}{c|}{\textbf{AUPRC}} & \multicolumn{2}{c|}{\textbf{AUROC}} & \multicolumn{2}{c|}{\textbf{AUP}} & \multicolumn{2}{c}{\textbf{AUR}} \\
\cmidrule(lr){2-3}\cmidrule(lr){4-5}\cmidrule(lr){6-7}\cmidrule(lr){8-9}
 & \textbf{Vector} & \textbf{Matrix} & \textbf{Vector} & \textbf{Matrix} & \textbf{Vector} & \textbf{Matrix} & \textbf{Vector} & \textbf{Matrix} \\
\midrule
Saliency & $0.551$ & \textcolor{hred}{$0.971$} & $0.954$ & \textcolor{hred}{$0.994$} & $0.130$ & \textcolor{hred}{$0.151$} & $0.950$ & \textcolor{hred}{$0.974$} \\
Gradient $\times$ input & $0.547$ & \textcolor{hred}{$0.968$} & $0.952$ & \textcolor{hred}{$0.994$} & $0.130$ & \textcolor{hred}{$0.151$} & $0.947$ & \textcolor{hred}{$0.974$} \\
Integrated gradients & $0.549$ & \textcolor{hred}{$0.968$} & $0.953$ & \textcolor{hred}{$0.994$} & $0.130$ & \textcolor{hred}{$0.151$} & $0.947$ & \textcolor{hred}{$0.974$} \\
Occlusion & $0.125$ & \textcolor{hred}{$0.150$} & $0.688$ & \textcolor{hred}{$0.838$} & $0.068$ & \textcolor{hred}{$0.084$} & $0.694$ & \textcolor{hred}{$0.822$} \\
Dynamask & $0.455$ & \textcolor{hred}{$0.756$} & $0.886$ & \textcolor{hred}{$0.947$} & $0.115$ & \textcolor{hred}{$0.134$} & $0.882$ & \textcolor{hred}{$0.925$} \\
ExtremalMask & $0.460$ & \textcolor{hred}{$0.705$} & $0.853$ & \textcolor{hred}{$0.903$} & $0.106$ & \textcolor{hred}{$0.118$} & $0.853$ & $0.847$ \\
ContraLSP & $0.478$ & \textcolor{hred}{$0.847$} & $0.888$ & \textcolor{hred}{$0.948$} & $0.116$ & \textcolor{hred}{$0.123$} & $0.886$ & \textcolor{hred}{$0.894$} \\
Input gradient & $0.550$ & \textcolor{hred}{$0.973$} & $0.954$ & \textcolor{hred}{$0.995$} & $0.130$ & \textcolor{hred}{$0.152$} & $0.950$ & \textcolor{hred}{$0.974$} \\
\bottomrule
\end{tabular}
\end{fittable}
\end{center}
\vspace{-3pt}
\end{table}

\begin{table}[t]
\caption{Application of HRX to various estimators (real-world data).}
\label{tab:methods}
\begin{center}
\small
{\setlength{\tabcolsep}{4pt}
\begin{fittable}\begin{tabular}{l|cc|cc|cc}
\toprule
\multirow{2.5}{*}{\textbf{Estimator}} & \multicolumn{2}{c|}{\textbf{$g_{\text{own}}$}} & \multicolumn{2}{c|}{\textbf{$g_{\text{shuffled}}$}} & \multicolumn{2}{c}{\textbf{$g_{\text{margin}}$}} \\
\cmidrule(lr){2-3}\cmidrule(lr){4-5}\cmidrule(lr){6-7}
 & \textbf{Median} & \textbf{95\% CI} & \textbf{Median} & \textbf{95\% CI} & \textbf{Median} & \textbf{95\% CI} \\
\midrule
Saliency & \textcolor{hred}{$0.098$} & $[0.090, 0.108]$ & \textcolor{hblue}{$-0.167$} & $[-0.179, -0.154]$ & \textcolor{hred}{$0.265$} & $[0.250, 0.280]$ \\
Gradient $\times$ input & \textcolor{hred}{$0.086$} & $[0.079, 0.095]$ & \textcolor{hblue}{$-0.142$} & $[-0.150, -0.132]$ & \textcolor{hred}{$0.228$} & $[0.215, 0.239]$ \\
Integrated gradients & \textcolor{hred}{$0.077$} & $[0.071, 0.085]$ & \textcolor{hblue}{$-0.153$} & $[-0.161, -0.145]$ & \textcolor{hred}{$0.230$} & $[0.221, 0.242]$ \\
Occlusion & \textcolor{hred}{$0.029$} & $[0.025, 0.036]$ & \textcolor{hblue}{$-0.044$} & $[-0.049, -0.039]$ & \textcolor{hred}{$0.073$} & $[0.067, 0.080]$ \\
Dynamask & \textcolor{hred}{$0.275$} & $[0.254, 0.292]$ & \textcolor{hblue}{$-0.362$} & $[-0.380, -0.344]$ & \textcolor{hred}{$0.637$} & $[0.606, 0.666]$ \\
ExtremalMask & \textcolor{hred}{$0.168$} & $[0.152, 0.187]$ & \textcolor{hblue}{$-0.193$} & $[-0.208, -0.180]$ & \textcolor{hred}{$0.361$} & $[0.338, 0.387]$ \\
ContraLSP & \textcolor{hred}{$0.158$} & $[0.142, 0.177]$ & \textcolor{hblue}{$-0.229$} & $[-0.242, -0.218]$ & \textcolor{hred}{$0.387$} & $[0.364, 0.413]$ \\
Input gradient & \textcolor{hred}{$0.141$} & $[0.129, 0.153]$ & \textcolor{hblue}{$-0.130$} & $[-0.140, -0.117]$ & \textcolor{hred}{$0.271$} & $[0.252, 0.287]$ \\
\bottomrule
\end{tabular}
\end{fittable}}
\end{center}
\vspace{-3pt}
\end{table}

\ifpairtables
\begin{table}[t]
\begin{minipage}[t]{0.50\linewidth}
\caption{\textbf{Gain $g_{\text{own}}$ by domain.} Rows split by whether the model beats both
naive baselines.}
\label{tab:domains}
\begin{center}
\small
\begin{fittable}\begin{tabular}{lcc}
\toprule
\textbf{Domain} & \textbf{$g_{\text{own}}$} & \textbf{95\% CI} \\
\midrule
\textbf{\textit{General}} & \textcolor{hred}{$\mathbf{+0.133}$} & $[+0.129, +0.136]$ \\
\quad Beats naive & \textcolor{hred}{$\mathbf{+0.155}$} & $[+0.151, +0.159]$ \\
\quad Loses to naive & \textcolor{hred}{$\mathbf{+0.060}$} & $[+0.055, +0.064]$ \\
\midrule
\textbf{\textit{Financial}} & \textcolor{hred}{$\mathbf{+0.211}$} & $[+0.199, +0.220]$ \\
\quad Log price & \textcolor{hred}{$\mathbf{+0.061}$} & $[+0.048, +0.072]$ \\
\qquad Beats naive & \textcolor{hred}{$\mathbf{+0.073}$} & $[+0.056, +0.089]$ \\
\qquad Loses to naive & \textcolor{hred}{$\mathbf{+0.055}$} & $[+0.039, +0.067]$ \\
\quad Log return & \textcolor{hred}{$\mathbf{+0.340}$} & $[+0.325, +0.356]$ \\
\qquad Beats naive & \textcolor{hred}{$\mathbf{+0.548}$} & $[+0.390, +0.695]$ \\
\qquad Loses to naive & \textcolor{hred}{$\mathbf{+0.336}$} & $[+0.322, +0.352]$ \\
\bottomrule
\end{tabular}
\end{fittable}
\end{center}
\end{minipage}
\end{table}
\fi

\noindent\textbf{Results on real-world datasets.}
Note that a method without the horizon axis uses one vector for every step, which makes all
three quantities \textit{exactly zero}, and every gain here is what the axis adds.
As shown in Table~\ref{tab:methods}, the gain is significantly positive for every estimator,
indicating that \textit{the improvement belongs to the horizon axis rather than to any
particular estimator}.
Every interval spans \cmpModels{} backbones, \cmpDatasets{} benchmarks, \cmpShapes{} window
shapes and \cmpSeeds{} random seeds.

\section{Analysis}

\leavevmode\par\vspace{-\baselineskip}

\ifpairtables\else
\begin{wraptable}{r}{0.52\linewidth}
\vspace{-\intextsep}
\vspace{-10.4pt}
\caption{\textbf{Gain $g_{\text{own}}$ by domain.} Each domain is split by whether the model
beats both naive baselines.}
\label{tab:domains}
\vspace{-4pt}
\begin{center}
\footnotesize\renewcommand{\arraystretch}{0.75}
\begin{fittable}\begin{tabular}{lcc}
\toprule
\textbf{Domain} & \textbf{$g_{\text{own}}$} & \textbf{95\% CI} \\
\midrule
\textbf{\textit{General}} & \textcolor{hred}{$\mathbf{+0.133}$} & $[+0.129, +0.136]$ \\
\quad Beats naive & \textcolor{hred}{$\mathbf{+0.155}$} & $[+0.151, +0.159]$ \\
\quad Loses to naive & \textcolor{hred}{$\mathbf{+0.060}$} & $[+0.055, +0.064]$ \\
\midrule
\textbf{\textit{Financial}} & \textcolor{hred}{$\mathbf{+0.211}$} & $[+0.199, +0.220]$ \\
\quad Log price & \textcolor{hred}{$\mathbf{+0.061}$} & $[+0.048, +0.072]$ \\
\qquad Beats naive & \textcolor{hred}{$\mathbf{+0.073}$} & $[+0.056, +0.089]$ \\
\qquad Loses to naive & \textcolor{hred}{$\mathbf{+0.055}$} & $[+0.039, +0.067]$ \\
\quad Log return & \textcolor{hred}{$\mathbf{+0.340}$} & $[+0.325, +0.356]$ \\
\qquad Beats naive & \textcolor{hred}{$\mathbf{+0.548}$} & $[+0.390, +0.695]$ \\
\qquad Loses to naive & \textcolor{hred}{$\mathbf{+0.336}$} & $[+0.322, +0.352]$ \\
\bottomrule
\end{tabular}
\end{fittable}
\end{center}
\vspace{-11pt}
\end{wraptable}
\fi

\noindent\textbf{Two domains: General \& Finance.}
\label{sec:domains}
To analyze the effect by the domain of a dataset, we report the general suite and the financial
series apart, and split each by whether the model beats two naive baselines, which are
predicting zero and repeating the last value.
As shown in Table~\ref{tab:domains}, runs that beat the baselines have a larger
$g_{\text{own}}$ than runs that do not, in all three groups, indicating that \textit{the
explanation gain grows with the forecasting skill of the model}.
Note that every group is positive, which means forecasting skill sets the size of the gain
rather than whether one exists.

\begin{figure}[t]
\centering
\includegraphics[width=\linewidth]{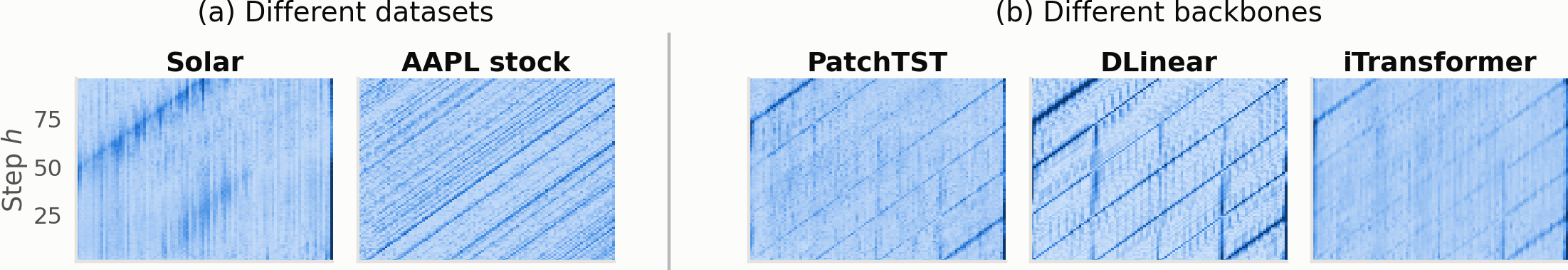}
\vspace{-6pt}
\caption{\textbf{Visualization of explanation matrices.} Rows are forecast steps and columns are
lookback positions, and the band tracks the step in (a) two datasets and (b) three backbones.}
\label{fig:hero}
\end{figure}

\noindent\textbf{Visualization of explanation matrices.}
To see what the matrix looks like in use, we vary the dataset in one group and the backbone in
the other.
As shown in Figure~\ref{fig:hero}(a), a clear band appears on both the Solar series (for
general domain) and the AAPL stock return (for financial domain).
Furthermore, the three backbones of Figure~\ref{fig:hero}(b) carry the same diagonal band on one
benchmark, and they differ only in how many parallel bands they hold.
In every panel the important lookback positions shift as the forecast step advances, which a
single shared vector cannot express.
Appendix~\ref{sec:qual} visualizes all the panels at full size, with the naive-baseline verdict
and the effective lookback of every row.

\begin{wraptable}{r}{0.372\linewidth}
\vspace{-\intextsep}
\vspace{-10.6pt}
\caption{\textbf{Horizon axis lives on time axis.} Effective rank of $E$ is
the participation ratio of its singular values.}
\label{tab:timechan}
\begin{center}
\small
\begin{tabular}{lcc}
\toprule
\textbf{Axis} & \textbf{Time} & \textbf{Channel} \\
\midrule
Gain $g_{\text{own}}$ & $\timeGain$ & $\chanGain$ \\
Effective rank & $\timeRank$ & $\chanRank$ \\
\bottomrule
\end{tabular}
\end{center}
\vspace{-11pt}
\end{wraptable}

\noindent\textbf{Time vs. channel axis.}
\label{sec:timechan}
To ask whether the same dissociation appears across channels, we run the identical protocol on
both input axes of multivariate forecasters.
As shown in Table~\ref{tab:timechan}, \textit{the time axis gives a clear gain} while
\textit{the channel axis gives almost none}, and its effective rank sits near $1$, which means
the channel rows are copies of one vector.
In other words, a forecaster changes \textit{when} it looks as the forecast step moves, but not
\textit{which} variable it looks at.

\begin{table}[t]
\begin{minipage}[t]{0.40\linewidth}
\caption{\textbf{Rank of $E$.} Each truncation is scored under the protocol of
Section~\ref{sec:protocol}, where rank $1$ is pinned to zero.}
\label{tab:lowrank}
\vspace{-6pt}
\begin{center}
\small
\begin{fittable}\begin{tabular}{l|cc}
\toprule
\textbf{Truncation} & \textbf{$g_{\text{own}}$} & \textbf{Fraction of full} \\
\midrule
$r = 1$ & $+0.000$ & 0.00 \\
$r = 2$ & \textcolor{hred}{$\mathbf{+0.044}$} & 0.19 \\
$r = 4$ & \textcolor{hred}{$\mathbf{+0.110}$} & 0.51 \\
$r = 8$ & \textcolor{hred}{$\mathbf{+0.153}$} & 0.71 \\
\midrule
Full ($r = H$) & \textcolor{hred}{$\mathbf{+0.228}$} & 1.00 \\
\bottomrule
\end{tabular}
\end{fittable}
\end{center}
\end{minipage}\hfill
\begin{minipage}[t]{0.57\linewidth}
\caption{\textbf{Rank predicts the gain.} Runs grouped by the effective rank of $E$
($r(E)$), where a rank near one means the shared vector of prior works loses nothing.}
\label{tab:rankgain}
\vspace{-6pt}
\begin{center}
\small
\begin{fittable}\begin{tabular}{l|cccc}
\toprule
\textbf{$r(E)$} & \textbf{Runs} & \textbf{$g_{\text{own}}$} & \textbf{95\% CI} & \textbf{Positive} \\
\midrule
$[1.0, 1.1)$ & 2{,}103 & \textcolor{hred}{$\mathbf{+0.015}$} & $[+0.012, +0.018]$ & \textcolor{hred}{\textbf{62\%}} \\
$[1.1, 1.25)$ & 3{,}977 & \textcolor{hred}{$\mathbf{+0.074}$} & $[+0.069, +0.079]$ & \textcolor{hred}{\textbf{71\%}} \\
$[1.25, 1.5)$ & 5{,}492 & \textcolor{hred}{$\mathbf{+0.109}$} & $[+0.102, +0.119]$ & \textcolor{hred}{\textbf{75\%}} \\
$[1.5, 2.0)$ & 6{,}675 & \textcolor{hred}{$\mathbf{+0.146}$} & $[+0.138, +0.150]$ & \textcolor{hred}{\textbf{83\%}} \\
$[2.0, 3.0)$ & 4{,}751 & \textcolor{hred}{$\mathbf{+0.250}$} & $[+0.244, +0.258]$ & \textcolor{hred}{\textbf{91\%}} \\
$[3.0, \infty)$ & 2{,}208 & \textcolor{hred}{$\mathbf{+0.323}$} & $[+0.316, +0.329]$ & \textcolor{hred}{\textbf{98\%}} \\
\bottomrule
\end{tabular}
\end{fittable}
\end{center}
\end{minipage}
\end{table}

\noindent\textbf{How many maps a forecast needs.}
\label{sec:lowrank}
To ask how many maps a forecast needs, we truncate $E$ to its leading directions and score every
truncation under the identical protocol.
As shown in Table~\ref{tab:lowrank}, a single direction recovers none of the gain, which is what
Section~\ref{sec:rank} predicts, while a handful of further directions recover most of it.
Table~\ref{tab:rankgain} reads the same structure the other way, where the gain rises with the
rank and the runs closest to one direction are the ones where the shared vector loses almost
nothing.
Appendix~\ref{sec:rankdetail} and Appendix~\ref{sec:offsetdetail} give the details of the rank
measure and the distance.

\noindent\textbf{Other analysis.}
The appendix reports three further analyses, and we summarize each one here.
{\setlength{\topsep}{-\parskip}\setlength{\partopsep}{0pt}
\begin{itemize}
\setlength{\itemsep}{0pt}\setlength{\parsep}{0pt}\setlength{\parskip}{0pt}
\item \textbf{Application to zero-shot TSFMs.} The gain is positive for almost every
foundation model run zero-shot, so the axis holds even when the model has never seen the
series (Appendix~\ref{sec:fmdetail}).
\item \textbf{Forecast length, lookback and depth.} The gain rises with both the forecast length
and the lookback length,
while depth moves in the opposite direction by far less (Appendix~\ref{sec:axes}).
\item \textbf{Cost of each estimator.} The mask methods cost two orders of magnitude more than a
gradient, since the axis makes each solve one problem per forecast step
(Appendix~\ref{sec:cost}).
\end{itemize}}

\section{Conclusion}
\label{sec:conclusion}

We showed that TS forecasters read \textit{different parts of the lookback window for different
forecast steps}, and we proposed horizon-resolved explanation, which recovers it as a matrix
without modifying the TS backbone, together with a protocol that isolates the horizon axis.
Two limitations remain, as no statistic of the series predicts the effect size, and none of the
downstream uses we tried gave a better forecast, which means HRX is a \textit{measurement tool}
for now.
Future work is to find what makes a series reward the axis, and to turn the matrix into a signal
a forecaster can use at training time.

\subsection*{AI use statement}

We used a generative AI tool to proofread this manuscript, where it checked grammar, wording,
and typographical errors in text we had already written, and we received minor coding assistance
from the same tool while implementing the experiment code.
The research question, the framing of the horizon axis as our contribution, the design of the
evaluation protocol, and the analysis of the results are our own.
Every number reported in this paper comes from an experiment we executed, and no result,
citation, or reference was produced by a generative model.
We reviewed every edit and every line of AI-assisted code we accepted, and we take
responsibility for the final content of this work.

\subsection*{Ethics statement}

This work studies how a trained forecaster uses its input, so it produces diagnostic
information about models rather than new predictive capability.
The general-domain benchmarks are standard public forecasting datasets, and they contain
aggregate measurements from sensors and meters rather than records about individuals.
The financial data are daily closing prices of publicly traded companies, retrieved from a
public market data source, and they contain no personal or proprietary information.
No human subjects were involved and no personally identifiable data were used at any stage.
One risk follows from the nature of explanation methods, which is that an explanation map
invites a causal reading it does not support.
Our protocol addresses this directly, as it compares a perturbed forecast against the model's
own unperturbed forecast and therefore measures what the model uses rather than what drives the
underlying process.
Furthermore, we report a forecasting baseline beside every result, so that an explanation of a
model with no forecasting skill is not read as evidence about the world.

\subsection*{Reproducibility statement}

Appendix~\ref{sec:setup} gives the backbones, benchmarks, swept configurations, and training
protocol, and Appendix~\ref{sec:morf} gives the deletion curve and the exact form of every
metric we report.
Appendix~\ref{sec:fmdetail} gives the checkpoint, context length, and verification procedure for
each pretrained forecaster.
Appendix~\ref{sec:seeds} reports the seed-to-seed variation of our main quantity,
Appendix~\ref{sec:gt} reports the same conclusion against constructed ground truth, and
Section~\ref{sec:stdmetrics} reports it under metrics of prior works.
Note that every number quoted in the text is generated from the stored per-run records by a
script rather than transcribed by hand, so the text and the tables cannot drift apart.
We submit the full source code as supplementary material, which includes the estimators, the
evaluation protocol, the data preparation scripts, and the scripts that produce every figure
and table in this paper.

\bibliography{refs}

@inproceedings{crabbe2021dynamask,
  title={Explaining time series predictions with dynamic masks},
  author={Crabb{\'e}, Jonathan and van der Schaar, Mihaela},
  booktitle={International Conference on Machine Learning},
  year={2021}
}

@inproceedings{enguehard2023extremal,
  title={Learning perturbations to explain time series predictions},
  author={Enguehard, Joseph},
  booktitle={International Conference on Machine Learning},
  year={2023}
}

@inproceedings{queen2023timex,
  title={Encoding time-series explanations through self-supervised model behavior consistency},
  author={Queen, Owen and Hartvigsen, Thomas and Koker, Teddy and He, Huan and Tsiligkaridis, Theodoros and Zitnik, Marinka},
  booktitle={Advances in Neural Information Processing Systems},
  year={2023}
}

@inproceedings{liu2024contralsp,
  title={Explaining time series via contrastive and locally sparse perturbations},
  author={Liu, Zichuan and Zhang, Yingying and Wang, Tianchun and Wang, Zefan and Luo, Dongsheng and Du, Mengnan and Wu, Min and Wang, Yi and Chen, Chunlin and Fan, Lunting and Wen, Qingsong},
  booktitle={International Conference on Learning Representations},
  year={2024}
}

@inproceedings{liu2024timexpp,
  title={{TimeX++}: Learning time-series explanations with information bottleneck},
  author={Liu, Zichuan and Wang, Tianchun and Shi, Jimeng and Zheng, Xu and Chen, Zhuomin and Song, Lei and Dong, Wenqian and Obeysekera, Jayantha and Shirani, Farhad and Luo, Dongsheng},
  booktitle={International Conference on Machine Learning},
  year={2024}
}

@inproceedings{ismail2020tsr,
  title={Benchmarking deep learning interpretability in time series predictions},
  author={Ismail, Aya Abdelsalam and Gunady, Mohamed and Corrada Bravo, H{\'e}ctor and Feizi, Soheil},
  booktitle={Advances in Neural Information Processing Systems},
  year={2020}
}

@inproceedings{tonekaboni2020fit,
  title={What went wrong and when? {Instance-wise} feature importance for time-series black-box models},
  author={Tonekaboni, Sana and Joshi, Shalmali and Campbell, Kieran and Duvenaud, David and Goldenberg, Anna},
  booktitle={Advances in Neural Information Processing Systems},
  year={2020}
}

@inproceedings{leung2023winit,
  title={Temporal dependencies in feature importance for time series prediction},
  author={Leung, Kin Kwan and Rooke, Clayton and Smith, Jonathan and Zuberi, Saba and Volkovs, Maksims},
  booktitle={International Conference on Learning Representations},
  year={2023}
}

@inproceedings{bento2021timeshap,
  title={{TimeSHAP}: Explaining recurrent models through sequence perturbations},
  author={Bento, Jo{\~a}o and Saleiro, Pedro and Cruz, Andr{\'e} F. and Figueiredo, M{\'a}rio A. T. and Bizarro, Pedro},
  booktitle={ACM SIGKDD Conference on Knowledge Discovery and Data Mining},
  year={2021}
}

@inproceedings{lundberg2017shap,
  title={A unified approach to interpreting model predictions},
  author={Lundberg, Scott M. and Lee, Su-In},
  booktitle={Advances in Neural Information Processing Systems},
  year={2017}
}

@inproceedings{sundararajan2017ig,
  title={Axiomatic attribution for deep networks},
  author={Sundararajan, Mukund and Taly, Ankur and Yan, Qiqi},
  booktitle={International Conference on Machine Learning},
  year={2017}
}

@article{simonyan2013saliency,
  title={Deep inside convolutional networks: Visualising image classification models and saliency maps},
  author={Simonyan, Karen and Vedaldi, Andrea and Zisserman, Andrew},
  journal={arXiv preprint arXiv:1312.6034},
  year={2013}
}

@article{smilkov2017smoothgrad,
  title={{SmoothGrad}: Removing noise by adding noise},
  author={Smilkov, Daniel and Thorat, Nikhil and Kim, Been and Vi{\'e}gas, Fernanda and Wattenberg, Martin},
  journal={arXiv preprint arXiv:1706.03825},
  year={2017}
}

@article{samek2016evaluating,
  title={Evaluating the visualization of what a deep neural network has learned},
  author={Samek, Wojciech and Binder, Alexander and Montavon, Gr{\'e}goire and Lapuschkin, Sebastian and M{\"u}ller, Klaus-Robert},
  journal={IEEE Transactions on Neural Networks and Learning Systems},
  year={2017}
}

@article{brocki2022evaluation,
  title={Fidelity of interpretability methods and perturbation artifacts in neural networks},
  author={Brocki, Lennart and Chung, Neo Christopher},
  journal={arXiv preprint arXiv:2203.02928},
  year={2022}
}

@inproceedings{adebayo2018sanity,
  title={Sanity checks for saliency maps},
  author={Adebayo, Julius and Gilmer, Justin and Muelly, Michael and Goodfellow, Ian and Hardt, Moritz and Kim, Been},
  booktitle={Advances in Neural Information Processing Systems},
  year={2018}
}

@inproceedings{hooker2019roar,
  title={A benchmark for interpretability methods in deep neural networks},
  author={Hooker, Sara and Erhan, Dumitru and Kindermans, Pieter-Jan and Kim, Been},
  booktitle={Advances in Neural Information Processing Systems},
  year={2019}
}

@inproceedings{zeng2023dlinear,
  title={Are {Transformers} effective for time series forecasting?},
  author={Zeng, Ailing and Chen, Muxi and Zhang, Lei and Xu, Qiang},
  booktitle={AAAI Conference on Artificial Intelligence},
  year={2023}
}

@inproceedings{nie2023patchtst,
  title={A time series is worth 64 words: Long-term forecasting with {Transformers}},
  author={Nie, Yuqi and Nguyen, Nam H. and Sinthong, Phanwadee and Kalagnanam, Jayant},
  booktitle={International Conference on Learning Representations},
  year={2023}
}

@inproceedings{liu2024itransformer,
  title={{iTransformer}: Inverted {Transformers} are effective for time series forecasting},
  author={Liu, Yong and Hu, Tengge and Zhang, Haoran and Wu, Haixu and Wang, Shiyu and Ma, Lintao and Long, Mingsheng},
  booktitle={International Conference on Learning Representations},
  year={2024}
}

@inproceedings{wu2023timesnet,
  title={{TimesNet}: Temporal {2D}-variation modeling for general time series analysis},
  author={Wu, Haixu and Hu, Tengge and Liu, Yong and Zhou, Hang and Wang, Jianmin and Long, Mingsheng},
  booktitle={International Conference on Learning Representations},
  year={2023}
}

@inproceedings{wu2021autoformer,
  title={Autoformer: Decomposition {Transformers} with auto-correlation for long-term series forecasting},
  author={Wu, Haixu and Xu, Jiehui and Wang, Jianmin and Long, Mingsheng},
  booktitle={Advances in Neural Information Processing Systems},
  year={2021}
}

@inproceedings{zhou2021informer,
  title={Informer: Beyond efficient {Transformer} for long sequence time-series forecasting},
  author={Zhou, Haoyi and Zhang, Shanghang and Peng, Jieqi and Zhang, Shuai and Li, Jianxin and Xiong, Hui and Zhang, Wancai},
  booktitle={AAAI Conference on Artificial Intelligence},
  year={2021}
}

@inproceedings{zhou2022fedformer,
  title={{FEDformer}: Frequency enhanced decomposed {Transformer} for long-term series forecasting},
  author={Zhou, Tian and Ma, Ziqing and Wen, Qingsong and Wang, Xue and Sun, Liang and Jin, Rong},
  booktitle={International Conference on Machine Learning},
  year={2022}
}

@inproceedings{wang2024timemixer,
  title={{TimeMixer}: Decomposable multiscale mixing for time series forecasting},
  author={Wang, Shiyu and Wu, Haixu and Shi, Xiaoming and Hu, Tengge and Luo, Huakun and Ma, Lintao and Zhang, James Y. and Zhou, Jun},
  booktitle={International Conference on Learning Representations},
  year={2024}
}

@article{lim2021tft,
  title={{Temporal Fusion Transformers} for interpretable multi-horizon time series forecasting},
  author={Lim, Bryan and Ar{\i}k, Sercan {\"O}. and Loeff, Nicolas and Pfister, Tomas},
  journal={International Journal of Forecasting},
  year={2021}
}

@inproceedings{oreshkin2020nbeats,
  title={{N-BEATS}: Neural basis expansion analysis for interpretable time series forecasting},
  author={Oreshkin, Boris N. and Carpov, Dmitri and Chapados, Nicolas and Bengio, Yoshua},
  booktitle={International Conference on Learning Representations},
  year={2020}
}

@article{wang2024tslib,
  title={Deep time series models: A comprehensive survey and benchmark},
  author={Wang, Yuxuan and Wu, Haixu and Dong, Jiaxiang and Liu, Yong and Wang, Chen and Long, Mingsheng and Wang, Jianmin},
  journal={arXiv preprint arXiv:2407.13278},
  year={2024}
}

@inproceedings{deyoung2020eraser,
  title={{ERASER}: A benchmark to evaluate rationalized {NLP} models},
  author={DeYoung, Jay and Jain, Sarthak and Rajani, Nazneen Fatema and Lehman, Eric and Xiong, Caiming and Socher, Richard and Wallace, Byron C.},
  booktitle={Annual Meeting of the Association for Computational Linguistics},
  year={2020}
}

@inproceedings{petsiuk2018rise,
  title={{RISE}: Randomized input sampling for explanation of black-box models},
  author={Petsiuk, Vitali and Das, Abir and Saenko, Kate},
  booktitle={British Machine Vision Conference},
  year={2018}
}

@article{ansari2024chronos,
  title={Chronos: Learning the language of time series},
  author={Ansari, Abdul Fatir and Stella, Lorenzo and Turkmen, Caner and Zhang, Xiyuan and Mercado, Pedro and Shen, Huibin and Shchur, Oleksandr and Rangapuram, Syama Sundar and Pineda Arango, Sebastian and Kapoor, Shubham and Zschiegner, Jasper and Maddix, Danielle C. and Wang, Hao and Mahoney, Michael W. and Torkkola, Kari and Gordon Wilson, Andrew and Bohlke-Schneider, Michael and Wang, Yuyang},
  journal={Transactions on Machine Learning Research},
  year={2024}
}

@inproceedings{das2024timesfm,
  title={A decoder-only foundation model for time-series forecasting},
  author={Das, Abhimanyu and Kong, Weihao and Sen, Rajat and Zhou, Yichen},
  booktitle={International Conference on Machine Learning},
  year={2024}
}

@inproceedings{woo2024moirai,
  title={Unified training of universal time series forecasting {Transformers}},
  author={Woo, Gerald and Liu, Chenghao and Kumar, Akshat and Xiong, Caiming and Savarese, Silvio and Sahoo, Doyen},
  booktitle={International Conference on Machine Learning},
  year={2024}
}

@article{aksu2024gifteval,
  title={{GIFT-Eval}: A benchmark for general time series forecasting model evaluation},
  author={Aksu, Taha and Woo, Gerald and Liu, Juncheng and Liu, Xu and Liu, Chenghao and Savarese, Silvio and Xiong, Caiming and Sahoo, Doyen},
  journal={arXiv preprint arXiv:2410.10393},
  year={2024}
}

@inproceedings{liu2024timer,
  title={Timer: Generative pre-trained {Transformers} are large time series models},
  author={Liu, Yong and Zhang, Haoran and Li, Chenyu and Huang, Xiangdong and Wang, Jianmin and Long, Mingsheng},
  booktitle={International Conference on Machine Learning},
  year={2024}
}

@inproceedings{shi2025timemoe,
  title={{Time-MoE}: Billion-scale time series foundation models with mixture of experts},
  author={Shi, Xiaoming and Wang, Shiyu and Nie, Yuqi and Li, Dianqi and Ye, Zhou and Wen, Qingsong and Jin, Ming},
  booktitle={International Conference on Learning Representations},
  year={2025}
}

@inproceedings{ekambaram2024ttm,
  title={Tiny time mixers ({TTMs}): Fast pre-trained models for enhanced zero/few-shot forecasting of multivariate time series},
  author={Ekambaram, Vijay and Jati, Arindam and Dayama, Pankaj and Mukherjee, Sumanta and Nguyen, Nam H. and Gifford, Wesley M. and Reddy, Chandra and Kalagnanam, Jayant},
  booktitle={Advances in Neural Information Processing Systems},
  year={2024}
}

@inproceedings{chen2024visionts,
  title={{VisionTS}: Visual masked autoencoders are free-lunch zero-shot time series forecasters},
  author={Chen, Mouxiang and Shen, Lefei and Li, Zhuo and Wang, Xiaoyun Joy and Sun, Jianling and Liu, Chenghao},
  booktitle={International Conference on Machine Learning},
  year={2025}
}

@inproceedings{goswami2024moment,
  title={{MOMENT}: A family of open time-series foundation models},
  author={Goswami, Mononito and Szafer, Konrad and Choudhry, Arjun and Cai, Yifu and Li, Shuo and Dubrawski, Artur},
  booktitle={International Conference on Machine Learning},
  year={2024}
}

@inproceedings{liu2025sundial,
  title={Sundial: A family of highly capable time series foundation models},
  author={Liu, Yong and Qin, Guo and Shi, Zhiyuan and Chen, Zhi and Yang, Caiyin and Huang, Xiangdong and Wang, Jianmin and Long, Mingsheng},
  booktitle={International Conference on Machine Learning},
  year={2025}
}

@article{lee2026exaone,
  title={{EXAONE Finance 1.0}: An attention-free time series foundation model for financial time series},
  author={Lee, Seunghan and Lee, Jaehoon and Seo, Jun and Lim, Tae Yoon and Kang, Dongwan and Choi, Hwanil and Kim, Minjae and Yoo, Sungdong and Kang, Junhyeok and Han, Sangjun and others},
  journal={arXiv preprint arXiv:2609.04239},
  year={2026}
}

@inproceedings{chung2024spectralx,
  title={Time is Not Enough: Time-Frequency based Explanation for Time-Series Black-Box Models},
  author={Chung, Hyunseung and Jo, Sumin and Kwon, Yeonsu and Choi, Edward},
  booktitle={ACM International Conference on Information and Knowledge Management},
  year={2024}
}

@inproceedings{brusch2025flextime,
  title={{FLEXtime}: Filterbank learning to explain time series},
  author={Br{\"u}sch, Thea and Wickstr{\o}m, Kristoffer K. and Schmidt, Mikkel N. and Jenssen, Robert and Alstr{\o}m, Tommy S.},
  booktitle={World Conference on eXplainable Artificial Intelligence},
  year={2025}
}

@article{chen2026freqlens,
  title={{FreqLens}: Interpretable Frequency Attribution for Time Series Forecasting},
  author={Chen, Chi-Sheng and Zhang, Xinyu and Kuo, En-Jui and Chen, Guan-Ying and Xie, Qiuzhe and Zhang, Fan},
  journal={arXiv preprint arXiv:2602.08768},
  year={2026}
}

@article{yeo2024financialxai,
  title={A comprehensive review on financial explainable {AI}},
  author={Yeo, Wei Jie and van der Heever, Wihan and Mao, Rui and Cambria, Erik and Satapathy, Ranjan and Mengaldo, Gianmarco},
  journal={Artificial Intelligence Review},
  year={2025}
}

@inproceedings{liu2022scinet,
  title={{SCINet}: Time series modeling and forecasting with sample convolution and interaction},
  author={Liu, Minhao and Zeng, Ailing and Chen, Muxi and Xu, Zhijian and Lai, Qiuxia and Ma, Lingna and Xu, Qiang},
  booktitle={Advances in Neural Information Processing Systems},
  year={2022}
}

@misc{yfinance,
  title={{yfinance}: {Yahoo!} {Finance} market data downloader},
  author={Aroussi, Ran},
  year={2024},
  howpublished={\url{https://github.com/ranaroussi/yfinance}}
}

@inproceedings{vaswani2017attention,
  title={Attention is all you need},
  author={Vaswani, Ashish and Shazeer, Noam and Parmar, Niki and Uszkoreit, Jakob and Jones, Llion and Gomez, Aidan N and Kaiser, Lukasz and Polosukhin, Illia},
  booktitle={Advances in Neural Information Processing Systems},
  year={2017}
}

@article{chen2023tsmixer,
  title={{TSMixer}: An all-{MLP} architecture for time series forecasting},
  author={Chen, Si-An and Li, Chun-Liang and Yoder, Nate and Arik, Sercan O. and Pfister, Tomas},
  journal={Transactions on Machine Learning Research},
  year={2023}
}

@article{zhang2022lightts,
  title={Less is more: Fast multivariate time series forecasting with light sampling-oriented {MLP} structures},
  author={Zhang, Tianping and Zhang, Yizhuo and Cao, Wei and Bian, Jiang and Yi, Xiaohan and Zheng, Shun and Li, Jian},
  journal={arXiv preprint arXiv:2207.01186},
  year={2022}
}

@inproceedings{zhou2022film,
  title={{FiLM}: Frequency improved {Legendre} memory model for long-term time series forecasting},
  author={Zhou, Tian and Ma, Ziqing and Wang, Xue and Wen, Qingsong and Sun, Liang and Yao, Tao and Yin, Wotao and Jin, Rong},
  booktitle={Advances in Neural Information Processing Systems},
  year={2022}
}

@inproceedings{yi2023frets,
  title={Frequency-domain {MLPs} are more effective learners in time series forecasting},
  author={Yi, Kun and Zhang, Qi and Fan, Wei and Wang, Shoujin and Wang, Pengyang and He, Hui and Lian, Defu and An, Ning and Cao, Longbing and Niu, Zhendong},
  booktitle={Advances in Neural Information Processing Systems},
  year={2023}
}

@article{das2023tide,
  title={Long-term forecasting with {TiDE}: Time-series dense encoder},
  author={Das, Abhimanyu and Kong, Weihao and Leach, Andrew and Mathur, Shaan and Sen, Rajat and Yu, Rose},
  journal={Transactions on Machine Learning Research},
  year={2023}
}

@inproceedings{wang2023micn,
  title={{MICN}: Multi-scale local and global context modeling for long-term series forecasting},
  author={Wang, Huiqiang and Peng, Jian and Huang, Feihu and Wang, Jince and Chen, Junhui and Xiao, Yifei},
  booktitle={International Conference on Learning Representations},
  year={2023}
}

@article{lin2023segrnn,
  title={{SegRNN}: Segment recurrent neural network for long-term time series forecasting},
  author={Lin, Shengsheng and Lin, Weiwei and Wu, Wentai and Zhao, Feiyu and Mo, Ruichao and Zhang, Haotong},
  journal={arXiv preprint arXiv:2308.11200},
  year={2023}
}

@inproceedings{cai2024msgnet,
  title={{MSGNet}: Learning multi-scale inter-series correlations for multivariate time series forecasting},
  author={Cai, Wanlin and Liang, Yuxuan and Liu, Xianggen and Feng, Jianshuai and Wu, Yuankai},
  booktitle={AAAI Conference on Artificial Intelligence},
  year={2024}
}

@inproceedings{liu2022pyraformer,
  title={Pyraformer: Low-complexity pyramidal attention for long-range time series modeling and forecasting},
  author={Liu, Shizhan and Yu, Hang and Liao, Cong and Li, Jianguo and Lin, Weiyao and Liu, Alex X. and Dustdar, Schahram},
  booktitle={International Conference on Learning Representations},
  year={2022}
}

@inproceedings{kitaev2020reformer,
  title={Reformer: The efficient {Transformer}},
  author={Kitaev, Nikita and Kaiser, Lukasz and Levskaya, Anselm},
  booktitle={International Conference on Learning Representations},
  year={2020}
}

@inproceedings{liu2022nonstationary,
  title={Non-stationary {Transformers}: Exploring the stationarity in time series forecasting},
  author={Liu, Yong and Wu, Haixu and Wang, Jianmin and Long, Mingsheng},
  booktitle={Advances in Neural Information Processing Systems},
  year={2022}
}

@inproceedings{zhang2023crossformer,
  title={Crossformer: {Transformer} utilizing cross-dimension dependency for multivariate time series forecasting},
  author={Zhang, Yunhao and Yan, Junchi},
  booktitle={International Conference on Learning Representations},
  year={2023}
}

@inproceedings{wang2024timexer,
  title={{TimeXer}: Empowering {Transformers} for time series forecasting with exogenous variables},
  author={Wang, Yuxuan and Wu, Haixu and Dong, Jiaxiang and Qin, Guo and Zhang, Haoran and Liu, Yong and Qiu, Yunzhong and Wang, Jianmin and Long, Mingsheng},
  booktitle={Advances in Neural Information Processing Systems},
  year={2024}
}

@article{multipatchformer,
  title={A multiscale model for multivariate time series forecasting},
  author={Naghashi, Vahid and Boukadoum, Mounir and Diallo, Abdoulaye Banire},
  journal={Scientific Reports},
  volume={15},
  number={1},
  pages={1565},
  year={2025}
}

@article{lee2026beyond,
  title={Beyond magnitude and shape: A direction-aware loss for time series forecasting},
  author={Lee, Seunghan and Lee, Jaehoon and Seo, Jun and Kang, Junhyeok and Han, Sangjun and Yoo, Sungdong and Kim, Minjae and Lim, Tae Yoon and Kang, Dongwan and Choi, Hwanil and Lee, Soonyoung and Ahn, Wonbin},
  journal={arXiv preprint arXiv:2608.01857},
  year={2026}
}

@inproceedings{lee2026finstar,
  title={{FinSTaR}: Towards financial reasoning with time series reasoning models},
  author={Lee, Seunghan and Seo, Jun and Lee, Jaehoon and Yoo, Sungdong and Kim, Minjae and Lim, Tae Yoon and Kang, Dongwan and Choi, Hwanil and Lee, Soonyoung and Ahn, Wonbin},
  booktitle={Conference on Empirical Methods in Natural Language Processing: Industry Track},
  year={2026}
}

@inproceedings{lee2024learning,
  title={Learning to embed time series patches independently},
  author={Lee, Seunghan and Park, Taeyoung and Lee, Kibok},
  booktitle={International Conference on Learning Representations},
  year={2024}
}

@inproceedings{roy2007erank,
  title={The effective rank: A measure of effective dimensionality},
  author={Roy, Olivier and Vetterli, Martin},
  booktitle={European Signal Processing Conference},
  pages={606--610},
  year={2007}
}

@article{litwinkumar2017dim,
  title={Optimal degrees of synaptic connectivity},
  author={Litwin-Kumar, Ashok and Harris, Kameron Decker and Axel, Richard and Sompolinsky, Haim and Abbott, L F},
  journal={Neuron},
  volume={93},
  number={5},
  pages={1153--1164},
  year={2017}
}
\bibliographystyle{iclr2027_conference}

\clearpage
\appendix

\section{Experimental Details}
\label{sec:appendix}
\label{sec:setup}

\noindent\textbf{Swept configurations.}
All experiments are univariate except the channel-axis comparison of
Section~\ref{sec:timechan}, and we sweep
$L \in \{48, 96, 192, 252, 336, 512\}$, $H \in \{24, 48, 96, 192, 336, 720\}$, and depth in
$\{1, 2, 3, 4, 6\}$.
Five benchmarks hold hundreds of series each, which are Solar, the four PEMS sets, and the
financial set.
For these we sample series at even intervals rather than running all of them, since several
hundred sensors on one road network carry nearly the same information.

\noindent\textbf{Reporting.}
Note that every configuration is run with three random seeds and that we report the seed
standard deviation separately from the across-configuration standard deviation, since pooling
them would make the estimate look more stable than it is.
Furthermore, we report an architecture only when it has at least \nMinRuns{} completed runs, so that a
backbone with a handful of runs does not sit beside one with thousands under the same visual
weight.
We store metrics and figures only rather than model weights or predicted series.

\noindent\textbf{Pretrained forecasters.}
The checkpoints are named in full in Table~\ref{tab:fm_models}, since each family releases
several sizes.
TimesFM requires the lookback to be a multiple of its input patch length, so we keep the whole
pretrained grid at $L \in \{96, 192, 320, 512\}$ and $H \in \{24, 48, 96, 192\}$ to compare the
three models under identical conditions.
GIFT-Eval is used at every dataset and sampling frequency whose test split admits a window at
the smallest of those settings, and the configurations this excludes are the ones whose series
are shorter than $L + H$.

\vspace{30pt}
\section{Datasets}
\label{sec:datasets}

\noindent\textbf{General domain.}
The suite is ETTh1, ETTh2, ETTm1 and ETTm2 \citep{zhou2021informer}, Electricity, Traffic,
Weather, Exchange and Solar-Energy \citep{wu2021autoformer}, and PEMS03, PEMS04, PEMS07 and
PEMS08 \citep{liu2022scinet}, which is the set the forecasting literature reports on.
We take every one of them from the Time-Series-Library \citep{wang2024tslib} so that the split
and the preprocessing are the ones the field already uses.

\noindent\textbf{Financial domain.}
We assemble this benchmark ourselves, since the general suite is dominated by strongly periodic
series and we want a domain where periodicity is nearly absent.
We retrieve daily closing prices from 2018 to 2025 with \texttt{yfinance} \citep{yfinance}, and
keep the 100 largest United States companies by market capitalization whose history is complete
over that window, so that no series is padded.
Trading days are the grid and we do not interpolate onto calendar days, since a non-trading day
is an absence of observation rather than a missing value.
The same 100 companies enter in two forms, which are the log price and the log return, and
Appendix~\ref{sec:domdetail} reports them apart, since the two differ in predictability.

\noindent\textbf{The 100 companies.}
AAPL, ABBV, ABT, ACN, ADBE, ADI, ADP, AMD, AMGN, AMT, AMZN, ANET, AON, APH, AVGO, AXP, BAC, BKNG, BLK, BMY, BRK-B, BSX, C, CAT, CB, CI, CL, CME, COST, CRM, CSCO, CVX, DE, DHR, DIS, DUK, ELV, EMR, EOG, ETN, GE, GILD, GOOG, GOOGL, GS, HD, HON, IBM, ICE, INTC, INTU, ISRG, ITW, JNJ, JPM, KLAC, KO, LIN, LLY, LMT, LOW, MA, MCD, MCK, MDT, META, MO, MRK, MS, MSFT, MU, NEO, NFLX, NKE, NOC, NOW, NVDA, ORCL, PANW, PEP, PFE, PG, PGR, PLD, PM, PYPL, QCOM, REGN, RTX, SBUX, SCHW, SHW, SO, SPGI, SYK, T, TJX, TMO, TSLA, TXN.

\clearpage
\section{Backbones}
\label{sec:backbones}

We explain \nModels{} backbones, grouped by family. Within each family they appear in the order
every table of this appendix uses, and we state the one idea each rests on.

\noindent\textbf{Basic architectures.}
\begin{itemize}
\setlength{\itemsep}{0pt}\setlength{\parsep}{0pt}\setlength{\parskip}{0pt}
\setlength{\topsep}{4pt}
\item \textbf{Linear.} A single map from the lookback window to the forecast.
\item \textbf{CNN.} A stack of temporal convolutions over the lookback window.
\item \textbf{Transformer} \citep{vaswani2017attention}. Vanilla self-attention over time steps.
\end{itemize}

\noindent\textbf{Linear family.}
\begin{itemize}
\setlength{\itemsep}{0pt}\setlength{\parsep}{0pt}\setlength{\parskip}{0pt}
\setlength{\topsep}{4pt}
\item \textbf{DLinear} \citep{zeng2023dlinear}. Splits trend from seasonality and fits a linear
map to each.
\item \textbf{TSMixer} \citep{chen2023tsmixer}. Alternates mixing over time and over channels.
\item \textbf{LightTS} \citep{zhang2022lightts}. Samples the series at two rates before mixing.
\item \textbf{FiLM} \citep{zhou2022film}. Projects the lookback onto Legendre polynomials.
\item \textbf{FreTS} \citep{yi2023frets}. Mixes in the frequency domain rather than in time.
\item \textbf{TiDE} \citep{das2023tide}. Encodes the lookback with a dense residual network.
\end{itemize}

\noindent\textbf{Convolution and recurrence.}
\begin{itemize}
\setlength{\itemsep}{0pt}\setlength{\parsep}{0pt}\setlength{\parskip}{0pt}
\setlength{\topsep}{4pt}
\item \textbf{SCINet} \citep{liu2022scinet}. Downsamples the series into branches that interact.
\item \textbf{MICN} \citep{wang2023micn}. Combines local convolution with a global view.
\item \textbf{TimesNet} \citep{wu2023timesnet}. Folds the series into two dimensions by period
and applies a vision backbone.
\item \textbf{SegRNN} \citep{lin2023segrnn}. Runs a recurrent unit over segments rather than
points.
\item \textbf{TimeMixer} \citep{wang2024timemixer}. Mixes across several sampling scales.
\item \textbf{MSGNet} \citep{cai2024msgnet}. Learns a graph over those scales.
\end{itemize}

\noindent\textbf{Transformer family.}
\begin{itemize}
\setlength{\itemsep}{0pt}\setlength{\parsep}{0pt}\setlength{\parskip}{0pt}
\setlength{\topsep}{4pt}
\item \textbf{Informer} \citep{zhou2021informer}. Sparsifies attention by a query score.
\item \textbf{Autoformer} \citep{wu2021autoformer}. Replaces attention with autocorrelation.
\item \textbf{FEDformer} \citep{zhou2022fedformer}. Attends in the frequency domain.
\item \textbf{Pyraformer} \citep{liu2022pyraformer}. Attends over a pyramid of resolutions.
\item \textbf{Reformer} \citep{kitaev2020reformer}. Hashes queries into buckets.
\item \textbf{Non-stationary Transformer} \citep{liu2022nonstationary}. Normalizes the input and
returns the statistics to the attention.
\item \textbf{Crossformer} \citep{zhang2023crossformer}. Attends across time and across channels
in turn.
\item \textbf{PatchTST} \citep{nie2023patchtst}. Tokenizes contiguous patches and keeps channels
independent.
\item \textbf{iTransformer} \citep{liu2024itransformer}. Treats each channel as a token.
\item \textbf{TimeXer} \citep{wang2024timexer}. Adds exogenous variables to the patch tokens.
\item \textbf{MultiPatchFormer} \citep{multipatchformer}. Reads several patch sizes at once.
\end{itemize}

\clearpage
\section{Seed Reproducibility}
\label{sec:seeds}

Table~\ref{tab:seeds} folds the three random seeds within each configuration before
aggregating, which separates two quantities that a pooled average would confuse.
The seed standard deviation is the spread of the same configuration across seeds, and the
configuration standard deviation is the spread across configurations.
We fold before aggregating on purpose, since configurations differ in how many runs
they have and a pooled average would weight them unequally.
The seed spread is smaller than the configuration spread for every quantity, so the variation
we report across architectures and benchmarks is a property of those settings rather than of
the random draw.

\begin{table}[h]
\caption{\textbf{Seed reproducibility.} Over the \sfGainN{} configurations for which all three
quantities completed on all three seeds, where each configuration is folded before aggregating.}
\label{tab:seeds}
\begin{center}
\begin{tabular}{lccc}
\toprule
\textbf{Quantity} & \textbf{Mean} & \textbf{Std (seed)} & \textbf{Std (configuration)} \\
\midrule
$g_{\text{own}}$ & \textcolor{hred}{$\mathbf{\sfGain}$} & $\sfGainSeed$ & $\sfGainConfig$ \\
$g_{\text{margin}}$ & \textcolor{hred}{$\mathbf{\sfMargin}$} & $\sfMarginSeed$ & $\sfMarginConfig$ \\
Effective rank of $E$ & $\sfRank$ & $\sfRankSeed$ & $\sfRankConfig$ \\
\bottomrule
\end{tabular}
\end{center}
\end{table}

\clearpage
\section{Estimator Details}
\label{sec:methoddetail}

\subsection{Filling the Matrix with Batched Backward Calls}
\label{sec:batching}

Automatic differentiation propagates from a single scalar, so the naive way to fill $E$ is to
select $\hat{y}[h]$, run a backward pass, keep the resulting row, and repeat for every $h$.
That costs $H$ backward passes per window, which is $720$ passes at our longest horizon.

\begin{figure}[h]
\centering
\includegraphics[width=\linewidth]{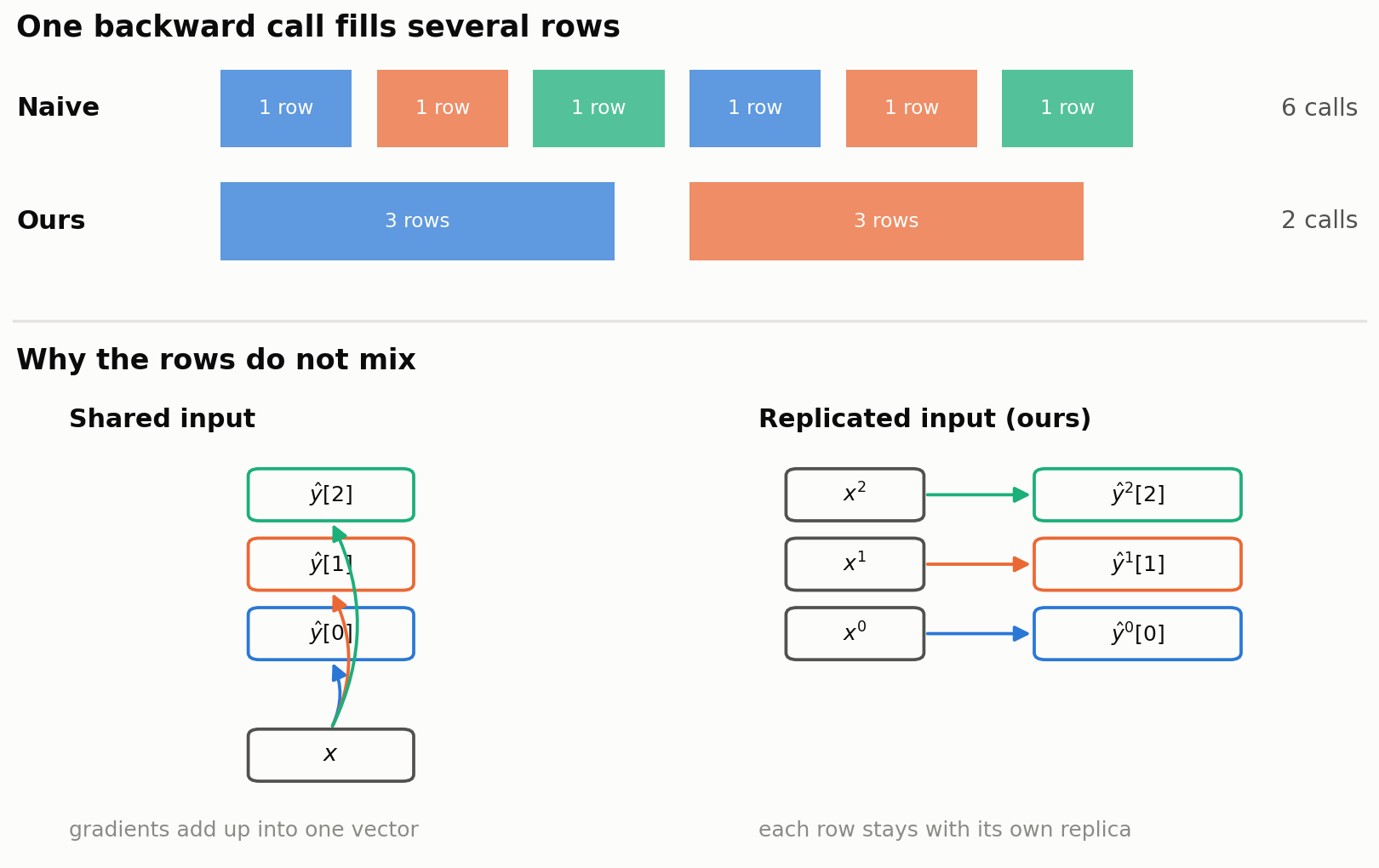}
\caption{\textbf{Filling the explanation matrix with batched backward calls.} \textrm{[Top]} Backward
calls for $H=6$, where the naive way spends one per row and ours spends one per group of $k=3$.
\textrm{[Bottom]} Why grouping is safe. A shared input adds the gradients together and returns
the shared vector, whereas one replica per step returns every row separately.}
\label{fig:batching}
\end{figure}

We instead replicate the input $k$ times and assign a different forecast step to each replica.
Writing $x^{i}$ for the $i$-th replica and $h_i$ for the step assigned to it, we differentiate
the single scalar $\sum_{i} \hat{y}^{i}[h_i]$.
The rows do not mix, because each replica is a separate leaf of the computation graph and the
terms belonging to the other replicas do not depend on $x^{i}$,
\begin{equation}
\frac{\partial}{\partial x^{i}} \sum_{j} \hat{y}^{j}[h_j]
\;=\; \frac{\partial \hat{y}^{i}[h_i]}{\partial x^{i}} .
\label{eq:batch}
\end{equation}
One backward call therefore returns $k$ distinct rows rather than their sum, as shown in
Figure~\ref{fig:batching}.
The replication is what makes this work, since differentiating the same sum through one shared
input would collapse the terms into $\sum_h \partial \hat{y}[h] / \partial x$, which is the
single vector of prior work and not a row of $E$.
We use $k=16$, so a horizon of $720$ needs $45$ backward calls instead of $720$.
The cost is paid in memory rather than time, as the batch grows by a factor of $k$, and $k$ sets
that trade-off.

\vspace{20pt}
\subsection{Implementation Details of the Estimator}
\label{sec:obstacles}
Reading an input gradient out of a TS backbone meets two difficulties, and one fails
\textit{loudly} while the other fails \textit{quietly}.
The first is that many backbones normalize their input in place, as in \texttt{x /= stdev}, which
autograd refuses to differentiate because the overwritten tensor is the one it saved for the
backward pass.
We therefore rewrite these operations out of place with a \texttt{TorchFunctionMode}, and
assert that the forward output is identical before and after, since code that relies on the side
effect of an in-place call would otherwise become a no-op without any sign.
The second is that a network can genuinely ignore its input, as we found one backbone whose
decoder outputs a constant at initialization, which makes both the gradient and the deletion effect
exactly zero.
Note that such a zero is a fact about the untrained model rather than a finding about
forecasting, so we validate the estimator on trained models only.

\vspace{20pt}
\subsection{Porting the Estimators to Forecasting}
\label{sec:porting}

Every estimator of Section~\ref{sec:methods} is reimplemented from the definition in its own
paper rather than taken from a library, since this puts all of them on the same forward
interface, the same perturbation operator, and the same protocol.
The quantity each one is asked to preserve is the only thing we change, and it becomes
$\hat{y}[h]$ in place of the scalar the original method reads.

\noindent\textbf{Gradient methods.}
Saliency, gradient $\times$ input, and integrated gradients differentiate $\hat{y}[h]$ instead
of a class score, and integrated gradients uses the mean of the lookback window as its baseline.
Occlusion corrupts one window of the lookback at a time and reads the change at step $h$ alone,
rather than the change in a class score.

\noindent\textbf{Mask methods.}
Dynamask, ExtremalMask, and ContraLSP were published for time series, but their objective
reduces the output to a scalar.
We keep their objective and let it preserve $\hat{y}[h]$ under the same budget on the mask, so
the axis makes each of them solve one optimization problem per forecast step.
Note that the same code with the axis removed preserves the whole forecast at once, which is the
form closest to the original papers and the one scored in Appendix~\ref{sec:plainmethods}.

\noindent\textbf{Rows we compute.}
A row is estimated at eight evenly spaced forecast steps rather than at all $H$ of them, since
the mask methods would otherwise need $H$ separate optimizations per window.
The protocol of Section~\ref{sec:protocol} scores exactly the steps that were estimated.

\clearpage
\section{Metric Details}
\label{sec:metricdetail}

\subsection{Definition of the Deletion Curve}
\label{sec:auccurve}

Section~\ref{sec:protocol} writes $\mathrm{AUC}(h; v)$ for the area under a deletion curve and
leaves the curve itself implicit, so we give it here.
Let $v \in \mathbb{R}^{L}$ be an ordering vector over the lookback window, let
$\rho = k / L \in [0, 1]$ be the deleted fraction, and let $S_{\rho}(v)$ be the $k$ positions that
$v$ ranks highest.
Writing $x \setminus S_{\rho}(v)$ for the input with those positions perturbed, the curve and its
area are
\begin{equation}
\varepsilon(h, \rho; v) \;=\;
  \bigl\lVert f\bigl(x \setminus S_{\rho}(v)\bigr)[h] - \hat{y}[h] \bigr\rVert^{2} ,
\qquad
\mathrm{AUC}(h; v) \;=\; \int_{0}^{1} \varepsilon(h, \rho; v) \, d\rho .
\label{eq:auccurve}
\end{equation}
Every quantity in Section~\ref{sec:protocol} is this one function read at a different ordering
vector, namely $v = E[h]$ for the row that belongs to the step, $v = \bar{E}$ for the single
vector a prior method supplies, and $v = E[\pi(h)]$ for the row of another step under the shuffle
control.
Note that the error is taken against the model's own unperturbed forecast $\hat{y}$ rather than
against the true future, so the curve reads faithfulness to the model and never forecast
accuracy, and $\varepsilon(h, 0; v) = 0$ holds for every $v$.
Perturbation resamples the selected positions from elsewhere in the same window instead of
setting them to zero, which keeps the perturbed input in distribution \citep{hooker2019roar}.
We evaluate the integral by the trapezoidal rule on nine fractions evenly spaced in
$[0, 1]$.

\vspace{20pt}
\subsection{Reporting the Deletion Curve}
\label{sec:morf}

A deletion curve can be swept in two orders, which are most-relevant-first (MoRF) and
least-relevant-first (LeRF).
We report the gap between the two areas rather than the MoRF area alone, since both curves end
at the same point when every position has been deleted.
Their areas therefore converge as $k$ approaches $L$ no matter which explanation was used.
The gap keeps the part of the curve where the two orders actually disagree.

\vspace{20pt}
\subsection{Definitions of the Borrowed Metrics}
\label{sec:stddefs}

Every metric in Section~\ref{sec:stdmetrics} was defined for classification, where the model
outputs one label, so we restate each definition for forecasting below.
We reuse the notation of Appendix~\ref{sec:auccurve} and write $\varepsilon(h, \rho; v)$ for the
deletion error and
$\tilde{\varepsilon}(h, \rho; v) = \lVert f\bigl(x \mid S_{\rho}(v)\bigr)[h] - \hat{y}[h] \rVert^{2}$
for the retention error, in which only $S_{\rho}(v)$ is left intact and the rest of the window is
perturbed.
Averaging over the swept grid $R$ of fractions gives
\begin{align*}
\text{Comprehensiveness}(h) &= \textstyle\frac{1}{|R|}\sum_{\rho \in R} \varepsilon(h, \rho; v) &&\text{higher is better} \\
\text{Sufficiency}(h) &= \textstyle\frac{1}{|R|}\sum_{\rho \in R} \tilde{\varepsilon}(h, \rho; v) &&\text{lower is better} \\
\text{AOPC}(h) &= \textstyle\int \varepsilon(h, \rho; v) \, d\rho &&\text{higher is better} \\
\text{Insertion}(h) &= \textstyle\frac{1}{|R|}\sum_{\rho \in R} \bigl[ 1 - \tilde{\varepsilon}(h, \rho; v) / \tilde{\varepsilon}(h, 0; v) \bigr] &&\text{higher is better}
\end{align*}
Here $\tilde{\varepsilon}(h, 0; v)$ is the retention error with nothing kept, which is the fully
perturbed window that Insertion starts from.
Note that Comprehensiveness and AOPC read the same deletion curve as
Equation~\ref{eq:auccurve}, and they differ from our gain only in how that curve becomes a
number.
The evidence independent of our own construction therefore comes from Sufficiency and Insertion,
which perturb in the opposite direction.
Each metric is evaluated twice, once at $v = E[h]$ and once at $v = \bar{E}$, and we sign the
difference so that positive always means the horizon axis helped.

\clearpage
\section{Synthetic Data and Calibration}

\noindent\textbf{Constructed ground truth.}
\label{sec:gt}
Every generator writes the target as $y_h = w^{(h)} \cdot x_{S(h)}$ and differs only in how the
set $S(h)$ of lookback positions moves with the forecast step $h$, and
Table~\ref{tab:gens} lists the six.
The weight vector $w^{(h)}$ has unit norm and varies smoothly with $h$, so that
the target moves like a series instead of staying constant.
Every $S(h)$ contains the last lookback position, so that the forecast starts from the value the
series just took.
The positions and weights are drawn once and reused for every window, since a map that
changed every window would leave nothing to recover.
We verified that a least-squares fit explains at least $98\%$ of the target variance on held-out
windows.

\begin{table}[h]
\caption{\textbf{The six synthetic generators.} Each row states how the set $S(h)$ read by step
$h$ is placed. Only \texttt{sparsenull} makes every row a multiple of one vector, which is the
negative control.}
\label{tab:gens}
\begin{center}
\small
\begin{fittable}
\begin{tabular}{lll}
\toprule
\textbf{Generator} & \textbf{Placement of $S(h)$} & \textbf{Role} \\
\midrule
\texttt{sparsenull} & Fixed, with $w^{(h)} = a_h w$ for one fixed $w$ & Negative control \\
\texttt{sparsesplit} & One set for $h < H/2$ and a disjoint one after & Two regimes \\
\texttt{sparseband} & One set per block of forecast steps & $B$ regimes \\
\texttt{sparseshift} & Slid left in proportion to $h$ & Gradual drift \\
\texttt{lagmix} & Placed at period-aligned lags & Seasonal structure \\
\texttt{plantrank} & Rows of a matrix $A$ of planted rank, $y = A x$ & Known rank \\
\bottomrule
\end{tabular}
\end{fittable}
\end{center}
\end{table}

\begin{figure}[h]
\centering
\includegraphics[width=\linewidth]{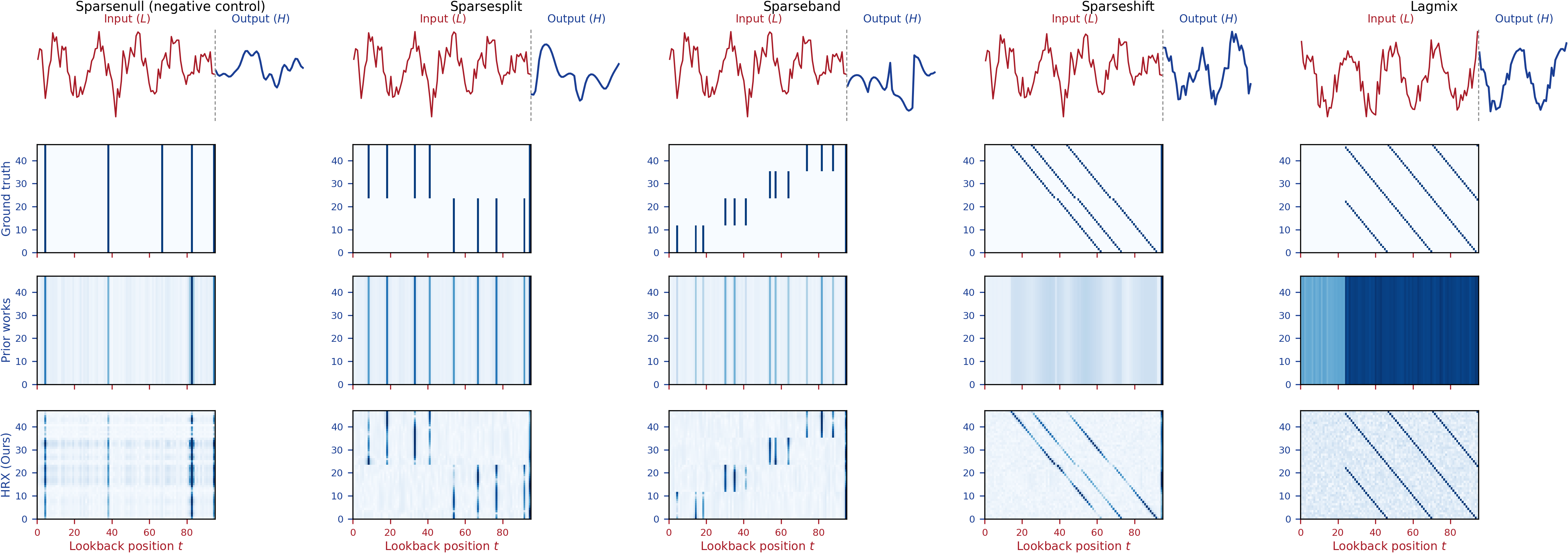}
\caption{\textbf{What each method recovers.} The rows are one window, the answer the generator
fixes, what a prior method returns, and our matrix. The last two come from the same trained
backbone, and only \texttt{sparsenull} makes them agree.}
\label{fig:synthfull}
\end{figure}

\noindent\textbf{Why only constructive labels.}
A label that only \textit{asserts} which inputs should matter is not enough, since it can be
wrong in a way the metric cannot detect.
Our first attempt asserted that a sinusoidal generator depends on the points one period back.
That assertion does not hold, since a sinusoid is fixed by any two points, which means a model
reading two other points is equally correct.
We discarded the conclusion it produced.
Note that the answer belongs to the generator while the explanation belongs to the model.
Agreement is therefore evidence that the measurement is sound, while disagreement can also mean
that the model found an equal solution the generator did not name.
Figure~\ref{fig:synthfull} shows what each method recovers on every generator,
Table~\ref{tab:synthall} repeats the body measurement for every backbone, and
Table~\ref{tab:synthby} splits the same measurement by backbone and generator.

\noindent\textbf{The negative control.}
\texttt{sparsenull} sets $w^{(h)} = a_h w$ for one fixed $w$, so every forecast step truly
depends on the same lookback positions and only the magnitude changes with $h$.
The horizon axis has nothing to resolve there, and the effective rank of the answer is one,
which is exactly the assumption prior works make.
It is included because the protocol of Section~\ref{sec:protocol} favors a concentrated
explanation, so a positive gain could in principle come from the instrument rather than from
the data.
As shown in Table~\ref{tab:synthby}, the gap on \texttt{sparsenull} is not positive on any
backbone, whereas it is positive on the other five generators.
The gain reported in Section~\ref{sec:sanity} therefore belongs to the horizon structure the
generator planted rather than to the way it is measured.

\begin{table}[h]
\caption{\textbf{Explanation performance on the synthetic dataset, all backbones.} The body
reports the three minimal architectures. This is the same measurement for every backbone.}
\label{tab:synthall}
\begin{center}
\tiny
\begin{fittable}\begin{NiceTabular}{l|c|c|c|cc|cc|cc|cc}
\toprule
\Block{3-1}{\textbf{Backbone}} & \multicolumn{3}{c|}{\textbf{Proposed metrics}\textsuperscript{1}} & \multicolumn{8}{c}{\textbf{Ground-truth metrics}\textsuperscript{2}} \\
\cmidrule(lr){2-4}\cmidrule(lr){5-12}
 & \Block{2-1}{\textbf{$g_{\text{own}}$}} & \Block{2-1}{\textbf{$g_{\text{shuffled}}$}} & \Block{2-1}{\textbf{$g_{\text{margin}}$}} & \multicolumn{2}{c|}{\textbf{AUPRC}} & \multicolumn{2}{c|}{\textbf{AUROC}} & \multicolumn{2}{c|}{\textbf{AUP}} & \multicolumn{2}{c}{\textbf{AUR}} \\
\cmidrule(lr){5-6}\cmidrule(lr){7-8}\cmidrule(lr){9-10}\cmidrule(lr){11-12}
 &  &  &  & Vector & Matrix & Vector & Matrix & Vector & Matrix & Vector & Matrix \\
\midrule
Linear & \textcolor{hred}{$\mathbf{0.142}$} & \textcolor{hblue}{$\mathbf{-0.215}$} & \textcolor{hred}{$\mathbf{0.357}$} & $0.586$ & \textcolor{hred}{$\mathbf{0.983}$} & $0.961$ & \textcolor{hred}{$\mathbf{0.998}$} & $0.132$ & \textcolor{hred}{$\mathbf{0.154}$} & $0.960$ & \textcolor{hred}{$\mathbf{0.985}$} \\
CNN & \textcolor{hred}{$\mathbf{0.161}$} & \textcolor{hblue}{$\mathbf{-0.234}$} & \textcolor{hred}{$\mathbf{0.395}$} & $0.578$ & \textcolor{hred}{$\mathbf{0.993}$} & $0.962$ & \textcolor{hred}{$\mathbf{0.999}$} & $0.132$ & \textcolor{hred}{$\mathbf{0.154}$} & $0.963$ & \textcolor{hred}{$\mathbf{0.984}$} \\
Transformer & \textcolor{hred}{$\mathbf{0.121}$} & \textcolor{hblue}{$\mathbf{-0.252}$} & \textcolor{hred}{$\mathbf{0.373}$} & $0.577$ & \textcolor{hred}{$\mathbf{0.993}$} & $0.960$ & \textcolor{hred}{$\mathbf{0.999}$} & $0.130$ & \textcolor{hred}{$\mathbf{0.154}$} & $0.963$ & \textcolor{hred}{$\mathbf{0.985}$} \\
DLinear & \textcolor{hred}{$\mathbf{0.153}$} & \textcolor{hblue}{$\mathbf{-0.197}$} & \textcolor{hred}{$\mathbf{0.350}$} & $0.586$ & \textcolor{hred}{$\mathbf{0.994}$} & $0.968$ & \textcolor{hred}{$\mathbf{1.000}$} & $0.133$ & \textcolor{hred}{$\mathbf{0.156}$} & $0.963$ & \textcolor{hred}{$\mathbf{0.987}$} \\
TSMixer & \textcolor{hred}{$\mathbf{0.140}$} & \textcolor{hblue}{$\mathbf{-0.210}$} & \textcolor{hred}{$\mathbf{0.350}$} & $0.585$ & \textcolor{hred}{$\mathbf{1.000}$} & $0.967$ & \textcolor{hred}{$\mathbf{1.000}$} & $0.133$ & \textcolor{hred}{$\mathbf{0.156}$} & $0.963$ & \textcolor{hred}{$\mathbf{0.987}$} \\
LightTS & \textcolor{hred}{$\mathbf{0.156}$} & \textcolor{hblue}{$\mathbf{-0.210}$} & \textcolor{hred}{$\mathbf{0.366}$} & $0.573$ & \textcolor{hred}{$\mathbf{0.991}$} & $0.963$ & \textcolor{hred}{$\mathbf{1.000}$} & $0.133$ & \textcolor{hred}{$\mathbf{0.156}$} & $0.961$ & \textcolor{hred}{$\mathbf{0.987}$} \\
FiLM & \textcolor{hred}{$\mathbf{0.073}$} & \textcolor{hblue}{$\mathbf{-0.194}$} & \textcolor{hred}{$\mathbf{0.267}$} & $0.569$ & \textcolor{hred}{$\mathbf{0.934}$} & $0.958$ & \textcolor{hred}{$\mathbf{0.979}$} & $0.133$ & \textcolor{hred}{$\mathbf{0.151}$} & $0.955$ & \textcolor{hred}{$\mathbf{0.973}$} \\
FreTS & \textcolor{hred}{$\mathbf{0.110}$} & \textcolor{hblue}{$\mathbf{-0.149}$} & \textcolor{hred}{$\mathbf{0.259}$} & $0.583$ & \textcolor{hred}{$\mathbf{1.000}$} & $0.967$ & \textcolor{hred}{$\mathbf{1.000}$} & $0.133$ & \textcolor{hred}{$\mathbf{0.156}$} & $0.963$ & \textcolor{hred}{$\mathbf{0.987}$} \\
TiDE & \textcolor{hred}{$\mathbf{0.128}$} & \textcolor{hblue}{$\mathbf{-0.265}$} & \textcolor{hred}{$\mathbf{0.393}$} & $0.566$ & \textcolor{hred}{$\mathbf{0.933}$} & $0.960$ & \textcolor{hred}{$\mathbf{0.973}$} & $0.131$ & \textcolor{hred}{$\mathbf{0.150}$} & $\mathbf{0.961}$ & $0.961$ \\
SCINet & \textcolor{hred}{$\mathbf{0.166}$} & \textcolor{hblue}{$\mathbf{-0.197}$} & \textcolor{hred}{$\mathbf{0.364}$} & $0.579$ & \textcolor{hred}{$\mathbf{0.981}$} & $0.960$ & \textcolor{hred}{$\mathbf{0.998}$} & $0.133$ & \textcolor{hred}{$\mathbf{0.154}$} & $0.961$ & \textcolor{hred}{$\mathbf{0.983}$} \\
MICN & \textcolor{hred}{$\mathbf{0.058}$} & \textcolor{hblue}{$\mathbf{-0.211}$} & \textcolor{hred}{$\mathbf{0.269}$} & $0.586$ & \textcolor{hred}{$\mathbf{0.994}$} & $0.967$ & \textcolor{hred}{$\mathbf{1.000}$} & $0.131$ & \textcolor{hred}{$\mathbf{0.156}$} & $0.963$ & \textcolor{hred}{$\mathbf{0.987}$} \\
TimesNet & \textcolor{hred}{$\mathbf{0.121}$} & \textcolor{hblue}{$\mathbf{-0.234}$} & \textcolor{hred}{$\mathbf{0.355}$} & $0.576$ & \textcolor{hred}{$\mathbf{0.966}$} & $0.960$ & \textcolor{hred}{$\mathbf{0.993}$} & $0.136$ & \textcolor{hred}{$\mathbf{0.154}$} & $0.963$ & \textcolor{hred}{$\mathbf{0.982}$} \\
SegRNN & \textcolor{hred}{$\mathbf{0.060}$} & \textcolor{hblue}{$\mathbf{-0.163}$} & \textcolor{hred}{$\mathbf{0.224}$} & $0.388$ & \textcolor{hred}{$\mathbf{0.829}$} & $0.918$ & \textcolor{hred}{$\mathbf{0.934}$} & $0.114$ & \textcolor{hred}{$\mathbf{0.139}$} & $\mathbf{0.917}$ & $0.903$ \\
TimeMixer & \textcolor{hred}{$\mathbf{0.126}$} & \textcolor{hblue}{$\mathbf{-0.243}$} & \textcolor{hred}{$\mathbf{0.369}$} & $0.566$ & \textcolor{hred}{$\mathbf{0.981}$} & $0.962$ & \textcolor{hred}{$\mathbf{0.998}$} & $0.135$ & \textcolor{hred}{$\mathbf{0.153}$} & $0.963$ & \textcolor{hred}{$\mathbf{0.980}$} \\
MSGNet & \textcolor{hred}{$\mathbf{0.136}$} & \textcolor{hblue}{$\mathbf{-0.187}$} & \textcolor{hred}{$\mathbf{0.323}$} & $0.571$ & \textcolor{hred}{$\mathbf{0.993}$} & $0.960$ & \textcolor{hred}{$\mathbf{1.000}$} & $0.133$ & \textcolor{hred}{$\mathbf{0.156}$} & $0.961$ & \textcolor{hred}{$\mathbf{0.986}$} \\
Informer & \textcolor{hred}{$\mathbf{0.081}$} & \textcolor{hblue}{$\mathbf{-0.137}$} & \textcolor{hred}{$\mathbf{0.218}$} & $0.536$ & \textcolor{hred}{$\mathbf{0.877}$} & $0.845$ & \textcolor{hred}{$\mathbf{0.959}$} & $0.128$ & \textcolor{hred}{$\mathbf{0.148}$} & $0.842$ & \textcolor{hred}{$\mathbf{0.939}$} \\
Autoformer & \textcolor{hred}{$\mathbf{0.065}$} & \textcolor{hblue}{$\mathbf{-0.102}$} & \textcolor{hred}{$\mathbf{0.168}$} & $\mathbf{0.273}$ & $0.184$ & $\mathbf{0.586}$ & $0.571$ & $\mathbf{0.068}$ & $0.068$ & $\mathbf{0.589}$ & $0.576$ \\
FEDformer & \textcolor{hred}{$\mathbf{0.037}$} & \textcolor{hblue}{$\mathbf{-0.103}$} & \textcolor{hred}{$\mathbf{0.141}$} & $0.375$ & \textcolor{hred}{$\mathbf{0.393}$} & $0.723$ & \textcolor{hred}{$\mathbf{0.758}$} & $0.098$ & \textcolor{hred}{$\mathbf{0.098}$} & $0.724$ & \textcolor{hred}{$\mathbf{0.745}$} \\
Pyraformer & \textcolor{hred}{$\mathbf{0.061}$} & \textcolor{hblue}{$\mathbf{-0.045}$} & \textcolor{hred}{$\mathbf{0.106}$} & $0.471$ & \textcolor{hred}{$\mathbf{0.833}$} & $0.824$ & \textcolor{hred}{$\mathbf{0.903}$} & $0.121$ & \textcolor{hred}{$\mathbf{0.135}$} & $0.825$ & \textcolor{hred}{$\mathbf{0.838}$} \\
Reformer & \textcolor{hred}{$\mathbf{0.021}$} & \textcolor{hblue}{$\mathbf{-0.117}$} & \textcolor{hred}{$\mathbf{0.139}$} & $\mathbf{0.369}$ & $0.320$ & $0.685$ & \textcolor{hred}{$\mathbf{0.760}$} & $0.088$ & \textcolor{hred}{$\mathbf{0.093}$} & $0.684$ & \textcolor{hred}{$\mathbf{0.757}$} \\
Nonstat. Trans. & \textcolor{hred}{$\mathbf{0.021}$} & \textcolor{hblue}{$\mathbf{-0.111}$} & \textcolor{hred}{$\mathbf{0.133}$} & $0.484$ & \textcolor{hred}{$\mathbf{0.882}$} & $0.917$ & \textcolor{hred}{$\mathbf{0.959}$} & $0.100$ & \textcolor{hred}{$\mathbf{0.123}$} & $0.918$ & \textcolor{hred}{$\mathbf{0.944}$} \\
Crossformer & \textcolor{hred}{$\mathbf{0.152}$} & \textcolor{hblue}{$\mathbf{-0.163}$} & \textcolor{hred}{$\mathbf{0.315}$} & $0.582$ & \textcolor{hred}{$\mathbf{1.000}$} & $0.956$ & \textcolor{hred}{$\mathbf{1.000}$} & $0.133$ & \textcolor{hred}{$\mathbf{0.156}$} & $0.952$ & \textcolor{hred}{$\mathbf{0.986}$} \\
PatchTST & \textcolor{hred}{$\mathbf{0.149}$} & \textcolor{hblue}{$\mathbf{-0.271}$} & \textcolor{hred}{$\mathbf{0.420}$} & $0.577$ & \textcolor{hred}{$\mathbf{0.994}$} & $0.967$ & \textcolor{hred}{$\mathbf{1.000}$} & $0.133$ & \textcolor{hred}{$\mathbf{0.155}$} & $0.963$ & \textcolor{hred}{$\mathbf{0.984}$} \\
TimeXer & \textcolor{hred}{$\mathbf{0.136}$} & \textcolor{hblue}{$\mathbf{-0.246}$} & \textcolor{hred}{$\mathbf{0.382}$} & $0.579$ & \textcolor{hred}{$\mathbf{1.000}$} & $0.967$ & \textcolor{hred}{$\mathbf{1.000}$} & $0.133$ & \textcolor{hred}{$\mathbf{0.156}$} & $0.963$ & \textcolor{hred}{$\mathbf{0.984}$} \\
MultiPatchFormer & \textcolor{hred}{$\mathbf{0.147}$} & \textcolor{hblue}{$\mathbf{-0.179}$} & \textcolor{hred}{$\mathbf{0.326}$} & $0.571$ & \textcolor{hred}{$\mathbf{1.000}$} & $0.955$ & \textcolor{hred}{$\mathbf{1.000}$} & $0.148$ & \textcolor{hred}{$\mathbf{0.158}$} & $0.950$ & \textcolor{hred}{$\mathbf{0.984}$} \\
iTransformer & \textcolor{hred}{$\mathbf{0.142}$} & \textcolor{hblue}{$\mathbf{-0.197}$} & \textcolor{hred}{$\mathbf{0.339}$} & $0.579$ & \textcolor{hred}{$\mathbf{0.980}$} & $0.960$ & \textcolor{hred}{$\mathbf{0.998}$} & $0.133$ & \textcolor{hred}{$\mathbf{0.156}$} & $0.962$ & \textcolor{hred}{$\mathbf{0.984}$} \\
\midrule
\textbf{Mean} & $0.110$ & $-0.186$ & $0.296$ & $0.533$ & $0.886$ & $0.915$ & $0.953$ & $0.125$ & $0.144$ & $0.913$ & $0.938$ \\
\bottomrule
\end{NiceTabular}
\end{fittable}
\end{center}
{\scriptsize\setlength{\parindent}{0pt}\setlength{\leftskip}{1.6em}
\textsuperscript{1}A prior explanation cannot be measured at a single forecast step, so $E[h]$
and $\bar{E}$ are the same vector there and every $g$ is exactly zero.\par}
\end{table}

\begin{table}[h]
\caption{\textbf{Explanation performance on the synthetic dataset, per backbone and generator.}
Every cell is the median gap in AUPRC between our matrix and the vector prior works return,
over \syN{} runs. Note that \texttt{sparsenull} is the negative control, so its gap should not
be positive.}
\label{tab:synthby}
\begin{center}
\tiny
\begin{fittable}\begin{tabular}{lcccccc}
\toprule
\textbf{Backbone} & \textbf{lagmix} & \textbf{plantrank} & \textbf{sparseband} & \textbf{sparsenull} & \textbf{sparseshift} & \textbf{sparsesplit} \\
\midrule
Linear & \textcolor{hred}{$\mathbf{+0.813}$} & \textcolor{hred}{$\mathbf{+0.556}$} & \textcolor{hred}{$\mathbf{+0.396}$} & \textcolor{hblue}{$\mathbf{-0.089}$} & \textcolor{hred}{$\mathbf{+0.426}$} & \textcolor{hred}{$\mathbf{+0.262}$} \\
CNN & \textcolor{hred}{$\mathbf{+0.817}$} & \textcolor{hred}{$\mathbf{+0.552}$} & \textcolor{hred}{$\mathbf{+0.408}$} & \textcolor{hblue}{$\mathbf{-0.125}$} & \textcolor{hred}{$\mathbf{+0.437}$} & \textcolor{hred}{$\mathbf{+0.276}$} \\
Transformer & \textcolor{hred}{$\mathbf{+0.818}$} & \textcolor{hred}{$\mathbf{+0.549}$} & \textcolor{hred}{$\mathbf{+0.403}$} & \textcolor{hblue}{$\mathbf{-0.110}$} & \textcolor{hred}{$\mathbf{+0.435}$} & \textcolor{hred}{$\mathbf{+0.278}$} \\
DLinear & \textcolor{hred}{$\mathbf{+0.814}$} & \textcolor{hred}{$\mathbf{+0.556}$} & \textcolor{hred}{$\mathbf{+0.397}$} & \textcolor{hblue}{$\mathbf{-0.062}$} & \textcolor{hred}{$\mathbf{+0.434}$} & \textcolor{hred}{$\mathbf{+0.275}$} \\
TSMixer & \textcolor{hred}{$\mathbf{+0.819}$} & \textcolor{hred}{$\mathbf{+0.543}$} & \textcolor{hred}{$\mathbf{+0.405}$} & \textcolor{hblue}{$\mathbf{-0.109}$} & \textcolor{hred}{$\mathbf{+0.436}$} & \textcolor{hred}{$\mathbf{+0.281}$} \\
LightTS & \textcolor{hred}{$\mathbf{+0.818}$} & \textcolor{hred}{$\mathbf{+0.557}$} & \textcolor{hred}{$\mathbf{+0.397}$} & \textcolor{hblue}{$\mathbf{-0.097}$} & \textcolor{hred}{$\mathbf{+0.435}$} & \textcolor{hred}{$\mathbf{+0.280}$} \\
FiLM & \textcolor{hred}{$\mathbf{+0.819}$} & \textcolor{hred}{$\mathbf{+0.189}$} & \textcolor{hred}{$\mathbf{+0.374}$} & \textcolor{hblue}{$\mathbf{-0.150}$} & \textcolor{hred}{$\mathbf{+0.404}$} & \textcolor{hred}{$\mathbf{+0.271}$} \\
FreTS & \textcolor{hred}{$\mathbf{+0.813}$} & \textcolor{hred}{$\mathbf{+0.551}$} & \textcolor{hred}{$\mathbf{+0.403}$} & \textcolor{hblue}{$\mathbf{-0.013}$} & \textcolor{hred}{$\mathbf{+0.435}$} & \textcolor{hred}{$\mathbf{+0.278}$} \\
TiDE & \textcolor{hred}{$\mathbf{+0.803}$} & \textcolor{hred}{$\mathbf{+0.318}$} & \textcolor{hred}{$\mathbf{+0.393}$} & \textcolor{hblue}{$\mathbf{-0.141}$} & \textcolor{hred}{$\mathbf{+0.384}$} & \textcolor{hred}{$\mathbf{+0.207}$} \\
SCINet & \textcolor{hred}{$\mathbf{+0.812}$} & \textcolor{hred}{$\mathbf{+0.359}$} & \textcolor{hred}{$\mathbf{+0.402}$} & \textcolor{hblue}{$\mathbf{-0.099}$} & \textcolor{hred}{$\mathbf{+0.422}$} & \textcolor{hred}{$\mathbf{+0.278}$} \\
MICN & \textcolor{hred}{$\mathbf{+0.820}$} & \textcolor{hred}{$\mathbf{+0.554}$} & \textcolor{hred}{$\mathbf{+0.397}$} & \textcolor{hblue}{$\mathbf{-0.152}$} & \textcolor{hred}{$\mathbf{+0.440}$} & \textcolor{hred}{$\mathbf{+0.276}$} \\
TimesNet & \textcolor{hred}{$\mathbf{+0.811}$} & \textcolor{hred}{$\mathbf{+0.356}$} & \textcolor{hred}{$\mathbf{+0.399}$} & \textcolor{hblue}{$\mathbf{-0.101}$} & \textcolor{hred}{$\mathbf{+0.413}$} & \textcolor{hred}{$\mathbf{+0.255}$} \\
SegRNN & \textcolor{hred}{$\mathbf{+0.812}$} & \textcolor{hred}{$\mathbf{+0.521}$} & \textcolor{hred}{$\mathbf{+0.525}$} & \textcolor{hblue}{$\mathbf{-0.060}$} & \textcolor{hred}{$\mathbf{+0.437}$} & \textcolor{hred}{$\mathbf{+0.374}$} \\
TimeMixer & \textcolor{hred}{$\mathbf{+0.819}$} & \textcolor{hred}{$\mathbf{+0.311}$} & \textcolor{hred}{$\mathbf{+0.398}$} & \textcolor{hblue}{$\mathbf{-0.141}$} & \textcolor{hred}{$\mathbf{+0.431}$} & \textcolor{hred}{$\mathbf{+0.246}$} \\
MSGNet & \textcolor{hred}{$\mathbf{+0.803}$} & \textcolor{hred}{$\mathbf{+0.373}$} & \textcolor{hred}{$\mathbf{+0.395}$} & \textcolor{hblue}{$\mathbf{-0.084}$} & \textcolor{hred}{$\mathbf{+0.436}$} & \textcolor{hred}{$\mathbf{+0.274}$} \\
Informer & \textcolor{hred}{$\mathbf{+0.258}$} & \textcolor{hred}{$\mathbf{+0.542}$} & \textcolor{hred}{$\mathbf{+0.371}$} & \textcolor{hblue}{$\mathbf{-0.137}$} & \textcolor{hred}{$\mathbf{+0.128}$} & \textcolor{hred}{$\mathbf{+0.233}$} \\
Autoformer & $-0.006$ & \textcolor{hred}{$\mathbf{+0.004}$} & $-0.000$ & \textcolor{hblue}{$\mathbf{-0.084}$} & $-0.080$ & $-0.097$ \\
FEDformer & \textcolor{hred}{$\mathbf{+0.425}$} & \textcolor{hred}{$\mathbf{+0.044}$} & $-0.037$ & \textcolor{hblue}{$\mathbf{-0.165}$} & \textcolor{hred}{$\mathbf{+0.034}$} & $-0.061$ \\
Pyraformer & \textcolor{hred}{$\mathbf{+0.810}$} & \textcolor{hred}{$\mathbf{+0.480}$} & \textcolor{hred}{$\mathbf{+0.286}$} & \textcolor{hblue}{$\mathbf{-0.131}$} & \textcolor{hred}{$\mathbf{+0.310}$} & \textcolor{hred}{$\mathbf{+0.227}$} \\
Reformer & \textcolor{hred}{$\mathbf{+0.125}$} & \textcolor{hred}{$\mathbf{+0.511}$} & \textcolor{hred}{$\mathbf{+0.004}$} & \textcolor{hblue}{$\mathbf{-0.216}$} & $-0.042$ & $-0.074$ \\
Nonstat. Trans. & \textcolor{hred}{$\mathbf{+0.416}$} & \textcolor{hred}{$\mathbf{+0.264}$} & \textcolor{hred}{$\mathbf{+0.361}$} & \textcolor{hblue}{$\mathbf{-0.103}$} & \textcolor{hred}{$\mathbf{+0.380}$} & \textcolor{hred}{$\mathbf{+0.218}$} \\
Crossformer & \textcolor{hred}{$\mathbf{+0.819}$} & \textcolor{hred}{$\mathbf{+0.553}$} & \textcolor{hred}{$\mathbf{+0.409}$} & \textcolor{hblue}{$\mathbf{-0.028}$} & \textcolor{hred}{$\mathbf{+0.446}$} & \textcolor{hred}{$\mathbf{+0.279}$} \\
PatchTST & \textcolor{hred}{$\mathbf{+0.816}$} & \textcolor{hred}{$\mathbf{+0.330}$} & \textcolor{hred}{$\mathbf{+0.407}$} & \textcolor{hblue}{$\mathbf{-0.071}$} & \textcolor{hred}{$\mathbf{+0.439}$} & \textcolor{hred}{$\mathbf{+0.281}$} \\
TimeXer & \textcolor{hred}{$\mathbf{+0.814}$} & \textcolor{hred}{$\mathbf{+0.360}$} & \textcolor{hred}{$\mathbf{+0.403}$} & \textcolor{hblue}{$\mathbf{-0.075}$} & \textcolor{hred}{$\mathbf{+0.439}$} & \textcolor{hred}{$\mathbf{+0.287}$} \\
MultiPatchFormer & \textcolor{hred}{$\mathbf{+0.809}$} & \textcolor{hred}{$\mathbf{+0.426}$} & \textcolor{hred}{$\mathbf{+0.408}$} & \textcolor{hblue}{$\mathbf{-0.048}$} & \textcolor{hred}{$\mathbf{+0.430}$} & \textcolor{hred}{$\mathbf{+0.286}$} \\
iTransformer & \textcolor{hred}{$\mathbf{+0.816}$} & \textcolor{hred}{$\mathbf{+0.368}$} & \textcolor{hred}{$\mathbf{+0.400}$} & \textcolor{hblue}{$\mathbf{-0.088}$} & \textcolor{hred}{$\mathbf{+0.420}$} & \textcolor{hred}{$\mathbf{+0.285}$} \\
\midrule
\textbf{Mean} & \textcolor{hred}{$\mathbf{+0.704}$} & \textcolor{hred}{$\mathbf{+0.413}$} & \textcolor{hred}{$\mathbf{+0.350}$} & \textcolor{hblue}{$\mathbf{-0.103}$} & \textcolor{hred}{$\mathbf{+0.358}$} & \textcolor{hred}{$\mathbf{+0.230}$} \\
\bottomrule
\end{tabular}
\end{fittable}
\end{center}
\end{table}

\clearpage
\begin{wrapfigure}{r}{0.5\linewidth}
\vspace{-\intextsep}\vspace{10.6pt}
\centering
\includegraphics[width=\linewidth]{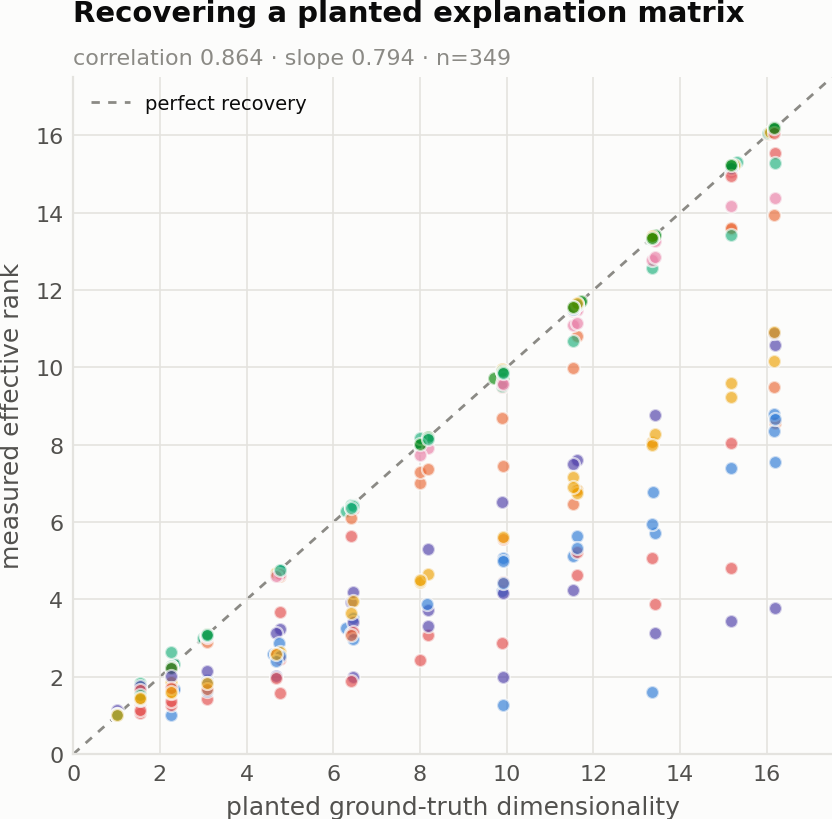}
\caption{\textbf{Recovering a planted rank.} The measured effective rank against the planted one
over \calibN{} runs, where color marks the backbone. The measure tracks the plant on every
backbone.}
\label{fig:calibration}
\end{wrapfigure}

\noindent\textbf{Instrument calibration.}
To ask whether the effective rank we report is a property of the matrix rather than of the
estimator that fills it, we calibrate the measure against a known answer.
We draw the target from a matrix $A$ of planted effective rank with the \texttt{plantrank}
generator of Table~\ref{tab:gens}, where the true Jacobian is $A$ itself and the answer is fixed
before any measurement.
The planted rank is swept from $1$ to $\calibRankMax{}$ over \calibModels{} backbones, and every
trained model is then read with our own estimator.
As shown in Figure~\ref{fig:calibration}, the measured rank rises with the planted one on every
backbone, exactly on some of them and at a lower slope on the rest.
The measure therefore reads the rank the matrix carries rather than one the estimator imposes,
which is what Appendix~\ref{sec:rankdetail} relies on when it reads the rank as a predictor of
the gain.

\begin{figure}[h]
\centering
\includegraphics[width=\linewidth]{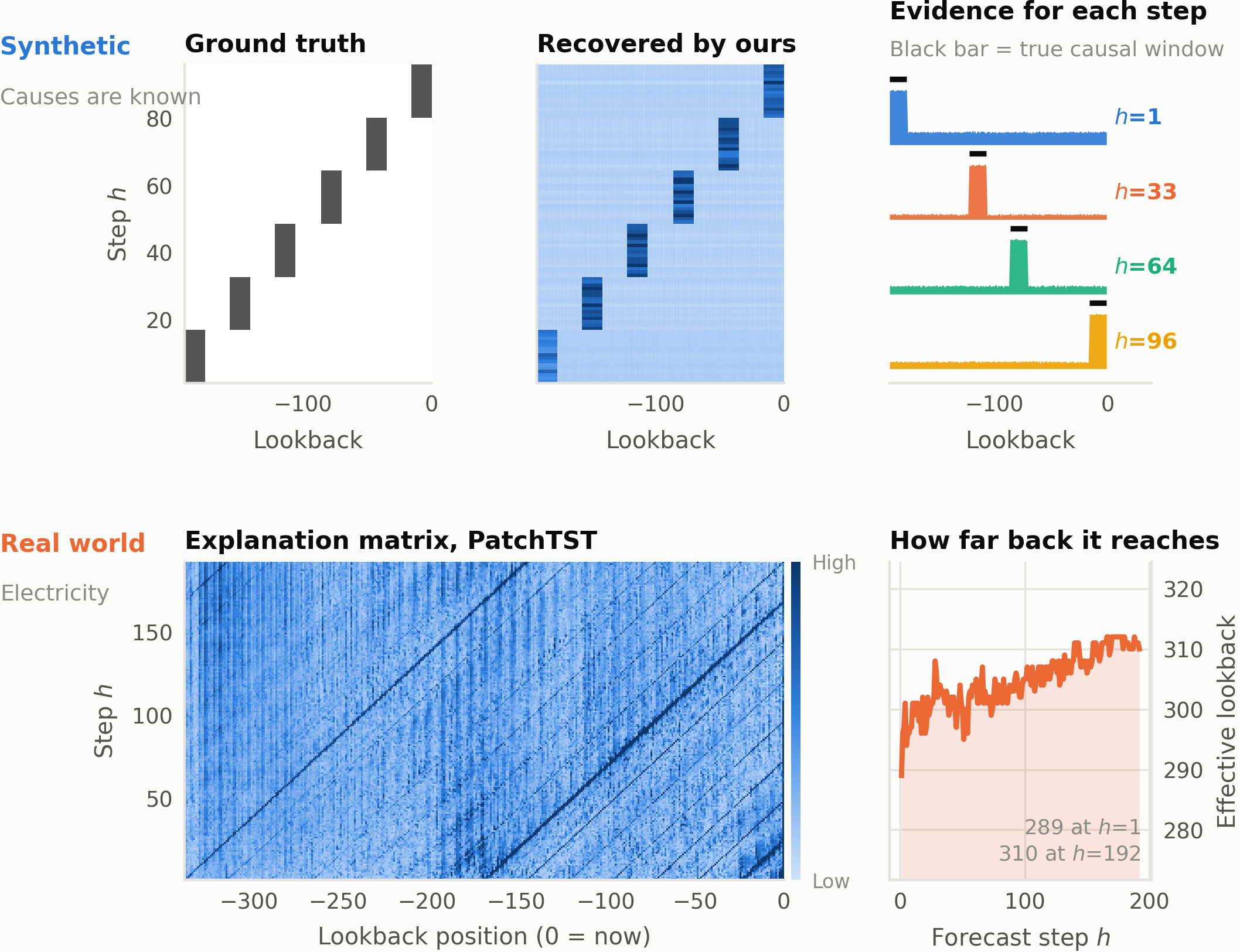}
\caption{\textbf{HRX against a known answer.} \textrm{[Top]} On a constructed generator, the recovered
matrix reproduces the ground-truth staircase, where black bars mark the true causal windows.
\textrm{[Bottom]} The same procedure on Electricity with PatchTST.}
\label{fig:toy}
\end{figure}

\noindent\textbf{Checking the matrix against a known answer.}
\label{sec:toy}
As shown in Figure~\ref{fig:toy}, we place a synthetic generator whose causal windows we control
beside a real benchmark, and the recovered matrix reproduces the staircase on the synthetic
side, which is the check Section~\ref{sec:methods} could not run.

\clearpage

\clearpage
\section{Results across Architectures and Benchmarks}
\label{sec:detailed}
Table~\ref{tab:models} and Table~\ref{tab:datasets} are the numbers behind the two panels of
Figure~\ref{fig:forest}, and they report all three quantities of Section~\ref{sec:protocol}
side by side, where $g_{\text{shuffled}}$ is recovered from
$g_{\text{margin}} = g_{\text{own}} - g_{\text{shuffled}}$.
Note that the ideal direction holds on every benchmark except PEMS04, which does not reach
significance, and that three backbones sit at or below zero.
Pyraformer is the one whose interval falls entirely below zero, while FEDformer and SegRNN have
intervals that contain it, and none of the three carries a gain of the size the other backbones
report.
Note also that a backbone is listed only when it has at least \nMinRuns{} completed runs, so
that a handful does not sit beside thousands.

\begin{table}[h]
\caption{\textbf{The three quantities of the protocol across architectures.} Aggregated over the
\nDatasets{} benchmarks and all swept settings. Every value is a median, and a check mark means
the bootstrap $95\%$ interval of $g_{\text{own}}$ excludes zero.}
\label{tab:models}
\begin{center}
\small
\begin{fittable}\begin{tabular}{lcccccc}
\toprule
\textbf{Backbone} & \textbf{$g_{\text{own}}$} & \textbf{95\% CI} & \textbf{$g_{\text{shuffled}}$} & \textbf{$g_{\text{margin}}$} & \textbf{$n$} & \textbf{Significant} \\
\midrule
Linear & \textcolor{hred}{$\mathbf{+0.199}$} & $[+0.192, +0.206]$ & \textcolor{hblue}{$\mathbf{-0.024}$} & \textcolor{hred}{$\mathbf{+0.223}$} & 3{,}513 & \checkmark \\
CNN & \textcolor{hred}{$\mathbf{+0.134}$} & $[+0.127, +0.140]$ & \textcolor{hblue}{$\mathbf{-0.084}$} & \textcolor{hred}{$\mathbf{+0.218}$} & 3{,}513 & \checkmark \\
Transformer & \textcolor{hred}{$\mathbf{+0.054}$} & $[+0.050, +0.061]$ & \textcolor{hblue}{$\mathbf{-0.071}$} & \textcolor{hred}{$\mathbf{+0.125}$} & 3{,}667 & \checkmark \\
DLinear & \textcolor{hred}{$\mathbf{+0.172}$} & $[+0.164, +0.181]$ & \textcolor{hblue}{$\mathbf{-0.036}$} & \textcolor{hred}{$\mathbf{+0.208}$} & 3{,}583 & \checkmark \\
TSMixer & \textcolor{hred}{$\mathbf{+0.169}$} & $[+0.157, +0.179]$ & \textcolor{hblue}{$\mathbf{-0.045}$} & \textcolor{hred}{$\mathbf{+0.214}$} & 662 & \checkmark \\
LightTS & \textcolor{hred}{$\mathbf{+0.229}$} & $[+0.213, +0.283]$ & \textcolor{hblue}{$\mathbf{-0.109}$} & \textcolor{hred}{$\mathbf{+0.338}$} & 284 & \checkmark \\
FiLM & \textcolor{hred}{$\mathbf{+0.256}$} & $[+0.221, +0.280]$ & \textcolor{hblue}{$\mathbf{-0.099}$} & \textcolor{hred}{$\mathbf{+0.355}$} & 284 & \checkmark \\
FreTS & \textcolor{hred}{$\mathbf{+0.101}$} & $[+0.086, +0.128]$ & \textcolor{hblue}{$\mathbf{-0.054}$} & \textcolor{hred}{$\mathbf{+0.155}$} & 284 & \checkmark \\
TiDE & \textcolor{hred}{$\mathbf{+0.264}$} & $[+0.246, +0.283]$ & \textcolor{hblue}{$\mathbf{-0.133}$} & \textcolor{hred}{$\mathbf{+0.397}$} & 578 & \checkmark \\
SCINet & \textcolor{hred}{$\mathbf{+0.238}$} & $[+0.216, +0.271]$ & \textcolor{hblue}{$\mathbf{-0.153}$} & \textcolor{hred}{$\mathbf{+0.391}$} & 578 & \checkmark \\
MICN & \textcolor{hred}{$\mathbf{+0.051}$} & $[+0.031, +0.071]$ & \textcolor{hblue}{$\mathbf{-0.037}$} & \textcolor{hred}{$\mathbf{+0.088}$} & 284 & \checkmark \\
TimesNet & \textcolor{hred}{$\mathbf{+0.224}$} & $[+0.197, +0.244]$ & \textcolor{hblue}{$\mathbf{-0.074}$} & \textcolor{hred}{$\mathbf{+0.298}$} & 654 & \checkmark \\
SegRNN & \textcolor{hred}{$\mathbf{+0.004}$} & $[-0.019, +0.012]$ & \textcolor{hblue}{$\mathbf{-0.076}$} & \textcolor{hred}{$\mathbf{+0.080}$} & 284 & \textrm{--} \\
TimeMixer & \textcolor{hred}{$\mathbf{+0.115}$} & $[+0.102, +0.130]$ & \textcolor{hblue}{$\mathbf{-0.064}$} & \textcolor{hred}{$\mathbf{+0.179}$} & 284 & \checkmark \\
MSGNet & \textcolor{hred}{$\mathbf{+0.071}$} & $[+0.052, +0.088]$ & \textcolor{hblue}{$\mathbf{-0.034}$} & \textcolor{hred}{$\mathbf{+0.105}$} & 282 & \checkmark \\
Informer & \textcolor{hred}{$\mathbf{+0.018}$} & $[+0.011, +0.035]$ & $+0.006$ & \textcolor{hred}{$\mathbf{+0.012}$} & 272 & \checkmark \\
Autoformer & \textcolor{hred}{$\mathbf{+0.098}$} & $[+0.070, +0.130]$ & \textcolor{hblue}{$\mathbf{-0.011}$} & \textcolor{hred}{$\mathbf{+0.109}$} & 284 & \checkmark \\
FEDformer & $-0.000$ & $[-0.018, +0.015]$ & \textcolor{hblue}{$\mathbf{-0.120}$} & \textcolor{hred}{$\mathbf{+0.120}$} & 283 & \textrm{--} \\
Pyraformer & $-0.003$ & $[-0.011, -0.000]$ & \textcolor{hblue}{$\mathbf{-0.048}$} & \textcolor{hred}{$\mathbf{+0.045}$} & 284 & \textrm{--} \\
Reformer & \textcolor{hred}{$\mathbf{+0.079}$} & $[+0.056, +0.102]$ & \textcolor{hblue}{$\mathbf{-0.137}$} & \textcolor{hred}{$\mathbf{+0.216}$} & 284 & \checkmark \\
Nonstat. Trans. & \textcolor{hred}{$\mathbf{+0.057}$} & $[+0.041, +0.078]$ & \textcolor{hblue}{$\mathbf{-0.027}$} & \textcolor{hred}{$\mathbf{+0.084}$} & 284 & \checkmark \\
Crossformer & \textcolor{hred}{$\mathbf{+0.055}$} & $[+0.040, +0.073]$ & \textcolor{hblue}{$\mathbf{-0.061}$} & \textcolor{hred}{$\mathbf{+0.116}$} & 284 & \checkmark \\
PatchTST & \textcolor{hred}{$\mathbf{+0.204}$} & $[+0.197, +0.212]$ & \textcolor{hblue}{$\mathbf{-0.030}$} & \textcolor{hred}{$\mathbf{+0.234}$} & 3{,}602 & \checkmark \\
ITransformer & \textcolor{hred}{$\mathbf{+0.279}$} & $[+0.257, +0.294]$ & \textcolor{hblue}{$\mathbf{-0.029}$} & \textcolor{hred}{$\mathbf{+0.308}$} & 662 & \checkmark \\
TimeXer & \textcolor{hred}{$\mathbf{+0.183}$} & $[+0.157, +0.223]$ & \textcolor{hblue}{$\mathbf{-0.044}$} & \textcolor{hred}{$\mathbf{+0.227}$} & 284 & \checkmark \\
MultiPatchFormer & \textcolor{hred}{$\mathbf{+0.178}$} & $[+0.144, +0.204]$ & \textcolor{hblue}{$\mathbf{-0.014}$} & \textcolor{hred}{$\mathbf{+0.192}$} & 260 & \checkmark \\
\bottomrule
\end{tabular}
\end{fittable}
\end{center}
\end{table}
\clearpage

\begin{table}[h]
\caption{\textbf{The three quantities of the protocol across benchmarks.} Aggregated over the
\nModels{} backbones in the univariate setting. Every value is a median, and a check mark means
the bootstrap $95\%$ interval of $g_{\text{own}}$ excludes zero. Note that the ranking is the normalized
one, and that the raw gain would put the two highest-variance benchmarks on top for unrelated
reasons.}
\label{tab:datasets}
\begin{center}
\begin{fittable}\begin{tabular}{lcccccc}
\toprule
\textbf{Benchmark} & \textbf{$g_{\text{own}}$} & \textbf{95\% CI} & \textbf{$g_{\text{shuffled}}$} & \textbf{$g_{\text{margin}}$} & \textbf{$n$} & \textbf{Significant} \\
\midrule
Finance\_ret & \textcolor{hred}{$\mathbf{+0.340}$} & $[+0.325, +0.356]$ & \textcolor{hblue}{$\mathbf{-0.018}$} & \textcolor{hred}{$\mathbf{+0.358}$} & 1{,}999 & \checkmark \\
Electricity & \textcolor{hred}{$\mathbf{+0.313}$} & $[+0.305, +0.320]$ & \textcolor{hblue}{$\mathbf{-0.079}$} & \textcolor{hred}{$\mathbf{+0.392}$} & 2{,}681 & \checkmark \\
ETTh2 & \textcolor{hred}{$\mathbf{+0.280}$} & $[+0.273, +0.286]$ & \textcolor{hblue}{$\mathbf{-0.010}$} & \textcolor{hred}{$\mathbf{+0.290}$} & 2{,}278 & \checkmark \\
ETTh1 & \textcolor{hred}{$\mathbf{+0.195}$} & $[+0.191, +0.199]$ & \textcolor{hblue}{$\mathbf{-0.047}$} & \textcolor{hred}{$\mathbf{+0.242}$} & 2{,}736 & \checkmark \\
ETTm2 & \textcolor{hred}{$\mathbf{+0.143}$} & $[+0.126, +0.158]$ & \textcolor{hblue}{$\mathbf{-0.018}$} & \textcolor{hred}{$\mathbf{+0.161}$} & 2{,}274 & \checkmark \\
Traffic & \textcolor{hred}{$\mathbf{+0.138}$} & $[+0.128, +0.151]$ & \textcolor{hblue}{$\mathbf{-0.050}$} & \textcolor{hred}{$\mathbf{+0.188}$} & 820 & \checkmark \\
Solar & \textcolor{hred}{$\mathbf{+0.112}$} & $[+0.102, +0.125]$ & \textcolor{hblue}{$\mathbf{-0.092}$} & \textcolor{hred}{$\mathbf{+0.204}$} & 681 & \checkmark \\
ETTm1 & \textcolor{hred}{$\mathbf{+0.079}$} & $[+0.073, +0.084]$ & \textcolor{hblue}{$\mathbf{-0.005}$} & \textcolor{hred}{$\mathbf{+0.084}$} & 2{,}326 & \checkmark \\
Finance & \textcolor{hred}{$\mathbf{+0.061}$} & $[+0.048, +0.072]$ & \textcolor{hblue}{$\mathbf{-0.062}$} & \textcolor{hred}{$\mathbf{+0.123}$} & 1{,}994 & \checkmark \\
Weather & \textcolor{hred}{$\mathbf{+0.056}$} & $[+0.053, +0.059]$ & \textcolor{hblue}{$\mathbf{-0.126}$} & \textcolor{hred}{$\mathbf{+0.182}$} & 2{,}730 & \checkmark \\
PEMS03 & \textcolor{hred}{$\mathbf{+0.055}$} & $[+0.019, +0.100]$ & \textcolor{hblue}{$\mathbf{-0.098}$} & \textcolor{hred}{$\mathbf{+0.153}$} & 682 & \checkmark \\
Exchange & \textcolor{hred}{$\mathbf{+0.045}$} & $[+0.039, +0.050]$ & \textcolor{hblue}{$\mathbf{-0.076}$} & \textcolor{hred}{$\mathbf{+0.121}$} & 2{,}278 & \checkmark \\
PEMS07 & \textcolor{hred}{$\mathbf{+0.037}$} & $[+0.004, +0.078]$ & \textcolor{hblue}{$\mathbf{-0.074}$} & \textcolor{hred}{$\mathbf{+0.111}$} & 679 & \checkmark \\
PEMS04 & \textcolor{hred}{$\mathbf{+0.007}$} & $[-0.004, +0.014]$ & \textcolor{hblue}{$\mathbf{-0.061}$} & \textcolor{hred}{$\mathbf{+0.068}$} & 679 & \textrm{--} \\
PEMS08 & \textcolor{hred}{$\mathbf{+0.005}$} & $[+0.003, +0.008]$ & \textcolor{hblue}{$\mathbf{-0.016}$} & \textcolor{hred}{$\mathbf{+0.021}$} & 680 & \checkmark \\
\bottomrule
\end{tabular}
\end{fittable}
\end{center}
\end{table}

\vspace{30pt}
\section{Results by Domain}
\label{sec:domdetail}

\noindent\textbf{The two naive baselines.}
The baselines are predicting the standardized target as zero and extending the last observation,
and a run counts as skilled only when its test error beats both.
Note that to our knowledge no prior TS explanation benchmark reports a forecasting baseline
beside its faithfulness score, which is why Section~\ref{sec:domains} reports the split rather
than a single number.

\noindent\textbf{What actually differs between the domains.}
The explanation matrices are of similar complexity in both, with an effective rank of
$\domGeneralRank$ in the general domain and $\domFinancialRank$ in the financial one.
What changes is which backbone exploits the axis rather than the shape of the explanation, as
the vanilla Transformer moves from one of the weakest in the general domain to one of the
strongest in the financial one.
Which backbone exploits the axis is therefore a property of the data as much as of the
architecture, and our ranking should not be read as a general one.

\clearpage
\section{Qualitative Panels in Full}
\label{sec:qual}
\begin{figure}[h]
\centering
\includegraphics[width=\linewidth]{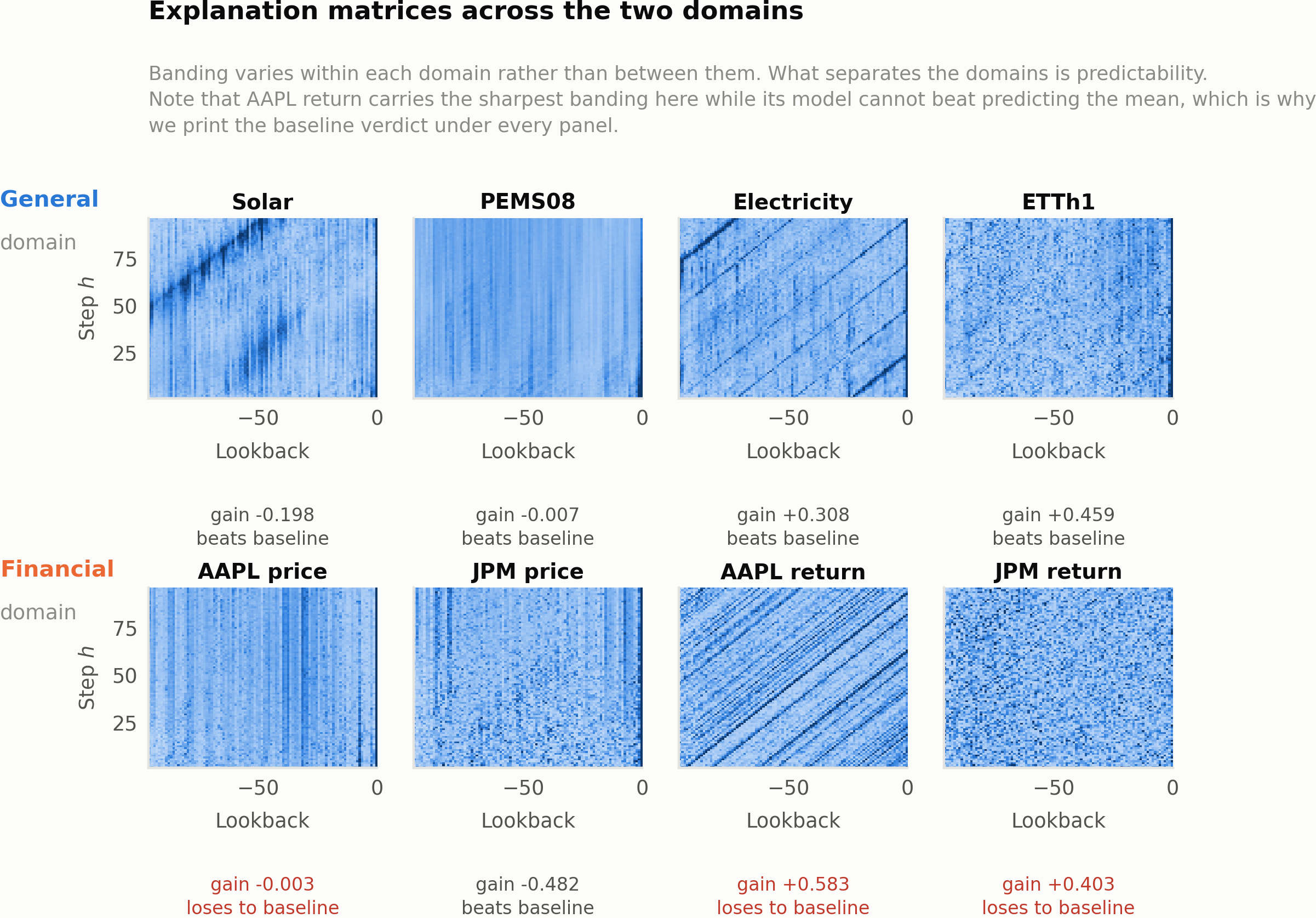}
\caption{\textbf{Explanation matrices across the two domains.} The naive-baseline verdict is under each
panel. Banding varies within each domain rather than between them, and the sharpest
banding here comes from a model that cannot forecast.}
\label{fig:domains}
\end{figure}

\begin{figure}[h]
\centering
\includegraphics[width=\linewidth]{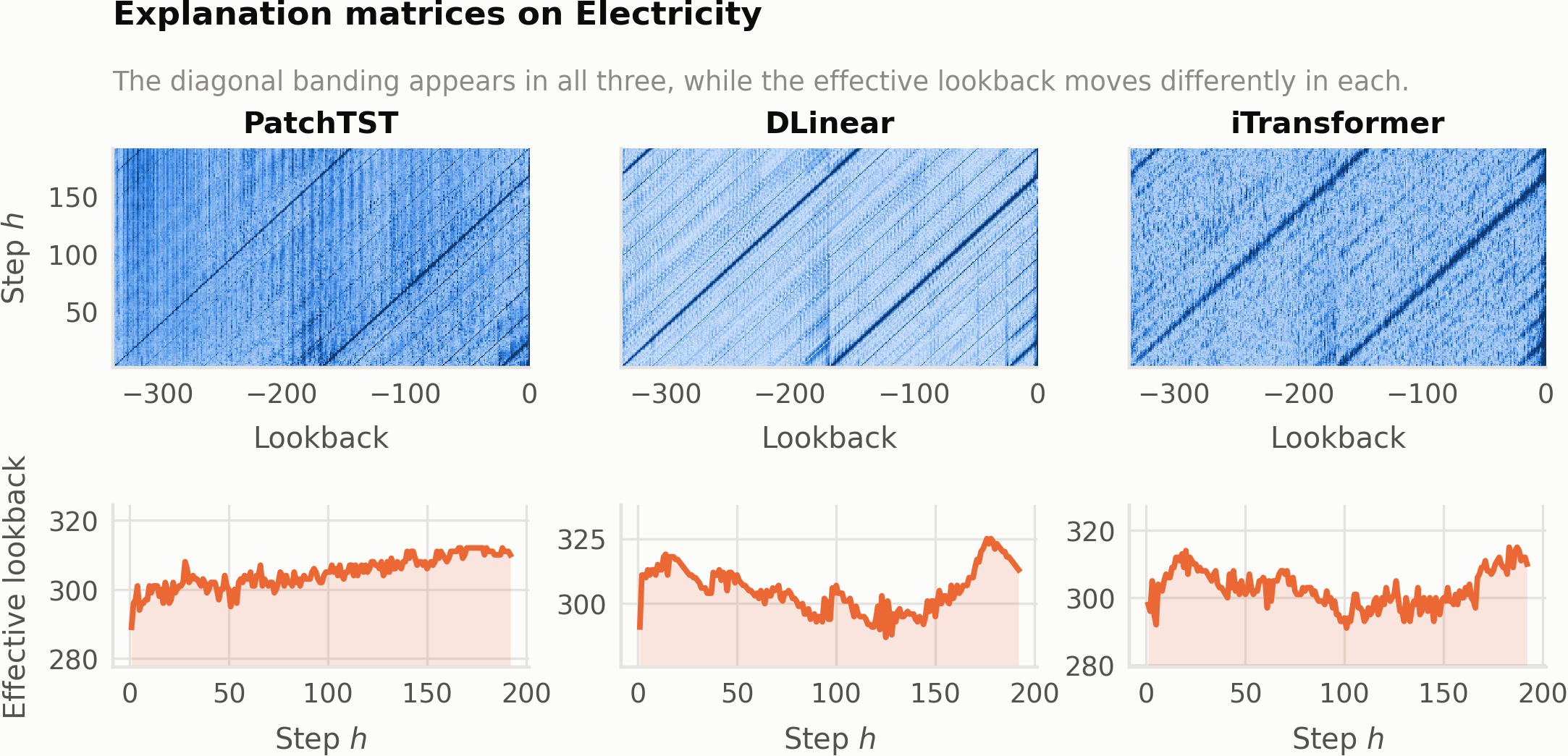}
\caption{\textbf{Explanation matrices for three architectures on Electricity.} Underneath each
is the effective lookback of every row, which is how many recent positions a row needs to reach
$90\%$ of its total. The banding appears in all three, while the effective lookback does not.}
\label{fig:application}
\end{figure}

Figure~\ref{fig:hero} shows five panels in one row for space. Figure~\ref{fig:domains} gives
the full domain comparison with the naive-baseline verdict under each panel, and
Figure~\ref{fig:application} gives the three backbones together with the effective lookback of
every row.

\clearpage
\section{Further Analyses}
\label{sec:further}

\subsection{Swept Axes}
\label{sec:axes}

To ask whether the horizon axis behaves the same at every setting, we contrast the endpoints of
the three swept settings, holding backbone, dataset, seed, and the other two settings fixed so
that the comparison is within-block.
The pairing is necessary because the set of backbones that completed at the longest horizon
differs from the set that completed at the shortest.
As shown in Table~\ref{tab:axes}, the gain rises with both the forecast length and the lookback
length, while depth moves in the opposite direction by an order of magnitude less.
We attribute this to the shape of the problem rather than to capacity, since a longer forecast
spans more of the seasonal structure and a longer lookback offers more evidence to choose among,
while adding layers offers neither.
Note that only the endpoint contrasts are significant, so we read the pattern as \textit{a
trend rather than a law}.

\begin{table}[h]
\caption{\textbf{Paired change in the gain along the three swept settings.} Each contrast holds
backbone, dataset, seed and the other two axes fixed. A plain average per level would be
misleading, since the backbones that completed at $H=720$ differ from those at $H=24$.}
\label{tab:axes}
\begin{center}
\begin{fittable}\begin{tabular}{lcccc}
\toprule
\textbf{Setting} & \textbf{Range} & \textbf{Paired change} & \textbf{95\% CI} & \textbf{Blocks} \\
\midrule
Forecast length $H$ & 24 $\to$ 720 & \textcolor{hred}{$\mathbf{+0.260}$} & $[+0.118, +0.402]$ & 100 \\
Lookback length $L$ & 48 $\to$ 512 & \textcolor{hred}{$\mathbf{+0.408}$} & $[+0.017, +0.800]$ & 20 \\
Depth (layers) & 1 $\to$ 6 & $-0.020$ & $[-0.039, -0.001]$ & 208 \\
\bottomrule
\end{tabular}
\end{fittable}
\end{center}
\end{table}

\begin{figure}[h]
\centering
\includegraphics[width=\linewidth]{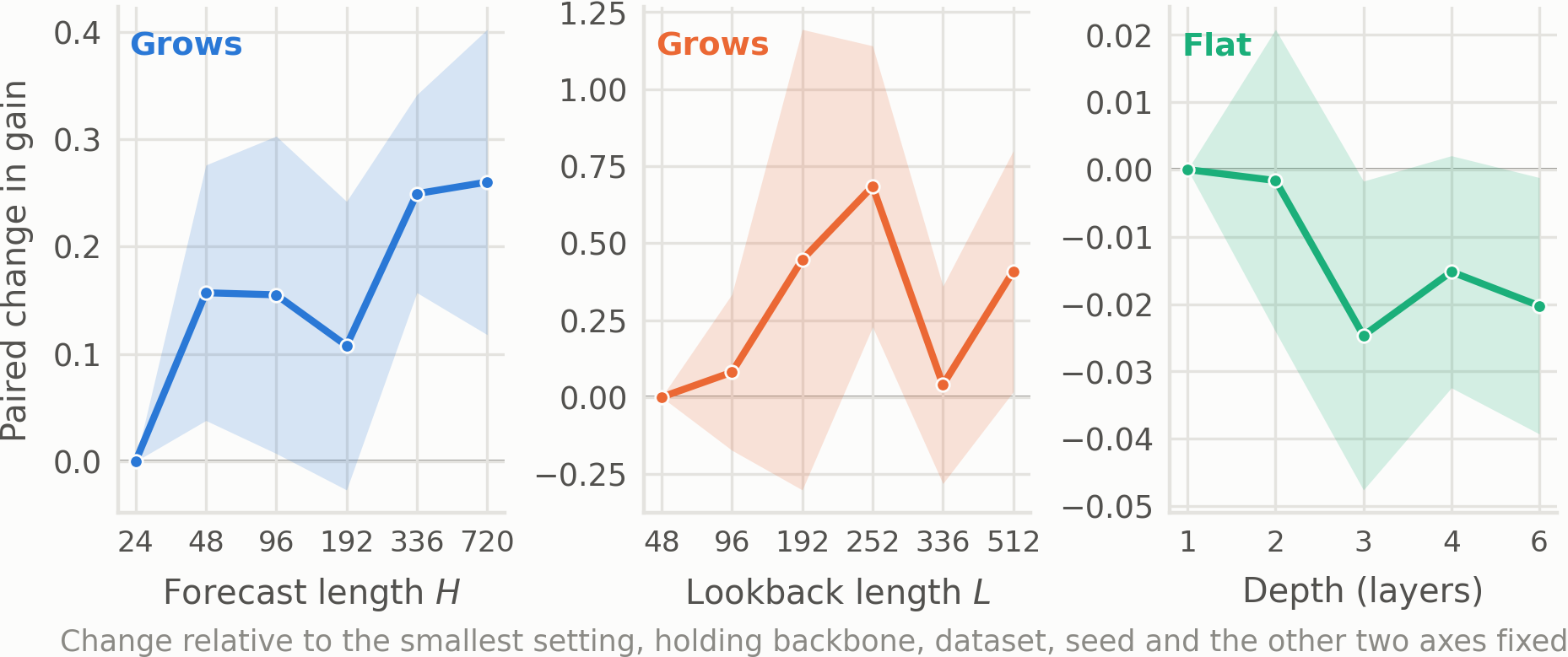}
\caption{\textbf{Paired change in the gain along the three swept settings.} Each point is the average
within-block difference against the smallest setting.}
\label{fig:axes}
\end{figure}

Figure~\ref{fig:axes} shows the same three contrasts as Table~\ref{tab:axes} at every
intermediate level, which makes clear that the increase is not monotone step by step.

\clearpage
\subsection{Cost of Each Estimator}
\label{sec:cost}

Table~\ref{tab:methods} in the body reports what the horizon axis adds to each estimator and
says nothing about what that costs.
As shown in Table~\ref{tab:methods_cost}, the mask methods are two orders of magnitude more
expensive than a plain gradient, because the axis makes each of them solve a separate
optimization problem for every forecast step.
Note that our default estimator is not the cheapest either, since filling the matrix takes more
than one backward pass, which Appendix~\ref{sec:batching} keeps under $H$.

\begin{table}[h]
\caption{\textbf{Cost of each estimator.} Wall-clock time relative to one forward pass of the
same backbone, measured on the same hardware for every method.}
\label{tab:methods_cost}
\begin{center}
\begin{fittable}\begin{tabular}{lc}
\toprule
\textbf{Estimator} & \textbf{Cost} \\
\midrule
Saliency & $3.1\times$ \\
Gradient $\times$ input & $3.1\times$ \\
Integrated gradients & $98\times$ \\
Occlusion & $14\times$ \\
Dynamask & $592\times$ \\
ExtremalMask & $643\times$ \\
ContraLSP & $675\times$ \\
Input gradient & $18\times$ \\
\bottomrule
\end{tabular}
\end{fittable}
\end{center}
\end{table}

\vspace{20pt}
\subsection{The Estimators as Plain Attribution Methods}
\label{sec:plainmethods}

Table~\ref{tab:methods} in the body asks what the horizon axis adds to each estimator, which
is a question about the axis rather than about the estimators themselves.
To ask the complementary question, we score the same estimators without the axis, in the marginal
form each one was published in, under the standard deletion gap between the most-relevant-first
and least-relevant-first orders.
As shown in Table~\ref{tab:methods_std}, our default input gradient is competitive but not the
best.
This is the position we take throughout, since the contribution is the axis and the estimator
that fills it is a free choice.

\begin{table}[h]
\caption{\textbf{The same estimators without the horizon axis.} Each is scored in its published
marginal form under the standard deletion gap, so the ordering here says nothing about the
horizon axis.}
\label{tab:methods_std}
\begin{center}
\begin{fittable}\begin{tabular}{lccc}
\toprule
\textbf{Estimator} & \textbf{Deletion gap} & \textbf{95\% CI} & \textbf{$n$} \\
\midrule
TSR & \textcolor{hred}{$\mathbf{+0.747}$} & $[+0.745, +0.749]$ & 15{,}898 \\
Integrated gradients + HRX & \textcolor{hred}{$\mathbf{+0.694}$} & $[+0.691, +0.697]$ & 15{,}898 \\
Dynamask + HRX & \textcolor{hred}{$\mathbf{+0.674}$} & $[+0.672, +0.677]$ & 15{,}898 \\
Gradient $\times$ input + HRX & \textcolor{hred}{$\mathbf{+0.634}$} & $[+0.628, +0.640]$ & 15{,}898 \\
\textbf{HRX (Ours)} & \textcolor{hred}{$\mathbf{+0.630}$} & $[+0.623, +0.635]$ & 15{,}668 \\
ContraLSP + HRX & \textcolor{hred}{$\mathbf{+0.588}$} & $[+0.582, +0.594]$ & 15{,}888 \\
Saliency + HRX & \textcolor{hred}{$\mathbf{+0.576}$} & $[+0.569, +0.586]$ & 15{,}898 \\
ExtremalMask + HRX & \textcolor{hred}{$\mathbf{+0.549}$} & $[+0.542, +0.556]$ & 15{,}889 \\
Occlusion + HRX & \textcolor{hred}{$\mathbf{+0.458}$} & $[+0.453, +0.464]$ & 15{,}898 \\
\bottomrule
\end{tabular}
\end{fittable}
\end{center}
\end{table}

\clearpage
\section{Limitations in Detail}
\label{sec:limdetail}

\noindent\textbf{We cannot predict the effect size.}
A practitioner cannot tell in advance how much the axis will buy on a new series.
We checked spectral concentration, autocorrelation, forecast-target correlation across steps, and
naive-forecast difficulty, and no candidate reached $\lvert r \rvert = \limMaxCorr{}$ on real
data.
An early analysis suggested that the number of stacked periods drives the gain, but that
generator scaled the target variance with the number of periods.
The correlation is \limPeriodCorr{} once the scale is fixed, so we report the failure rather
than the artifact.

\noindent\textbf{We found no downstream utility.}
None of the uses we tried turned the matrix into a better forecast or a better diagnosis.
Neither explanation-guided input selection nor horizon-specific pruning gave a gain
distinguishable from zero, which means HRX is a \textit{measurement tool} and nothing more.

\noindent\textbf{Two properties of our estimator.}
Our default estimator inherits the known weaknesses of input gradients, and our evidence on real
data is faithfulness-based, which is why we anchor our method on constructed labels first.
The effective rank reads $\lvert E \rvert$ and therefore measures how many components the
magnitude map is built from, not how many the signed Jacobian is.
Structure in the signs alone is invisible to it, which is why the calibration of
Section~\ref{sec:sanity} plants the rank in the magnitudes.

\vspace{30pt}
\section{The Rank of the Explanation Matrix}
\label{sec:rankdetail}

\noindent\textbf{What the effective rank counts.}
Eq.~\ref{eq:rank} is the participation ratio of the singular values of $E$, which counts how many
directions carry the magnitude of the matrix rather than how many are nonzero.
A matrix whose rows are positive multiples of one vector has a participation ratio of one, and
every quantity of Table~\ref{tab:protocol} is then pinned to zero, because the deletion order of
Section~\ref{sec:protocol} depends on a row only through the order it induces.
We compute the ratio on $\lvert E \rvert$ for the same reason the protocol does, which is that
the order a row induces does not see the sign.
Structure that lives in the signs alone is therefore invisible to the ratio, which is why the
synthetic generators of Section~\ref{sec:sanity} plant it in the magnitudes.

\noindent\textbf{Why this measure and not the entropy-based one.}
Both Eq.~\ref{eq:rank} and the entropy-based effective rank \citep{roy2007erank} are
exponentials of a
R\'enyi entropy of a normalized singular value distribution, and they differ in two places.
Theirs takes $p_k = \sigma_k / \lVert \sigma \rVert_1$ and the Shannon entropy, that is
R\'enyi order one, giving $\exp(-\sum_k p_k \log p_k)$.
Ours takes $p_k = \sigma_k^2 / \sum_j \sigma_j^2$ and R\'enyi order two, giving
$1 / \sum_k p_k^2$, which is Eq.~\ref{eq:rank} and the quantity reported as a dimension in
prior work \citep{litwinkumar2017dim}.
We use order two because $\sigma_k^2$ is the magnitude a direction carries, so the criterion
says how much of $E$ a direction accounts for rather than how probable it is.
Order two is also less sensitive to the long tail of small singular values, which a matrix of
this size always has, and Jensen's inequality makes it the smaller of the two, so the criterion
never overstates how many directions are in use.

\noindent\textbf{The truncation sweep.}
Table~\ref{tab:lowrank} truncates $E$ to its leading directions and scores every truncation under
the identical protocol, and the rank-one row is the analytic check just described.
Note that the deletion in this sweep is paired, which means every truncation is scored against
the same corruption pattern rather than a fresh draw, since an unpaired comparison puts the
Monte Carlo spread of the corruption into the difference between two truncations.
Without the pairing the rank-one row reads away from its analytic value of zero at the scale of
the spread.
The same pairing is used for the distance sweep of Appendix~\ref{sec:offsetdetail} and nowhere
else.

\noindent\textbf{Raw and normalized quantities.}
Every quantity in the paper is divided by the forecast variance $\mathrm{Var}(\hat{y})$.
That variance is positive on every run, so the normalization cannot change the sign of a
quantity and the fraction of runs in the ideal direction is identical with or without it.
It does change how the benchmarks rank against each other, since a benchmark whose forecasts
vary more carries a larger raw gain at the same normalized gain.
We report the normalized reading throughout because the question is whether the horizon axis
carries information, which the forecast scale should not decide.
Table~\ref{tab:datasets} is the ordering under the normalized reading, and its caption names the
two benchmarks that the raw reading would move to the top.

\clearpage
\section{Distance Between Forecast Steps}
\label{sec:offsetdetail}

\noindent\textbf{Why a distance is needed.}
The protocol of Section~\ref{sec:protocol} draws a permutation $\pi$ uniformly over the horizon,
which validates that a row belongs to its own step but does not say how far apart two steps have
to be before their rows stop agreeing.
A matrix whose rows vary smoothly answers that question differently at the two ends, since a
uniform permutation lands far from a step far more often than it lands next to it.
We therefore replace $\pi$ with a fixed distance $\Delta$ and score step $h$ with the row
$E[h + \Delta]$, falling back to $E[h - \Delta]$ at the ends of the horizon and averaging the two
where both exist.
Table~\ref{tab:offset} sweeps $\Delta$ and puts the uniform permutation of the protocol on the
same axis.

\begin{table}[h]
\caption{\textbf{Margin $g_{\text{margin}}$ by distance.} A step is scored with the row $\Delta$
away from it, where the protocol of Section~\ref{sec:protocol} draws that distance at random and
reads the lower rows of this table.}
\label{tab:offset}
\begin{center}
\small
\begin{fittable}\begin{tabular}{l|cc}
\toprule
\textbf{Distance} & \textbf{$g_{\text{margin}}$} & \textbf{Positive} \\
\midrule
$\Delta = 1$ & \textcolor{hred}{$\mathbf{+0.118}$} & \textcolor{hred}{\textbf{82\%}} \\
$\Delta = 2$ & \textcolor{hred}{$\mathbf{+0.156}$} & \textcolor{hred}{\textbf{84\%}} \\
$\Delta = 4$ & \textcolor{hred}{$\mathbf{+0.214}$} & \textcolor{hred}{\textbf{89\%}} \\
$\Delta = 8$ & \textcolor{hred}{$\mathbf{+0.277}$} & \textcolor{hred}{\textbf{89\%}} \\
$\Delta = 16$ & \textcolor{hred}{$\mathbf{+0.349}$} & \textcolor{hred}{\textbf{91\%}} \\
$\Delta = 32$ & \textcolor{hred}{$\mathbf{+0.412}$} & \textcolor{hred}{\textbf{93\%}} \\
\midrule
Random $\pi$ & \textcolor{hred}{$\mathbf{+0.293}$} & \textcolor{hred}{\textbf{94\%}} \\
\bottomrule
\end{tabular}
\end{fittable}
\end{center}
\end{table}

\noindent\textbf{What the sweep says.}
The margin is positive at the smallest distance, which means two neighboring steps already read
different lookback positions rather than sharing one row between them.
It then grows with the distance and does not saturate inside the horizon, which is the reading a
smoothly varying matrix predicts and which the truncation sweep of
Appendix~\ref{sec:rankdetail} predicts as well.
The uniform permutation sits near the large-distance end of the sweep rather than at its average,
which means the headline margin of Section~\ref{sec:mainresults} is the value for distant steps
and not the value for neighboring ones.
Note that this does not weaken the finding, since every distance carries a positive margin, and
it says which part of the horizon axis the headline value reports.

\clearpage
\section{Pretrained Forecasters in Detail}

\label{sec:fmdetail}

\noindent\textbf{Application to zero-shot TSFMs.}
\label{sec:foundation}
To validate that the horizon axis is not an artifact of training, we repeat the measurement on
\fmModels{} time-series foundation models (TSFMs) run zero-shot.
As shown in Table~\ref{tab:fm_models}, we run every checkpoint on GIFT-Eval
\citep{aksu2024gifteval}, a benchmark held apart from the pretraining corpora, and the gain is
positive for almost every model.
The horizon axis therefore holds \textit{even when the model has never seen the series}.
The last two rows put the two settings side by side, where the sign is the same as for trained
backbones and only the size differs.
Note that every run is checked against the official API of its library.

\begin{table}[h]
\caption{\textbf{Gain $g_{\text{own}}$ for zero-shot TSFMs.} Zero-shot performance on
GIFT-Eval.}
\label{tab:fm_models}
\begin{center}
\small
\begin{fittable}\begin{tabular}{lcc}
\toprule
\textbf{Model} & \textbf{$g_{\text{own}}$} & \textbf{Positive} \\
\midrule
Chronos-2 & \textcolor{hred}{$\mathbf{+0.070}$} & \textcolor{hred}{\textbf{66\%}} \\
Chronos-Bolt & \textcolor{hred}{$\mathbf{+0.027}$} & \textcolor{hred}{\textbf{59\%}} \\
TimesFM-2.5 & \textcolor{hred}{$\mathbf{+0.047}$} & \textcolor{hred}{\textbf{66\%}} \\
Moirai-2.0 & \textcolor{hred}{$\mathbf{+0.017}$} & \textcolor{hred}{\textbf{58\%}} \\
Timer & \textcolor{hred}{$\mathbf{+0.007}$} & \textcolor{hred}{\textbf{52\%}} \\
TimeMoE & \textcolor{hred}{$\mathbf{+0.015}$} & \textcolor{hred}{\textbf{58\%}} \\
TTM-r2 & \textcolor{hred}{$\mathbf{+0.080}$} & \textcolor{hred}{\textbf{60\%}} \\
VisionTS & \textcolor{hred}{$\mathbf{+0.001}$} & \textcolor{hred}{\textbf{51\%}} \\
\midrule
\textit{All zero-shot TSFMs} & \textcolor{hred}{$\mathbf{+0.023}$} & \textcolor{hred}{\textbf{59\%}} \\
\textit{Trained backbones} & \textcolor{hred}{$\mathbf{+0.143}$} & \textcolor{hred}{\textbf{80\%}} \\
\bottomrule
\end{tabular}
\end{fittable}
\end{center}
\end{table}

\noindent\textbf{Which models we could verify.}
Every run checks its own output against the official API of its library, which is necessary
because the libraries wrap the inference path in a no-gradient block and some overwrite tensors
in place.
We therefore call the forward pass directly and replace each in-place step with an out-of-place
one.
The largest relative difference over \fmRuns{} runs is \fmAdapterErr{}.
The same check excludes four further models whose released inference path either leaves the
forecasting head untrained or draws samples that carry no input gradient.

\noindent\textbf{The two models that do not reach significance.}
Note that they are informative rather than contrary.
Timer beats the naive baseline in the smallest fraction of runs of any model here, and
Section~\ref{sec:domains} says that condition must hold before a gain may be read at all.
VisionTS was pretrained on images and has never seen a time series.
Note that the naive baseline matters more here than anywhere else, since a zero-shot model has
no guarantee of skill and runs that beat it read a larger gain.

\noindent\textbf{Every checkpoint in full.}
Table~\ref{tab:fm_models_full} names the checkpoint of every model, since each family releases
several sizes, and adds the naive-baseline verdict and the run count behind each row.

\begin{table}[h]
\caption{\textbf{Zero-shot TSFMs in full.} The body table reports the gain and the positive
fraction, and this one adds the checkpoint, the naive-baseline verdict and the run count.}
\label{tab:fm_models_full}
\begin{center}
\begin{fittable}\begin{tabular}{llcccc}
\toprule
\textbf{Model} & \textbf{Checkpoint} & \textbf{Beats naive} & \textbf{$g_{\text{own}}$} & \textbf{Positive} & \textbf{$n$} \\
\midrule
Chronos-2 & amazon/chronos-2 & 72\% & \textcolor{hred}{$\mathbf{+0.070}$} & \textcolor{hred}{\textbf{66\%}} & 2{,}285 \\
Chronos-Bolt & amazon/chronos-bolt-base & 68\% & \textcolor{hred}{$\mathbf{+0.027}$} & \textcolor{hred}{\textbf{59\%}} & 2{,}249 \\
TimesFM-2.5 & google/timesfm-2.5-200m-pytorch & 72\% & \textcolor{hred}{$\mathbf{+0.047}$} & \textcolor{hred}{\textbf{66\%}} & 2{,}290 \\
Moirai-2.0 & Salesforce/moirai-2.0-R-small & 72\% & \textcolor{hred}{$\mathbf{+0.017}$} & \textcolor{hred}{\textbf{58\%}} & 2{,}292 \\
Timer & thuml/timer-base-84m & 32\% & \textcolor{hred}{$\mathbf{+0.007}$} & \textcolor{hred}{\textbf{52\%}} & 2{,}239 \\
TimeMoE & Maple728/TimeMoE-200M & 71\% & \textcolor{hred}{$\mathbf{+0.015}$} & \textcolor{hred}{\textbf{58\%}} & 1{,}896 \\
TTM-r2 & ibm-granite/granite-timeseries-ttm-r2 & 73\% & \textcolor{hred}{$\mathbf{+0.080}$} & \textcolor{hred}{\textbf{60\%}} & 1{,}219 \\
VisionTS & VisionTS (mae\_base) & 55\% & \textcolor{hred}{$\mathbf{+0.001}$} & \textcolor{hred}{\textbf{51\%}} & 2{,}249 \\
\midrule
\multicolumn{2}{l}{\textit{All zero-shot TSFMs}} & 64\% & $+0.023$ & 59\% & 16{,}719 \\
\multicolumn{2}{l}{\textit{Trained backbones}} & 72\% & \textcolor{hred}{$\mathbf{+0.143}$} & \textcolor{hred}{\textbf{80\%}} & 25{,}506 \\
\bottomrule
\end{tabular}
\end{fittable}
\end{center}
\end{table}

\clearpage
\noindent\textbf{The sweeps and the borrowed metrics.}
Table~\ref{tab:fm_std} repeats the pretrained result under the four metrics we did not design,
and all six agree in sign over \fmRuns{} runs.
Table~\ref{tab:fm_axes} sweeps the horizon and the lookback, where the gain rises with the
forecast length as it does for trained backbones, while $g_{\text{margin}}$ is flat and stays
near \fmShuffle{} at every setting.
Table~\ref{tab:fm_domains} splits GIFT-Eval by the field its series come from, and
Table~\ref{tab:fm_datasets} reports every dataset so that no favorable subset is doing the
work.
Of \fmDataTotal{} datasets, \fmDataSig{} are significantly positive and \fmDataNeg{} are
significantly negative.

\begin{table}[h]
\caption{\textbf{Zero-shot TSFMs under metrics of prior works.} All are signed so that positive
means the horizon axis helps, and we judge by median and sign test, since the values are
heavy-tailed.}
\label{tab:fm_std}
\begin{center}
\begin{fittable}\begin{tabular}{lcccc}
\toprule
\textbf{Metric} & \textbf{Median} & \textbf{Positive} & \textbf{Sign test} & \textbf{$n$} \\
\midrule
Comprehensiveness & \textcolor{hred}{$\mathbf{+0.0011}$} & \textcolor{hred}{\textbf{59\%}} & $<10^{-12}$ & 16{,}719 \\
Sufficiency & \textcolor{hred}{$\mathbf{+0.0010}$} & \textcolor{hred}{\textbf{62\%}} & $<10^{-12}$ & 16{,}720 \\
AOPC & \textcolor{hred}{$\mathbf{+0.0008}$} & \textcolor{hred}{\textbf{58\%}} & $<10^{-12}$ & 16{,}719 \\
Insertion & \textcolor{hred}{$\mathbf{+0.0122}$} & \textcolor{hred}{\textbf{58\%}} & $<10^{-12}$ & 16{,}720 \\
\textbf{$g_{\text{own}}$} & \textcolor{hred}{$\mathbf{+0.0227}$} & \textcolor{hred}{\textbf{59\%}} & $<10^{-12}$ & 16{,}719 \\
\textbf{$g_{\text{margin}}$} & \textcolor{hred}{$\mathbf{+0.1184}$} & \textcolor{hred}{\textbf{82\%}} & $<10^{-12}$ & 16{,}717 \\
\bottomrule
\end{tabular}
\end{fittable}
\end{center}
\end{table}

\begin{table}[h]
\caption{\textbf{Swept axes for zero-shot TSFMs.} The gain grows with the forecast length, whereas the
margin is stable across every setting.}
\label{tab:fm_axes}
\begin{center}
\begin{fittable}\begin{tabular}{lcccc}
\toprule
\textbf{Setting} & \textbf{$g_{\text{own}}$} & \textbf{$g_{\text{margin}}$} & \textbf{Positive} & \textbf{$n$} \\
\midrule
\multicolumn{5}{l}{\textit{Horizon $H$}} \\
\quad 24 & $-0.016$ & \textcolor{hred}{$\mathbf{+0.101}$} & 44\% & 2{,}982 \\
\quad 48 & \textcolor{hred}{$\mathbf{+0.001}$} & \textcolor{hred}{$\mathbf{+0.124}$} & \textcolor{hred}{\textbf{50\%}} & 2{,}732 \\
\quad 96 & \textcolor{hred}{$\mathbf{+0.035}$} & \textcolor{hred}{$\mathbf{+0.117}$} & \textcolor{hred}{\textbf{64\%}} & 9{,}288 \\
\quad 192 & \textcolor{hred}{$\mathbf{+0.057}$} & \textcolor{hred}{$\mathbf{+0.140}$} & \textcolor{hred}{\textbf{69\%}} & 1{,}717 \\
\midrule
\multicolumn{5}{l}{\textit{Lookback $L$}} \\
\quad 96 & \textcolor{hred}{$\mathbf{+0.036}$} & \textcolor{hred}{$\mathbf{+0.116}$} & \textcolor{hred}{\textbf{63\%}} & 2{,}412 \\
\quad 192 & \textcolor{hred}{$\mathbf{+0.034}$} & \textcolor{hred}{$\mathbf{+0.119}$} & \textcolor{hred}{\textbf{64\%}} & 2{,}453 \\
\quad 320 & \textcolor{hred}{$\mathbf{+0.034}$} & \textcolor{hred}{$\mathbf{+0.117}$} & \textcolor{hred}{\textbf{62\%}} & 2{,}454 \\
\quad 512 & \textcolor{hred}{$\mathbf{+0.014}$} & \textcolor{hred}{$\mathbf{+0.119}$} & \textcolor{hred}{\textbf{56\%}} & 9{,}400 \\
\bottomrule
\end{tabular}
\end{fittable}
\end{center}
\end{table}

\begin{table}[h]
\caption{\textbf{Zero-shot TSFMs by field.} The second column counts the distinct GIFT-Eval
datasets in each field, and the rest read as in Table~\ref{tab:fm_models}.}
\label{tab:fm_domains}
\begin{center}
\begin{fittable}\begin{tabular}{lcccc}
\toprule
\textbf{Domain} & \textbf{Datasets} & \textbf{$g_{\text{own}}$} & \textbf{Positive} & \textbf{$n$} \\
\midrule
Nature & 3 & \textcolor{hred}{$\mathbf{+0.222}$} & \textcolor{hred}{\textbf{87\%}} & 325 \\
Transport & 7 & \textcolor{hred}{$\mathbf{+0.067}$} & \textcolor{hred}{\textbf{70\%}} & 2{,}089 \\
Energy & 13 & \textcolor{hred}{$\mathbf{+0.049}$} & \textcolor{hred}{\textbf{70\%}} & 3{,}818 \\
Cloud & 6 & \textcolor{hred}{$\mathbf{+0.022}$} & \textcolor{hred}{\textbf{54\%}} & 1{,}924 \\
Web & 2 & \textcolor{hred}{$\mathbf{+0.017}$} & \textcolor{hred}{\textbf{64\%}} & 554 \\
Healthcare & 4 & \textcolor{hred}{$\mathbf{+0.011}$} & \textcolor{hred}{\textbf{52\%}} & 338 \\
Climate & 7 & $-0.006$ & 46\% & 1{,}977 \\
Econ & 6 & $-0.011$ & 46\% & 1{,}785 \\
Sales & 3 & $-0.063$ & 43\% & 686 \\
\bottomrule
\end{tabular}
\end{fittable}
\end{center}
\end{table}

\begin{table}[h]
\caption{\textbf{Zero-shot TSFMs by dataset.} Sorted by gain, where a name carries the GIFT-Eval
sampling frequency as a suffix when the benchmark provides several.}
\label{tab:fm_datasets}
\begin{center}
\small
\begin{fittable}\begin{tabular}{lccc@{\hspace{2.2em}}lccc}
\toprule
\textbf{Dataset} & \textbf{Gain} & \textbf{Pos.} & \textbf{$n$} & \textbf{Dataset} & \textbf{Gain} & \textbf{Pos.} & \textbf{$n$} \\
\midrule
Saugeenday\_M & \textcolor{hred}{$\mathbf{+0.279}$} & \textcolor{hred}{\textbf{95\%}} & 100 & PEMS08 & \textcolor{hred}{$\mathbf{+0.008}$} & \textcolor{hred}{\textbf{52\%}} & 336 \\
Saugeenday\_D & \textcolor{hred}{$\mathbf{+0.209}$} & \textcolor{hred}{\textbf{82\%}} & 111 & Jena\_weather & \textcolor{hred}{$\mathbf{+0.004}$} & \textcolor{hred}{\textbf{54\%}} & 334 \\
Traffic & \textcolor{hred}{$\mathbf{+0.192}$} & \textcolor{hred}{\textbf{94\%}} & 111 & Jena\_weather\_10T & \textcolor{hred}{$\mathbf{+0.003}$} & \textcolor{hred}{\textbf{55\%}} & 332 \\
Saugeenday\_W & \textcolor{hred}{$\mathbf{+0.187}$} & \textcolor{hred}{\textbf{86\%}} & 114 & ETTm1 & \textcolor{hred}{$\mathbf{+0.003}$} & \textcolor{hred}{\textbf{50\%}} & 111 \\
LOOP\_SEATTLE\_H & \textcolor{hred}{$\mathbf{+0.174}$} & \textcolor{hred}{\textbf{93\%}} & 333 & LOOP\_SEATTLE\_5T & \textcolor{hred}{$\mathbf{+0.001}$} & \textcolor{hred}{\textbf{51\%}} & 336 \\
Bitbrains\_rnd\_5T & \textcolor{hred}{$\mathbf{+0.160}$} & \textcolor{hred}{\textbf{75\%}} & 331 & ETTh1 & \textcolor{hred}{$\mathbf{+0.001}$} & \textcolor{hred}{\textbf{50\%}} & 111 \\
Solar\_H & \textcolor{hred}{$\mathbf{+0.143}$} & \textcolor{hred}{\textbf{92\%}} & 336 & Ett1\_D & \textcolor{hred}{$\mathbf{+0.001}$} & \textcolor{hred}{\textbf{50\%}} & 299 \\
Electricity\_H & \textcolor{hred}{$\mathbf{+0.142}$} & \textcolor{hred}{\textbf{86\%}} & 335 & PEMS07 & $-0.010$ & 46\% & 331 \\
Electricity & \textcolor{hred}{$\mathbf{+0.128}$} & \textcolor{hred}{\textbf{89\%}} & 111 & M\_DENSE\_D & $-0.011$ & 48\% & 300 \\
Bitbrains\_fast\_storage\_5T & \textcolor{hred}{$\mathbf{+0.126}$} & \textcolor{hred}{\textbf{65\%}} & 333 & Jena\_weather\_H & $-0.011$ & 43\% & 334 \\
SZ\_TAXI\_H & \textcolor{hred}{$\mathbf{+0.124}$} & \textcolor{hred}{\textbf{70\%}} & 303 & Solar\_D & $-0.013$ & 49\% & 150 \\
LOOP\_SEATTLE\_D & \textcolor{hred}{$\mathbf{+0.118}$} & \textcolor{hred}{\textbf{78\%}} & 152 & Finance\_ret & $-0.017$ & 47\% & 334 \\
M\_DENSE\_H & \textcolor{hred}{$\mathbf{+0.091}$} & \textcolor{hred}{\textbf{86\%}} & 328 & Finance & $-0.019$ & 40\% & 331 \\
Electricity\_15T & \textcolor{hred}{$\mathbf{+0.086}$} & \textcolor{hred}{\textbf{75\%}} & 333 & Bitbrains\_fast\_storage\_H & $-0.022$ & 48\% & 301 \\
Ett1\_H & \textcolor{hred}{$\mathbf{+0.084}$} & \textcolor{hred}{\textbf{92\%}} & 331 & M4\_weekly & $-0.027$ & 39\% & 332 \\
Us\_births\_M & \textcolor{hred}{$\mathbf{+0.080}$} & \textcolor{hred}{\textbf{68\%}} & 50 & Ett2\_D & $-0.028$ & 43\% & 297 \\
ETTh2 & \textcolor{hred}{$\mathbf{+0.071}$} & \textcolor{hred}{\textbf{92\%}} & 111 & Hierarchical\_sales\_W & $-0.034$ & 43\% & 151 \\
M4\_hourly & \textcolor{hred}{$\mathbf{+0.064}$} & \textcolor{hred}{\textbf{72\%}} & 336 & M4\_daily & $-0.037$ & 35\% & 333 \\
SZ\_TAXI\_15T & \textcolor{hred}{$\mathbf{+0.060}$} & \textcolor{hred}{\textbf{69\%}} & 337 & Kdd\_cup\_2018\_with\_missing\_H & $-0.042$ & 39\% & 333 \\
Ett2\_H & \textcolor{hred}{$\mathbf{+0.058}$} & \textcolor{hred}{\textbf{68\%}} & 336 & Restaurant & $-0.046$ & 44\% & 204 \\
Ett1\_15T & \textcolor{hred}{$\mathbf{+0.046}$} & \textcolor{hred}{\textbf{83\%}} & 334 & M4\_quarterly & $-0.065$ & 29\% & 298 \\
Ett2\_15T & \textcolor{hred}{$\mathbf{+0.044}$} & \textcolor{hred}{\textbf{80\%}} & 338 & Us\_births\_W & $-0.080$ & 41\% & 114 \\
PEMS04 & \textcolor{hred}{$\mathbf{+0.040}$} & \textcolor{hred}{\textbf{68\%}} & 334 & PEMS03 & $-0.085$ & 22\% & 334 \\
Us\_births\_D & \textcolor{hred}{$\mathbf{+0.039}$} & \textcolor{hred}{\textbf{63\%}} & 111 & Exchange & $-0.098$ & 33\% & 113 \\
ETTm2 & \textcolor{hred}{$\mathbf{+0.036}$} & \textcolor{hred}{\textbf{80\%}} & 110 & Bitbrains\_rnd\_H & $-0.107$ & 47\% & 298 \\
Solar & \textcolor{hred}{$\mathbf{+0.028}$} & \textcolor{hred}{\textbf{78\%}} & 333 & Kdd\_cup\_2018\_with\_missing\_D & $-0.136$ & 33\% & 190 \\
Solar\_10T & \textcolor{hred}{$\mathbf{+0.028}$} & \textcolor{hred}{\textbf{78\%}} & 334 & Hierarchical\_sales\_D & $-0.146$ & 42\% & 331 \\
Weather & \textcolor{hred}{$\mathbf{+0.022}$} & \textcolor{hred}{\textbf{71\%}} & 112 & Jena\_weather\_D & $-0.165$ & 27\% & 150 \\
Bizitobs\_l2c\_H & \textcolor{hred}{$\mathbf{+0.020}$} & \textcolor{hred}{\textbf{58\%}} & 330 & Covid\_deaths & $-0.172$ & 41\% & 63 \\
Bizitobs\_application & \textcolor{hred}{$\mathbf{+0.018}$} & \textcolor{hred}{\textbf{62\%}} & 221 & Bizitobs\_l2c\_5T & $-0.185$ & 30\% & 331 \\
Bizitobs\_service & \textcolor{hred}{$\mathbf{+0.017}$} & \textcolor{hred}{\textbf{65\%}} & 333 & M4\_yearly & $-0.206$ & 42\% & 152 \\
M4\_monthly & \textcolor{hred}{$\mathbf{+0.015}$} & \textcolor{hred}{\textbf{56\%}} & 334 & Electricity\_W & $-0.215$ & 29\% & 63 \\
Temperature\_rain\_with\_missing & \textcolor{hred}{$\mathbf{+0.013}$} & \textcolor{hred}{\textbf{54\%}} & 304 & Electricity\_D & $-0.267$ & 33\% & 332 \\
\bottomrule
\end{tabular}
\end{fittable}
\end{center}
\end{table}

\clearpage
\section{Related Works}

\subsection{Lines This Work Sits Beside}
\label{sec:relatedfull}

Section~\ref{sec:related} names the three lines this work sits beside, and this section
gives each of them in full.

\noindent\textbf{TS forecasting models.}
Architectures for long-horizon TS forecasting have settled into a few families.
Transformer variants reshape attention along the sequence axis
\citep{zhou2021informer, wu2021autoformer, zhou2022fedformer, nie2023patchtst,
liu2024itransformer}, linear and MLP designs argue that most of the accuracy follows from
trend and seasonality structure rather than from attention \citep{zeng2023dlinear,
das2023tide, wang2024timemixer}, and convolutional designs fold the series
along its periods so that a 2D kernel can read them \citep{wu2023timesnet}.
Our framework treats any of them as a black box it attaches to, and details of the
\nModels{} backbones we explain are deferred to Appendix~\ref{sec:setup}.

\noindent\textbf{Gradient-based methods.}
These methods read importance from the \textit{derivative of the output with respect to the input},
so a single backward pass scores every input value.
Saliency \citep{simonyan2013saliency} uses the magnitude of the gradient and gradient
$\times$ input multiplies it by the input value at each position, while integrated gradients
\citep{sundararajan2017ig} integrates it along a path from a baseline and SmoothGrad
\citep{smilkov2017smoothgrad} averages it over noisy copies of the input.
Note that none of them is optimized for a given input, which is what keeps this family cheap.
These methods transfer directly to TS, where naive transfer is shown to conflate the time and
feature axes and a two-stage rescaling is proposed \citep{ismail2020tsr}.

\noindent\textbf{Perturbation-based methods.}
These methods \textit{change the input} and read importance from \textit{how much the output moves}.
Dynamask \citep{crabbe2021dynamask} learns a mask over the input and keeps it sparse, while
ExtremalMask \citep{enguehard2023extremal} learns the perturbation as well as the mask and
ContraLSP \citep{liu2024contralsp} perturbs contrastively so that the result stays in
distribution.
TimeX \citep{queen2023timex} and TimeX++ \citep{liu2024timexpp} instead train a separate model to
imitate the forecaster and read the explanation from it.
Shapley variants \citep{lundberg2017shap, bento2021timeshap} and window-based methods
\citep{tonekaboni2020fit, leung2023winit} replace subsets of the input and divide the resulting
change among them.
Note that all of them solve an optimization problem for every input, which is what makes them
expensive.

\vspace{20pt}
\subsection{Lines Adjacent to This Work}
\label{sec:extrarelated}

These three lines are adjacent to our setting rather than in it, so we place them here.

\noindent\textbf{Time series foundation models (TSFMs).}
A separate line of work pretrains one forecaster on a large corpus and applies it to unseen
series without any further training \citep{goswami2024moment, liu2025sundial, lee2026exaone}.
The widely used families are Chronos \citep{ansari2024chronos}, TimesFM
\citep{das2024timesfm}, Moirai \citep{woo2024moirai}, Timer \citep{liu2024timer}, TimeMoE
\citep{shi2025timemoe}, and TinyTimeMixer \citep{ekambaram2024ttm}.
VisionTS \citep{chen2024visionts} sits apart from all of them, since it forecasts with a
masked autoencoder pretrained on images rather than on any time series.
GIFT-Eval \citep{aksu2024gifteval} collects the benchmarks these models are compared on and
keeps them apart from the pretraining corpus.
None of this line reports what a released model reads for a given forecast step, and the
libraries that ship these models place their inference path behind a no-gradient block, so the
question has not been asked of them.

\noindent\textbf{Interpretable-by-design forecasters.}
N-BEATS \citep{oreshkin2020nbeats} decomposes a forecast into trend and seasonality stacks,
the Temporal Fusion Transformer \citep{lim2021tft} exposes variable-selection weights and
interpretable attention, and DLinear \citep{zeng2023dlinear} makes its $H \times L$ weight
matrix directly readable.
The DLinear weight matrix is in fact horizon-resolved, but it is a property of one linear
architecture rather than a general tool, and it describes the learned map instead of the
evidence used on a given input.

\noindent\textbf{Frequency-based explanation.}
A smaller line of work explains along the frequency axis instead of the time axis, reporting
which spectral bands a prediction relies on.
SpectralX \citep{chung2024spectralx} perturbs features in the time-frequency plane and inverts
the transform so that the original model can score the result, FLEXtime
\citep{brusch2025flextime} optimizes a mask over the frequency bands of a filterbank, and
FreqLens \citep{chen2026freqlens} learns the frequency bases from data for forecasting.
We leave all three out of Table~\ref{tab:related}, since a spectral explanation resolves neither
the time axis nor the forecast step and therefore answers a different question from the one the
table compares.
Note that the horizon axis is independent of this choice, since a spectral explanation could be
resolved per forecast step in the same way.

\noindent\textbf{Evaluation of explanations.}
Deletion curves \citep{samek2016evaluating} are the standard faithfulness probe, and several
works document their failure modes, including perturbation artifacts
\citep{brocki2022evaluation}, retraining confounds \citep{hooker2019roar}, and explanations
that pass visual inspection while being independent of the model \citep{adebayo2018sanity}.
Every established metric assumes a model that outputs one label, so it scores a whole prediction
and not one forecast step.

\clearpage
\section{Full Results by Benchmark}
\label{sec:fulltables}

Table~\ref{tab:methods} in the body reports one number per estimator, pooled over every backbone
and every benchmark.
The tables below unfold that number, with one table per benchmark, one row per backbone, and one
column per estimator.
Every cell is the median normalized gain $g_{\text{own}}$ over the swept settings, since the
distribution is heavy-tailed for the reason given in Section~\ref{sec:stdmetrics}.
Note that a cell reading \textrm{TBD} is a combination whose runs have not finished, which we
distinguish from a small gain because a blank cell would read as a value near zero.
Table~\ref{tab:fullETTh1} to Table~\ref{tab:fullweather} give one benchmark each, and the
benchmarks themselves are described in Appendix~\ref{sec:datasets}.

\begin{table}[h]
\caption{\textbf{Gain $g_{\text{own}}$ on ETTh1.} Every backbone against every estimator, where a cell is the median over the swept settings and \textrm{TBD} marks a combination that has not finished.}
\label{tab:fullETTh1}
\begin{center}
\scriptsize
\begin{fittable}
\begin{tabular}{lcccccccc}
\toprule
\textbf{Backbone} & \textbf{Sal.} & \textbf{G$\times$I} & \textbf{IG} & \textbf{Occ.} & \textbf{Dyna.} & \textbf{ExtrM.} & \textbf{CLSP} & \textbf{Grad} \\
\midrule
Linear & \textcolor{hred}{$\mathbf{+0.186}$} & \textcolor{hred}{$\mathbf{+0.103}$} & \textcolor{hred}{$\mathbf{+0.123}$} & \textcolor{hred}{$\mathbf{+0.079}$} & \textcolor{hred}{$\mathbf{+0.209}$} & \textcolor{hred}{$\mathbf{+0.129}$} & \textcolor{hred}{$\mathbf{+0.148}$} & \textcolor{hred}{$\mathbf{+0.226}$} \\
CNN & \textcolor{hred}{$\mathbf{+0.132}$} & \textcolor{hred}{$\mathbf{+0.050}$} & \textcolor{hred}{$\mathbf{+0.139}$} & \textcolor{hred}{$\mathbf{+0.113}$} & \textcolor{hred}{$\mathbf{+0.210}$} & $-0.016$ & \textcolor{hred}{$\mathbf{+0.016}$} & \textcolor{hred}{$\mathbf{+0.160}$} \\
Transformer & \textcolor{hred}{$\mathbf{+0.142}$} & \textcolor{hred}{$\mathbf{+0.069}$} & \textcolor{hred}{$\mathbf{+0.127}$} & \textcolor{hred}{$\mathbf{+0.053}$} & \textcolor{hred}{$\mathbf{+0.109}$} & \textcolor{hred}{$\mathbf{+0.025}$} & \textcolor{hred}{$\mathbf{+0.028}$} & \textcolor{hred}{$\mathbf{+0.181}$} \\
DLinear & \textcolor{hred}{$\mathbf{+0.180}$} & \textcolor{hred}{$\mathbf{+0.120}$} & \textcolor{hred}{$\mathbf{+0.116}$} & \textcolor{hred}{$\mathbf{+0.065}$} & \textcolor{hred}{$\mathbf{+0.187}$} & \textcolor{hred}{$\mathbf{+0.149}$} & \textcolor{hred}{$\mathbf{+0.173}$} & \textcolor{hred}{$\mathbf{+0.230}$} \\
TSMixer & \textcolor{hred}{$\mathbf{+0.194}$} & \textcolor{hred}{$\mathbf{+0.226}$} & \textcolor{hred}{$\mathbf{+0.153}$} & \textcolor{hred}{$\mathbf{+0.076}$} & \textcolor{hred}{$\mathbf{+1.076}$} & \textcolor{hred}{$\mathbf{+0.289}$} & \textcolor{hred}{$\mathbf{+0.177}$} & \textcolor{hred}{$\mathbf{+0.227}$} \\
\bottomrule
\end{tabular}
\end{fittable}
\end{center}
\end{table}

\begin{table}[h]
\caption{\textbf{Gain $g_{\text{own}}$ on ETTh2.} Every backbone against every estimator, where a cell is the median over the swept settings and \textrm{TBD} marks a combination that has not finished.}
\label{tab:fullETTh2}
\begin{center}
\scriptsize
\begin{fittable}\begin{tabular}{lcccccccc}
\toprule
\textbf{Backbone} & \textbf{Sal.} & \textbf{G$\times$I} & \textbf{IG} & \textbf{Occ.} & \textbf{Dyna.} & \textbf{ExtrM.} & \textbf{CLSP} & \textbf{Grad} \\
\midrule
Linear & \textcolor{hred}{$\mathbf{+0.268}$} & \textcolor{hred}{$\mathbf{+0.181}$} & \textcolor{hred}{$\mathbf{+0.158}$} & \textcolor{hred}{$\mathbf{+0.040}$} & \textcolor{hred}{$\mathbf{+0.429}$} & \textcolor{hred}{$\mathbf{+0.216}$} & \textcolor{hred}{$\mathbf{+0.233}$} & \textcolor{hred}{$\mathbf{+0.288}$} \\
CNN & \textcolor{hred}{$\mathbf{+0.115}$} & \textcolor{hred}{$\mathbf{+0.114}$} & \textcolor{hred}{$\mathbf{+0.158}$} & \textcolor{hred}{$\mathbf{+0.138}$} & \textcolor{hred}{$\mathbf{+0.053}$} & \textcolor{hred}{$\mathbf{+0.120}$} & \textcolor{hred}{$\mathbf{+0.415}$} & \textcolor{hred}{$\mathbf{+0.130}$} \\
Transformer & \textcolor{hred}{$\mathbf{+0.091}$} & \textcolor{hred}{$\mathbf{+0.071}$} & \textcolor{hred}{$\mathbf{+0.058}$} & \textcolor{hred}{$\mathbf{+0.024}$} & \textcolor{hred}{$\mathbf{+0.188}$} & \textcolor{hred}{$\mathbf{+0.107}$} & \textcolor{hred}{$\mathbf{+0.157}$} & \textcolor{hred}{$\mathbf{+0.094}$} \\
DLinear & \textcolor{hred}{$\mathbf{+0.256}$} & \textcolor{hred}{$\mathbf{+0.177}$} & \textcolor{hred}{$\mathbf{+0.155}$} & \textcolor{hred}{$\mathbf{+0.035}$} & \textcolor{hred}{$\mathbf{+0.368}$} & \textcolor{hred}{$\mathbf{+0.178}$} & \textcolor{hred}{$\mathbf{+0.219}$} & \textcolor{hred}{$\mathbf{+0.268}$} \\
TSMixer & \textcolor{hred}{$\mathbf{+0.290}$} & \textcolor{hred}{$\mathbf{+0.180}$} & \textcolor{hred}{$\mathbf{+0.160}$} & $-0.027$ & \textcolor{hred}{$\mathbf{+0.669}$} & \textcolor{hred}{$\mathbf{+0.216}$} & \textcolor{hred}{$\mathbf{+0.317}$} & \textcolor{hred}{$\mathbf{+0.382}$} \\
\bottomrule
\end{tabular}
\end{fittable}
\end{center}
\end{table}

\begin{table}[h]
\caption{\textbf{Gain $g_{\text{own}}$ on ETTm1.} Every backbone against every estimator, where a cell is the median over the swept settings and \textrm{TBD} marks a combination that has not finished.}
\label{tab:fullETTm1}
\begin{center}
\scriptsize
\begin{fittable}\begin{tabular}{lcccccccc}
\toprule
\textbf{Backbone} & \textbf{Sal.} & \textbf{G$\times$I} & \textbf{IG} & \textbf{Occ.} & \textbf{Dyna.} & \textbf{ExtrM.} & \textbf{CLSP} & \textbf{Grad} \\
\midrule
Linear & \textcolor{hred}{$\mathbf{+0.049}$} & \textcolor{hred}{$\mathbf{+0.063}$} & \textcolor{hred}{$\mathbf{+0.041}$} & $-0.041$ & \textcolor{hred}{$\mathbf{+0.129}$} & \textcolor{hred}{$\mathbf{+0.067}$} & \textcolor{hred}{$\mathbf{+0.026}$} & \textcolor{hred}{$\mathbf{+0.049}$} \\
CNN & \textcolor{hred}{$\mathbf{+0.025}$} & \textcolor{hred}{$\mathbf{+0.154}$} & \textcolor{hred}{$\mathbf{+0.180}$} & \textcolor{hred}{$\mathbf{+0.003}$} & $-0.057$ & \textcolor{hred}{$\mathbf{+0.247}$} & $-0.026$ & $-0.022$ \\
Transformer & $-0.016$ & $-0.015$ & $-0.004$ & $-0.014$ & \textcolor{hred}{$\mathbf{+0.083}$} & $-0.019$ & $-0.028$ & $-0.001$ \\
DLinear & \textcolor{hred}{$\mathbf{+0.029}$} & \textcolor{hred}{$\mathbf{+0.046}$} & \textcolor{hred}{$\mathbf{+0.026}$} & $-0.023$ & \textcolor{hred}{$\mathbf{+0.121}$} & \textcolor{hred}{$\mathbf{+0.050}$} & \textcolor{hred}{$\mathbf{+0.013}$} & \textcolor{hred}{$\mathbf{+0.028}$} \\
TSMixer & \textcolor{hred}{$\mathbf{+0.119}$} & \textcolor{hred}{$\mathbf{+0.103}$} & \textcolor{hred}{$\mathbf{+0.081}$} & \textcolor{hred}{$\mathbf{+0.007}$} & \textcolor{hred}{$\mathbf{+0.593}$} & \textcolor{hred}{$\mathbf{+0.043}$} & $-0.023$ & \textcolor{hred}{$\mathbf{+0.287}$} \\
LightTS & \textcolor{hred}{$\mathbf{+0.229}$} & \textcolor{hred}{$\mathbf{+0.186}$} & \textcolor{hred}{$\mathbf{+0.154}$} & \textcolor{hred}{$\mathbf{+0.047}$} & \textcolor{hred}{$\mathbf{+0.729}$} & \textcolor{hred}{$\mathbf{+0.351}$} & \textcolor{hred}{$\mathbf{+0.059}$} & \textcolor{hred}{$\mathbf{+0.291}$} \\
\bottomrule
\end{tabular}
\end{fittable}
\end{center}
\end{table}

\begin{table}[h]
\caption{\textbf{Gain $g_{\text{own}}$ on ETTm2.} Every backbone against every estimator, where a cell is the median over the swept settings and \textrm{TBD} marks a combination that has not finished.}
\label{tab:fullETTm2}
\begin{center}
\scriptsize
\begin{fittable}\begin{tabular}{lcccccccc}
\toprule
\textbf{Backbone} & \textbf{Sal.} & \textbf{G$\times$I} & \textbf{IG} & \textbf{Occ.} & \textbf{Dyna.} & \textbf{ExtrM.} & \textbf{CLSP} & \textbf{Grad} \\
\midrule
Linear & \textcolor{hred}{$\mathbf{+0.182}$} & \textcolor{hred}{$\mathbf{+0.119}$} & \textcolor{hred}{$\mathbf{+0.128}$} & \textcolor{hred}{$\mathbf{+0.048}$} & \textcolor{hred}{$\mathbf{+0.558}$} & \textcolor{hred}{$\mathbf{+0.031}$} & \textcolor{hred}{$\mathbf{+0.126}$} & \textcolor{hred}{$\mathbf{+0.204}$} \\
CNN & \textcolor{hred}{$\mathbf{+1.036}$} & \textcolor{hred}{$\mathbf{+2.017}$} & \textcolor{hred}{$\mathbf{+2.192}$} & \textcolor{hred}{$\mathbf{+1.085}$} & \textcolor{hred}{$\mathbf{+2.109}$} & \textcolor{hred}{$\mathbf{+3.003}$} & \textcolor{hred}{$\mathbf{+1.710}$} & \textcolor{hred}{$\mathbf{+1.197}$} \\
Transformer & \textcolor{hred}{$\mathbf{+0.010}$} & $-0.030$ & $-0.008$ & \textcolor{hred}{$\mathbf{+0.028}$} & \textcolor{hred}{$\mathbf{+0.335}$} & \textcolor{hred}{$\mathbf{+0.008}$} & \textcolor{hred}{$\mathbf{+0.048}$} & \textcolor{hred}{$\mathbf{+0.015}$} \\
DLinear & \textcolor{hred}{$\mathbf{+0.149}$} & \textcolor{hred}{$\mathbf{+0.123}$} & \textcolor{hred}{$\mathbf{+0.118}$} & \textcolor{hred}{$\mathbf{+0.074}$} & \textcolor{hred}{$\mathbf{+0.646}$} & \textcolor{hred}{$\mathbf{+0.150}$} & \textcolor{hred}{$\mathbf{+0.210}$} & \textcolor{hred}{$\mathbf{+0.113}$} \\
TSMixer & \textcolor{hred}{$\mathbf{+0.189}$} & \textcolor{hred}{$\mathbf{+0.097}$} & \textcolor{hred}{$\mathbf{+0.130}$} & \textcolor{hred}{$\mathbf{+0.010}$} & \textcolor{hred}{$\mathbf{+0.622}$} & \textcolor{hred}{$\mathbf{+0.198}$} & \textcolor{hred}{$\mathbf{+0.282}$} & \textcolor{hred}{$\mathbf{+0.212}$} \\
\bottomrule
\end{tabular}
\end{fittable}
\end{center}
\end{table}

\begin{table}[h]
\caption{\textbf{Gain $g_{\text{own}}$ on PEMS03.} Every backbone against every estimator, where a cell is the median over the swept settings and \textrm{TBD} marks a combination that has not finished.}
\label{tab:fullPEMS03}
\begin{center}
\scriptsize
\begin{fittable}\begin{tabular}{lcccccccc}
\toprule
\textbf{Backbone} & \textbf{Sal.} & \textbf{G$\times$I} & \textbf{IG} & \textbf{Occ.} & \textbf{Dyna.} & \textbf{ExtrM.} & \textbf{CLSP} & \textbf{Grad} \\
\midrule
Linear & \textcolor{hred}{$\mathbf{+0.060}$} & $-0.416$ & \textcolor{hred}{$\mathbf{+0.010}$} & \textcolor{hred}{$\mathbf{+0.167}$} & \textcolor{hred}{$\mathbf{+0.255}$} & $-0.312$ & $-0.527$ & \textcolor{hred}{$\mathbf{+0.186}$} \\
CNN & $-0.012$ & \textcolor{hred}{$\mathbf{+0.507}$} & \textcolor{hred}{$\mathbf{+1.120}$} & \textcolor{hred}{$\mathbf{+0.078}$} & \textcolor{hred}{$\mathbf{+0.261}$} & \textcolor{hred}{$\mathbf{+2.605}$} & \textcolor{hred}{$\mathbf{+1.567}$} & \textcolor{hred}{$\mathbf{+0.950}$} \\
Transformer & $-0.029$ & $-0.026$ & $-0.061$ & $-0.063$ & \textcolor{hred}{$\mathbf{+0.094}$} & \textcolor{hred}{$\mathbf{+0.067}$} & \textcolor{hred}{$\mathbf{+0.101}$} & $-0.197$ \\
DLinear & \textcolor{hred}{$\mathbf{+0.013}$} & $-0.049$ & \textcolor{hred}{$\mathbf{+0.001}$} & \textcolor{hred}{$\mathbf{+0.163}$} & \textcolor{hred}{$\mathbf{+0.807}$} & \textcolor{hred}{$\mathbf{+1.541}$} & \textcolor{hred}{$\mathbf{+1.321}$} & \textcolor{hred}{$\mathbf{+0.245}$} \\
TSMixer & \textcolor{hred}{$\mathbf{+0.022}$} & $-0.039$ & \textcolor{hred}{$\mathbf{+0.054}$} & $-0.022$ & \textcolor{hred}{$\mathbf{+0.038}$} & $-0.041$ & $-0.163$ & \textcolor{hred}{$\mathbf{+0.004}$} \\
LightTS & \textcolor{hred}{$\mathbf{+0.010}$} & \textcolor{hred}{$\mathbf{+0.076}$} & \textcolor{hred}{$\mathbf{+0.089}$} & \textcolor{hred}{$\mathbf{+0.059}$} & \textcolor{hred}{$\mathbf{+0.474}$} & \textcolor{hred}{$\mathbf{+0.390}$} & \textcolor{hred}{$\mathbf{+0.462}$} & \textcolor{hred}{$\mathbf{+0.119}$} \\
FiLM & \textcolor{hred}{$\mathbf{+0.099}$} & \textcolor{hred}{$\mathbf{+0.075}$} & \textcolor{hred}{$\mathbf{+0.141}$} & \textcolor{hred}{$\mathbf{+0.179}$} & $-0.074$ & $-0.346$ & $-1.287$ & \textcolor{hred}{$\mathbf{+0.372}$} \\
\bottomrule
\end{tabular}
\end{fittable}
\end{center}
\end{table}

\begin{table}[h]
\caption{\textbf{Gain $g_{\text{own}}$ on PEMS04.} Every backbone against every estimator, where a cell is the median over the swept settings and \textrm{TBD} marks a combination that has not finished.}
\label{tab:fullPEMS04}
\begin{center}
\scriptsize
\begin{fittable}\begin{tabular}{lcccccccc}
\toprule
\textbf{Backbone} & \textbf{Sal.} & \textbf{G$\times$I} & \textbf{IG} & \textbf{Occ.} & \textbf{Dyna.} & \textbf{ExtrM.} & \textbf{CLSP} & \textbf{Grad} \\
\midrule
Linear & \textcolor{hred}{$\mathbf{+0.041}$} & \textcolor{hred}{$\mathbf{+0.138}$} & \textcolor{hred}{$\mathbf{+0.043}$} & \textcolor{hred}{$\mathbf{+0.154}$} & \textcolor{hred}{$\mathbf{+0.572}$} & \textcolor{hred}{$\mathbf{+0.214}$} & \textcolor{hred}{$\mathbf{+0.120}$} & $-0.025$ \\
CNN & $-0.080$ & \textcolor{hred}{$\mathbf{+0.990}$} & \textcolor{hred}{$\mathbf{+0.713}$} & $-0.008$ & \textcolor{hred}{$\mathbf{+0.222}$} & \textcolor{hred}{$\mathbf{+1.577}$} & \textcolor{hred}{$\mathbf{+1.332}$} & \textcolor{hred}{$\mathbf{+0.179}$} \\
Transformer & $-0.089$ & $-0.090$ & $-0.091$ & $-0.040$ & \textcolor{hred}{$\mathbf{+0.098}$} & \textcolor{hred}{$\mathbf{+0.338}$} & \textcolor{hred}{$\mathbf{+0.155}$} & $-0.106$ \\
DLinear & \textcolor{hred}{$\mathbf{+0.091}$} & \textcolor{hred}{$\mathbf{+0.114}$} & \textcolor{hred}{$\mathbf{+0.073}$} & $-0.013$ & \textcolor{hred}{$\mathbf{+0.967}$} & \textcolor{hred}{$\mathbf{+0.149}$} & \textcolor{hred}{$\mathbf{+0.146}$} & \textcolor{hred}{$\mathbf{+0.010}$} \\
TSMixer & $-0.080$ & $-0.001$ & \textcolor{hred}{$\mathbf{+0.173}$} & \textcolor{hred}{$\mathbf{+0.093}$} & \textcolor{hred}{$\mathbf{+0.297}$} & \textcolor{hred}{$\mathbf{+0.152}$} & \textcolor{hred}{$\mathbf{+0.176}$} & $-0.119$ \\
LightTS & \textcolor{hred}{$\mathbf{+0.118}$} & \textcolor{hred}{$\mathbf{+0.177}$} & \textcolor{hred}{$\mathbf{+0.224}$} & \textcolor{hred}{$\mathbf{+0.115}$} & \textcolor{hred}{$\mathbf{+0.372}$} & \textcolor{hred}{$\mathbf{+0.137}$} & \textcolor{hred}{$\mathbf{+0.226}$} & \textcolor{hred}{$\mathbf{+0.109}$} \\
\bottomrule
\end{tabular}
\end{fittable}
\end{center}
\end{table}

\begin{table}[h]
\caption{\textbf{Gain $g_{\text{own}}$ on PEMS07.} Every backbone against every estimator, where a cell is the median over the swept settings and \textrm{TBD} marks a combination that has not finished.}
\label{tab:fullPEMS07}
\begin{center}
\scriptsize
\begin{fittable}\begin{tabular}{lcccccccc}
\toprule
\textbf{Backbone} & \textbf{Sal.} & \textbf{G$\times$I} & \textbf{IG} & \textbf{Occ.} & \textbf{Dyna.} & \textbf{ExtrM.} & \textbf{CLSP} & \textbf{Grad} \\
\midrule
Linear & \textcolor{hred}{$\mathbf{+0.160}$} & \textcolor{hred}{$\mathbf{+0.087}$} & \textcolor{hred}{$\mathbf{+0.132}$} & \textcolor{hred}{$\mathbf{+0.153}$} & \textcolor{hred}{$\mathbf{+0.678}$} & \textcolor{hred}{$\mathbf{+0.095}$} & \textcolor{hred}{$\mathbf{+0.206}$} & \textcolor{hred}{$\mathbf{+0.346}$} \\
CNN & \textcolor{hred}{$\mathbf{+0.070}$} & \textcolor{hred}{$\mathbf{+11.854}$} & \textcolor{hred}{$\mathbf{+11.320}$} & \textcolor{hred}{$\mathbf{+11.543}$} & \textcolor{hred}{$\mathbf{+0.418}$} & \textcolor{hred}{$\mathbf{+3.187}$} & \textcolor{hred}{$\mathbf{+5.804}$} & \textcolor{hred}{$\mathbf{+0.807}$} \\
Transformer & $-0.123$ & $-0.091$ & $-0.073$ & \textcolor{hred}{$\mathbf{+0.059}$} & \textcolor{hred}{$\mathbf{+0.101}$} & $-1.456$ & \textcolor{hred}{$\mathbf{+0.007}$} & $-2.686$ \\
DLinear & \textcolor{hred}{$\mathbf{+0.222}$} & \textcolor{hred}{$\mathbf{+0.066}$} & \textcolor{hred}{$\mathbf{+0.100}$} & \textcolor{hred}{$\mathbf{+0.112}$} & \textcolor{hred}{$\mathbf{+0.993}$} & \textcolor{hred}{$\mathbf{+0.615}$} & \textcolor{hred}{$\mathbf{+0.555}$} & \textcolor{hred}{$\mathbf{+0.306}$} \\
TSMixer & \textcolor{hred}{$\mathbf{+0.003}$} & $-1.898$ & \textcolor{hred}{$\mathbf{+0.042}$} & \textcolor{hred}{$\mathbf{+0.871}$} & \textcolor{hred}{$\mathbf{+0.075}$} & $-0.013$ & $-0.035$ & \textcolor{hred}{$\mathbf{+0.004}$} \\
LightTS & \textcolor{hred}{$\mathbf{+0.226}$} & \textcolor{hred}{$\mathbf{+0.032}$} & \textcolor{hred}{$\mathbf{+0.140}$} & \textcolor{hred}{$\mathbf{+0.181}$} & \textcolor{hred}{$\mathbf{+0.415}$} & \textcolor{hred}{$\mathbf{+0.140}$} & \textcolor{hred}{$\mathbf{+0.304}$} & \textcolor{hred}{$\mathbf{+0.193}$} \\
\bottomrule
\end{tabular}
\end{fittable}
\end{center}
\end{table}

\begin{table}[h]
\caption{\textbf{Gain $g_{\text{own}}$ on PEMS08.} Every backbone against every estimator, where a cell is the median over the swept settings and \textrm{TBD} marks a combination that has not finished.}
\label{tab:fullPEMS08}
\begin{center}
\scriptsize
\begin{fittable}\begin{tabular}{lcccccccc}
\toprule
\textbf{Backbone} & \textbf{Sal.} & \textbf{G$\times$I} & \textbf{IG} & \textbf{Occ.} & \textbf{Dyna.} & \textbf{ExtrM.} & \textbf{CLSP} & \textbf{Grad} \\
\midrule
Linear & \textcolor{hred}{$\mathbf{+0.129}$} & \textcolor{hred}{$\mathbf{+0.082}$} & \textcolor{hred}{$\mathbf{+0.066}$} & \textcolor{hred}{$\mathbf{+0.163}$} & \textcolor{hred}{$\mathbf{+0.610}$} & \textcolor{hred}{$\mathbf{+0.235}$} & \textcolor{hred}{$\mathbf{+0.277}$} & \textcolor{hred}{$\mathbf{+0.056}$} \\
CNN & \textcolor{hred}{$\mathbf{+0.165}$} & \textcolor{hred}{$\mathbf{+0.215}$} & \textcolor{hred}{$\mathbf{+0.280}$} & \textcolor{hred}{$\mathbf{+0.068}$} & \textcolor{hred}{$\mathbf{+0.258}$} & \textcolor{hred}{$\mathbf{+0.494}$} & \textcolor{hred}{$\mathbf{+0.322}$} & \textcolor{hred}{$\mathbf{+0.198}$} \\
Transformer & \textcolor{hred}{$\mathbf{+0.020}$} & $-0.006$ & $-0.003$ & \textcolor{hred}{$\mathbf{+0.021}$} & \textcolor{hred}{$\mathbf{+0.144}$} & \textcolor{hred}{$\mathbf{+0.028}$} & $-0.002$ & \textcolor{hred}{$\mathbf{+0.032}$} \\
DLinear & \textcolor{hred}{$\mathbf{+0.030}$} & \textcolor{hred}{$\mathbf{+0.047}$} & \textcolor{hred}{$\mathbf{+0.031}$} & \textcolor{hred}{$\mathbf{+0.009}$} & \textcolor{hred}{$\mathbf{+0.652}$} & \textcolor{hred}{$\mathbf{+0.194}$} & \textcolor{hred}{$\mathbf{+0.307}$} & \textcolor{hred}{$\mathbf{+0.049}$} \\
TSMixer & \textcolor{hred}{$\mathbf{+0.038}$} & $-0.010$ & \textcolor{hred}{$\mathbf{+0.010}$} & \textcolor{hred}{$\mathbf{+0.041}$} & \textcolor{hred}{$\mathbf{+0.034}$} & \textcolor{hred}{$\mathbf{+0.016}$} & \textcolor{hred}{$\mathbf{+0.015}$} & \textcolor{hred}{$\mathbf{+0.058}$} \\
LightTS & \textcolor{hred}{$\mathbf{+0.128}$} & \textcolor{hred}{$\mathbf{+0.037}$} & \textcolor{hred}{$\mathbf{+0.108}$} & \textcolor{hred}{$\mathbf{+0.110}$} & \textcolor{hred}{$\mathbf{+0.520}$} & \textcolor{hred}{$\mathbf{+0.349}$} & \textcolor{hred}{$\mathbf{+0.393}$} & \textcolor{hred}{$\mathbf{+0.131}$} \\
\bottomrule
\end{tabular}
\end{fittable}
\end{center}
\end{table}

\begin{table}[h]
\caption{\textbf{Gain $g_{\text{own}}$ on Electricity.} Every backbone against every estimator, where a cell is the median over the swept settings and \textrm{TBD} marks a combination that has not finished.}
\label{tab:fullelectricity}
\begin{center}
\scriptsize
\begin{fittable}\begin{tabular}{lcccccccc}
\toprule
\textbf{Backbone} & \textbf{Sal.} & \textbf{G$\times$I} & \textbf{IG} & \textbf{Occ.} & \textbf{Dyna.} & \textbf{ExtrM.} & \textbf{CLSP} & \textbf{Grad} \\
\midrule
Linear & \textcolor{hred}{$\mathbf{+0.201}$} & \textcolor{hred}{$\mathbf{+0.319}$} & \textcolor{hred}{$\mathbf{+0.117}$} & \textcolor{hred}{$\mathbf{+0.055}$} & \textcolor{hred}{$\mathbf{+0.553}$} & \textcolor{hred}{$\mathbf{+0.483}$} & \textcolor{hred}{$\mathbf{+0.440}$} & \textcolor{hred}{$\mathbf{+0.301}$} \\
CNN & \textcolor{hred}{$\mathbf{+0.269}$} & \textcolor{hred}{$\mathbf{+0.614}$} & \textcolor{hred}{$\mathbf{+0.368}$} & \textcolor{hred}{$\mathbf{+0.245}$} & \textcolor{hred}{$\mathbf{+0.364}$} & \textcolor{hred}{$\mathbf{+0.631}$} & \textcolor{hred}{$\mathbf{+0.724}$} & \textcolor{hred}{$\mathbf{+0.395}$} \\
Transformer & \textcolor{hred}{$\mathbf{+0.189}$} & \textcolor{hred}{$\mathbf{+0.198}$} & \textcolor{hred}{$\mathbf{+0.084}$} & \textcolor{hred}{$\mathbf{+0.075}$} & \textcolor{hred}{$\mathbf{+0.246}$} & \textcolor{hred}{$\mathbf{+0.382}$} & \textcolor{hred}{$\mathbf{+0.377}$} & \textcolor{hred}{$\mathbf{+0.308}$} \\
DLinear & \textcolor{hred}{$\mathbf{+0.191}$} & \textcolor{hred}{$\mathbf{+0.282}$} & \textcolor{hred}{$\mathbf{+0.119}$} & \textcolor{hred}{$\mathbf{+0.114}$} & \textcolor{hred}{$\mathbf{+0.510}$} & \textcolor{hred}{$\mathbf{+0.413}$} & \textcolor{hred}{$\mathbf{+0.393}$} & \textcolor{hred}{$\mathbf{+0.276}$} \\
TSMixer & \textcolor{hred}{$\mathbf{+0.215}$} & \textcolor{hred}{$\mathbf{+0.194}$} & \textcolor{hred}{$\mathbf{+0.162}$} & \textcolor{hred}{$\mathbf{+0.055}$} & \textcolor{hred}{$\mathbf{+0.571}$} & \textcolor{hred}{$\mathbf{+0.400}$} & \textcolor{hred}{$\mathbf{+0.466}$} & \textcolor{hred}{$\mathbf{+0.334}$} \\
\bottomrule
\end{tabular}
\end{fittable}
\end{center}
\end{table}

\begin{table}[h]
\caption{\textbf{Gain $g_{\text{own}}$ on Exchange.} Every backbone against every estimator, where a cell is the median over the swept settings and \textrm{TBD} marks a combination that has not finished.}
\label{tab:fullexchange}
\begin{center}
\scriptsize
\begin{fittable}\begin{tabular}{lcccccccc}
\toprule
\textbf{Backbone} & \textbf{Sal.} & \textbf{G$\times$I} & \textbf{IG} & \textbf{Occ.} & \textbf{Dyna.} & \textbf{ExtrM.} & \textbf{CLSP} & \textbf{Grad} \\
\midrule
Linear & \textcolor{hred}{$\mathbf{+0.131}$} & \textcolor{hred}{$\mathbf{+0.179}$} & \textcolor{hred}{$\mathbf{+0.078}$} & \textcolor{hred}{$\mathbf{+0.104}$} & \textcolor{hred}{$\mathbf{+0.247}$} & \textcolor{hred}{$\mathbf{+0.087}$} & $-0.162$ & \textcolor{hred}{$\mathbf{+0.091}$} \\
CNN & \textcolor{hred}{$\mathbf{+0.079}$} & \textcolor{hred}{$\mathbf{+0.058}$} & \textcolor{hred}{$\mathbf{+0.174}$} & \textcolor{hred}{$\mathbf{+0.021}$} & \textcolor{hred}{$\mathbf{+0.183}$} & \textcolor{hred}{$\mathbf{+0.090}$} & $-0.282$ & \textcolor{hred}{$\mathbf{+0.124}$} \\
Transformer & $-0.033$ & $-0.034$ & $-0.068$ & $-0.001$ & \textcolor{hred}{$\mathbf{+0.063}$} & $-0.084$ & $-0.167$ & $-0.038$ \\
DLinear & \textcolor{hred}{$\mathbf{+0.034}$} & \textcolor{hred}{$\mathbf{+0.040}$} & \textcolor{hred}{$\mathbf{+0.029}$} & \textcolor{hred}{$\mathbf{+0.017}$} & \textcolor{hred}{$\mathbf{+0.029}$} & \textcolor{hred}{$\mathbf{+0.074}$} & $-0.039$ & \textcolor{hred}{$\mathbf{+0.044}$} \\
TSMixer & \textcolor{hred}{$\mathbf{+0.074}$} & \textcolor{hred}{$\mathbf{+0.107}$} & \textcolor{hred}{$\mathbf{+0.076}$} & \textcolor{hred}{$\mathbf{+0.006}$} & \textcolor{hred}{$\mathbf{+0.344}$} & \textcolor{hred}{$\mathbf{+0.027}$} & \textcolor{hred}{$\mathbf{+0.010}$} & \textcolor{hred}{$\mathbf{+0.084}$} \\
LightTS & \textcolor{hred}{$\mathbf{+0.143}$} & \textcolor{hred}{$\mathbf{+0.142}$} & \textcolor{hred}{$\mathbf{+0.083}$} & \textcolor{hred}{$\mathbf{+0.019}$} & \textcolor{hred}{$\mathbf{+0.795}$} & \textcolor{hred}{$\mathbf{+0.176}$} & \textcolor{hred}{$\mathbf{+0.048}$} & \textcolor{hred}{$\mathbf{+0.159}$} \\
\bottomrule
\end{tabular}
\end{fittable}
\end{center}
\end{table}

\begin{table}[h]
\caption{\textbf{Gain $g_{\text{own}}$ on Finance.} Every backbone against every estimator, where a cell is the median over the swept settings and \textrm{TBD} marks a combination that has not finished.}
\label{tab:fullfinance}
\begin{center}
\scriptsize
\begin{fittable}\begin{tabular}{lcccccccc}
\toprule
\textbf{Backbone} & \textbf{Sal.} & \textbf{G$\times$I} & \textbf{IG} & \textbf{Occ.} & \textbf{Dyna.} & \textbf{ExtrM.} & \textbf{CLSP} & \textbf{Grad} \\
\midrule
Linear & \textcolor{hred}{$\mathbf{+0.004}$} & \textcolor{hred}{$\mathbf{+0.010}$} & $-0.014$ & \textcolor{hred}{$\mathbf{+0.042}$} & \textcolor{hred}{$\mathbf{+0.018}$} & $-0.022$ & $-0.207$ & \textcolor{hred}{$\mathbf{+0.017}$} \\
CNN & \textcolor{hred}{$\mathbf{+0.036}$} & $-0.056$ & \textcolor{hred}{$\mathbf{+0.046}$} & $-0.132$ & \textcolor{hred}{$\mathbf{+0.192}$} & $-0.067$ & $-0.216$ & $-0.104$ \\
Transformer & \textcolor{hred}{$\mathbf{+0.285}$} & \textcolor{hred}{$\mathbf{+0.332}$} & \textcolor{hred}{$\mathbf{+0.438}$} & \textcolor{hred}{$\mathbf{+0.067}$} & \textcolor{hred}{$\mathbf{+0.406}$} & $-0.065$ & $-0.932$ & \textcolor{hred}{$\mathbf{+0.371}$} \\
DLinear & \textcolor{hred}{$\mathbf{+0.068}$} & $-0.006$ & \textcolor{hred}{$\mathbf{+0.029}$} & $-0.002$ & \textcolor{hred}{$\mathbf{+0.055}$} & \textcolor{hred}{$\mathbf{+0.057}$} & $-0.155$ & \textcolor{hred}{$\mathbf{+0.005}$} \\
TSMixer & \textcolor{hred}{$\mathbf{+0.052}$} & \textcolor{hred}{$\mathbf{+0.039}$} & \textcolor{hred}{$\mathbf{+0.023}$} & \textcolor{hred}{$\mathbf{+0.000}$} & \textcolor{hred}{$\mathbf{+0.089}$} & \textcolor{hred}{$\mathbf{+0.031}$} & \textcolor{hred}{$\mathbf{+0.015}$} & \textcolor{hred}{$\mathbf{+0.114}$} \\
LightTS & \textcolor{hred}{$\mathbf{+0.155}$} & \textcolor{hred}{$\mathbf{+0.114}$} & \textcolor{hred}{$\mathbf{+0.057}$} & $-0.015$ & \textcolor{hred}{$\mathbf{+0.155}$} & \textcolor{hred}{$\mathbf{+0.116}$} & \textcolor{hred}{$\mathbf{+0.058}$} & \textcolor{hred}{$\mathbf{+0.134}$} \\
FiLM & \textcolor{hred}{$\mathbf{+0.101}$} & $-0.280$ & \textcolor{hred}{$\mathbf{+0.072}$} & \textcolor{hred}{$\mathbf{+0.032}$} & \textcolor{hred}{$\mathbf{+1.356}$} & \textcolor{hred}{$\mathbf{+0.752}$} & $-0.395$ & \textcolor{hred}{$\mathbf{+1.674}$} \\
\bottomrule
\end{tabular}
\end{fittable}
\end{center}
\end{table}

\begin{table}[h]
\caption{\textbf{Gain $g_{\text{own}}$ on Finance\_ret.} Every backbone against every estimator, where a cell is the median over the swept settings and \textrm{TBD} marks a combination that has not finished.}
\label{tab:fullfinanceret}
\begin{center}
\scriptsize
\begin{fittable}\begin{tabular}{lcccccccc}
\toprule
\textbf{Backbone} & \textbf{Sal.} & \textbf{G$\times$I} & \textbf{IG} & \textbf{Occ.} & \textbf{Dyna.} & \textbf{ExtrM.} & \textbf{CLSP} & \textbf{Grad} \\
\midrule
Linear & \textcolor{hred}{$\mathbf{+0.535}$} & \textcolor{hred}{$\mathbf{+0.271}$} & \textcolor{hred}{$\mathbf{+0.233}$} & $-0.007$ & \textcolor{hred}{$\mathbf{+0.923}$} & \textcolor{hred}{$\mathbf{+0.698}$} & \textcolor{hred}{$\mathbf{+0.337}$} & \textcolor{hred}{$\mathbf{+0.591}$} \\
CNN & \textcolor{hred}{$\mathbf{+0.149}$} & \textcolor{hred}{$\mathbf{+0.079}$} & \textcolor{hred}{$\mathbf{+0.057}$} & \textcolor{hred}{$\mathbf{+0.019}$} & \textcolor{hred}{$\mathbf{+0.761}$} & \textcolor{hred}{$\mathbf{+0.148}$} & \textcolor{hred}{$\mathbf{+0.403}$} & \textcolor{hred}{$\mathbf{+0.141}$} \\
Transformer & \textcolor{hred}{$\mathbf{+0.232}$} & \textcolor{hred}{$\mathbf{+0.201}$} & \textcolor{hred}{$\mathbf{+0.203}$} & \textcolor{hred}{$\mathbf{+0.119}$} & \textcolor{hred}{$\mathbf{+0.569}$} & \textcolor{hred}{$\mathbf{+0.242}$} & \textcolor{hred}{$\mathbf{+0.484}$} & \textcolor{hred}{$\mathbf{+0.297}$} \\
DLinear & \textcolor{hred}{$\mathbf{+0.144}$} & \textcolor{hred}{$\mathbf{+0.113}$} & \textcolor{hred}{$\mathbf{+0.152}$} & \textcolor{hred}{$\mathbf{+0.009}$} & \textcolor{hred}{$\mathbf{+1.040}$} & \textcolor{hred}{$\mathbf{+0.039}$} & \textcolor{hred}{$\mathbf{+0.079}$} & \textcolor{hred}{$\mathbf{+0.349}$} \\
TSMixer & \textcolor{hred}{$\mathbf{+0.229}$} & \textcolor{hred}{$\mathbf{+0.179}$} & \textcolor{hred}{$\mathbf{+0.105}$} & \textcolor{hred}{$\mathbf{+0.011}$} & \textcolor{hred}{$\mathbf{+0.774}$} & $-0.014$ & \textcolor{hred}{$\mathbf{+0.102}$} & \textcolor{hred}{$\mathbf{+0.305}$} \\
LightTS & \textcolor{hred}{$\mathbf{+0.316}$} & \textcolor{hred}{$\mathbf{+0.133}$} & \textcolor{hred}{$\mathbf{+0.199}$} & \textcolor{hred}{$\mathbf{+0.064}$} & \textcolor{hred}{$\mathbf{+1.361}$} & $-0.021$ & \textcolor{hred}{$\mathbf{+0.205}$} & \textcolor{hred}{$\mathbf{+0.614}$} \\
\bottomrule
\end{tabular}
\end{fittable}
\end{center}
\end{table}

\begin{table}[h]
\caption{\textbf{Gain $g_{\text{own}}$ on Solar.} Every backbone against every estimator, where a cell is the median over the swept settings and \textrm{TBD} marks a combination that has not finished.}
\label{tab:fullsolar}
\begin{center}
\scriptsize
\begin{fittable}\begin{tabular}{lcccccccc}
\toprule
\textbf{Backbone} & \textbf{Sal.} & \textbf{G$\times$I} & \textbf{IG} & \textbf{Occ.} & \textbf{Dyna.} & \textbf{ExtrM.} & \textbf{CLSP} & \textbf{Grad} \\
\midrule
Linear & \textcolor{hred}{$\mathbf{+0.123}$} & \textcolor{hred}{$\mathbf{+0.182}$} & \textcolor{hred}{$\mathbf{+0.076}$} & \textcolor{hred}{$\mathbf{+0.047}$} & \textcolor{hred}{$\mathbf{+0.256}$} & \textcolor{hred}{$\mathbf{+0.241}$} & \textcolor{hred}{$\mathbf{+0.284}$} & \textcolor{hred}{$\mathbf{+0.185}$} \\
CNN & \textcolor{hred}{$\mathbf{+0.151}$} & \textcolor{hred}{$\mathbf{+0.344}$} & \textcolor{hred}{$\mathbf{+0.169}$} & \textcolor{hred}{$\mathbf{+0.062}$} & \textcolor{hred}{$\mathbf{+0.298}$} & \textcolor{hred}{$\mathbf{+0.629}$} & \textcolor{hred}{$\mathbf{+1.278}$} & \textcolor{hred}{$\mathbf{+0.211}$} \\
Transformer & \textcolor{hred}{$\mathbf{+0.100}$} & \textcolor{hred}{$\mathbf{+0.081}$} & \textcolor{hred}{$\mathbf{+0.035}$} & \textcolor{hred}{$\mathbf{+0.011}$} & \textcolor{hred}{$\mathbf{+0.086}$} & \textcolor{hred}{$\mathbf{+0.180}$} & \textcolor{hred}{$\mathbf{+0.203}$} & \textcolor{hred}{$\mathbf{+0.127}$} \\
DLinear & \textcolor{hred}{$\mathbf{+0.186}$} & \textcolor{hred}{$\mathbf{+0.137}$} & \textcolor{hred}{$\mathbf{+0.088}$} & \textcolor{hred}{$\mathbf{+0.045}$} & \textcolor{hred}{$\mathbf{+0.414}$} & \textcolor{hred}{$\mathbf{+0.468}$} & \textcolor{hred}{$\mathbf{+0.740}$} & \textcolor{hred}{$\mathbf{+0.174}$} \\
TSMixer & \textcolor{hred}{$\mathbf{+0.100}$} & \textcolor{hred}{$\mathbf{+0.085}$} & \textcolor{hred}{$\mathbf{+0.097}$} & \textcolor{hred}{$\mathbf{+0.013}$} & \textcolor{hred}{$\mathbf{+0.339}$} & \textcolor{hred}{$\mathbf{+0.209}$} & \textcolor{hred}{$\mathbf{+0.214}$} & \textcolor{hred}{$\mathbf{+0.143}$} \\
\bottomrule
\end{tabular}
\end{fittable}
\end{center}
\end{table}

\begin{table}[h]
\caption{\textbf{Gain $g_{\text{own}}$ on Traffic.} Every backbone against every estimator, where a cell is the median over the swept settings and \textrm{TBD} marks a combination that has not finished.}
\label{tab:fulltraffic}
\begin{center}
\scriptsize
\begin{fittable}\begin{tabular}{lcccccccc}
\toprule
\textbf{Backbone} & \textbf{Sal.} & \textbf{G$\times$I} & \textbf{IG} & \textbf{Occ.} & \textbf{Dyna.} & \textbf{ExtrM.} & \textbf{CLSP} & \textbf{Grad} \\
\midrule
Linear & \textcolor{hred}{$\mathbf{+0.087}$} & \textcolor{hred}{$\mathbf{+0.110}$} & \textcolor{hred}{$\mathbf{+0.074}$} & \textcolor{hred}{$\mathbf{+0.206}$} & \textcolor{hred}{$\mathbf{+0.474}$} & \textcolor{hred}{$\mathbf{+0.324}$} & \textcolor{hred}{$\mathbf{+0.417}$} & \textcolor{hred}{$\mathbf{+0.378}$} \\
CNN & \textcolor{hred}{$\mathbf{+0.028}$} & \textcolor{hred}{$\mathbf{+0.008}$} & \textcolor{hred}{$\mathbf{+0.009}$} & \textcolor{hred}{$\mathbf{+0.115}$} & \textcolor{hred}{$\mathbf{+0.183}$} & \textcolor{hred}{$\mathbf{+0.333}$} & \textcolor{hred}{$\mathbf{+0.486}$} & $-0.027$ \\
Transformer & \textcolor{hred}{$\mathbf{+0.113}$} & \textcolor{hred}{$\mathbf{+0.093}$} & \textcolor{hred}{$\mathbf{+0.048}$} & \textcolor{hred}{$\mathbf{+0.067}$} & \textcolor{hred}{$\mathbf{+0.262}$} & \textcolor{hred}{$\mathbf{+0.230}$} & \textcolor{hred}{$\mathbf{+0.293}$} & \textcolor{hred}{$\mathbf{+0.177}$} \\
DLinear & \textcolor{hred}{$\mathbf{+0.076}$} & \textcolor{hred}{$\mathbf{+0.090}$} & \textcolor{hred}{$\mathbf{+0.079}$} & \textcolor{hred}{$\mathbf{+0.188}$} & \textcolor{hred}{$\mathbf{+0.467}$} & \textcolor{hred}{$\mathbf{+0.268}$} & \textcolor{hred}{$\mathbf{+0.378}$} & \textcolor{hred}{$\mathbf{+0.377}$} \\
TSMixer & \textcolor{hred}{$\mathbf{+0.123}$} & \textcolor{hred}{$\mathbf{+0.053}$} & \textcolor{hred}{$\mathbf{+0.078}$} & \textcolor{hred}{$\mathbf{+0.051}$} & \textcolor{hred}{$\mathbf{+0.391}$} & \textcolor{hred}{$\mathbf{+0.146}$} & \textcolor{hred}{$\mathbf{+0.205}$} & \textcolor{hred}{$\mathbf{+0.174}$} \\
\bottomrule
\end{tabular}
\end{fittable}
\end{center}
\end{table}

\begin{table}[h]
\caption{\textbf{Gain $g_{\text{own}}$ on Weather.} Every backbone against every estimator, where a cell is the median over the swept settings and \textrm{TBD} marks a combination that has not finished.}
\label{tab:fullweather}
\begin{center}
\scriptsize
\begin{fittable}\begin{tabular}{lcccccccc}
\toprule
\textbf{Backbone} & \textbf{Sal.} & \textbf{G$\times$I} & \textbf{IG} & \textbf{Occ.} & \textbf{Dyna.} & \textbf{ExtrM.} & \textbf{CLSP} & \textbf{Grad} \\
\midrule
Linear & \textcolor{hred}{$\mathbf{+0.098}$} & \textcolor{hred}{$\mathbf{+0.072}$} & \textcolor{hred}{$\mathbf{+0.063}$} & \textcolor{hred}{$\mathbf{+0.001}$} & \textcolor{hred}{$\mathbf{+0.108}$} & \textcolor{hred}{$\mathbf{+0.048}$} & \textcolor{hred}{$\mathbf{+0.064}$} & \textcolor{hred}{$\mathbf{+0.081}$} \\
CNN & \textcolor{hred}{$\mathbf{+0.182}$} & \textcolor{hred}{$\mathbf{+0.168}$} & \textcolor{hred}{$\mathbf{+0.165}$} & \textcolor{hred}{$\mathbf{+0.095}$} & \textcolor{hred}{$\mathbf{+0.043}$} & \textcolor{hred}{$\mathbf{+0.732}$} & \textcolor{hred}{$\mathbf{+0.860}$} & \textcolor{hred}{$\mathbf{+0.125}$} \\
Transformer & \textcolor{hred}{$\mathbf{+0.018}$} & $-0.007$ & $-0.007$ & \textcolor{hred}{$\mathbf{+0.011}$} & \textcolor{hred}{$\mathbf{+0.077}$} & \textcolor{hred}{$\mathbf{+0.027}$} & \textcolor{hred}{$\mathbf{+0.044}$} & \textcolor{hred}{$\mathbf{+0.011}$} \\
DLinear & \textcolor{hred}{$\mathbf{+0.099}$} & \textcolor{hred}{$\mathbf{+0.079}$} & \textcolor{hred}{$\mathbf{+0.074}$} & \textcolor{hred}{$\mathbf{+0.012}$} & \textcolor{hred}{$\mathbf{+0.165}$} & \textcolor{hred}{$\mathbf{+0.085}$} & \textcolor{hred}{$\mathbf{+0.066}$} & \textcolor{hred}{$\mathbf{+0.079}$} \\
TSMixer & \textcolor{hred}{$\mathbf{+0.212}$} & \textcolor{hred}{$\mathbf{+0.067}$} & \textcolor{hred}{$\mathbf{+0.077}$} & $-0.008$ & \textcolor{hred}{$\mathbf{+0.348}$} & \textcolor{hred}{$\mathbf{+0.109}$} & \textcolor{hred}{$\mathbf{+0.225}$} & \textcolor{hred}{$\mathbf{+0.283}$} \\
LightTS & \textcolor{hred}{$\mathbf{+0.142}$} & \textcolor{hred}{$\mathbf{+0.051}$} & \textcolor{hred}{$\mathbf{+0.075}$} & $-0.007$ & \textcolor{hred}{$\mathbf{+0.475}$} & \textcolor{hred}{$\mathbf{+0.507}$} & \textcolor{hred}{$\mathbf{+0.803}$} & \textcolor{hred}{$\mathbf{+0.194}$} \\
\bottomrule
\end{tabular}
\end{fittable}
\end{center}
\end{table}

\end{document}